\documentclass[10pt,twocolumn,letterpaper]{article}

\newif\ifarxiv
\arxivtrue

\ifarxiv
\usepackage[pagenumbers]{cvpr} %
\else
\usepackage{cvpr}              %
\fi

\usepackage{amsmath}

\usepackage{comment}

\usepackage{multirow}
\usepackage[table]{xcolor}
\usepackage{overpic}
\usepackage{pifont}
\usepackage{xparse}
\usepackage{makecell}

\usepackage[accsupp]{axessibility}  %

\newcommand{\cmark}{\ding{51}}%
\newcommand{\xmark}{\ding{55}}%

\newcounter{subsubsubsection}[subsubsection]

\newcommand{\sssubsection}[2][]{%
  \refstepcounter{subsubsubsection}%
  \phantomsection%
  \textbf{#2}%
  \ifx&#1&\else\label{#1}\fi%
}

\newcommand{\notsosmall}{\fontsize{10pt}{12pt}\selectfont}

\definecolor{mycyan}{HTML}{00FFFF}

\definecolor{firstcolor}{HTML}{BDE6CD}%
\definecolor{secondcolor}{HTML}{E2EEBC}%
\definecolor{thirdcolor}{HTML}{FFF8C5}%

\newcommand{\fst}[1]{\cellcolor{firstcolor}\bfseries #1}
\newcommand{\snd}[1]{\cellcolor{secondcolor}#1}

\definecolor{cvprblue}{rgb}{0.21,0.49,0.74}
\ifarxiv
\usepackage[breaklinks,colorlinks,allcolors=cvprblue,bookmarks=true,bookmarksnumbered=true]{hyperref}
\hypersetup{
  pdftitle={EventHub: Data Factory for Generalizable Event-Based Stereo Networks without Active Sensors},
  pdfsubject={Computer Vision, Deep Learning, Robotics},
  pdfauthor={Luca Bartolomei, Fabio Tosi, Matteo Poggi, Stefano Mattoccia, Guillermo Gallego},
  pdfkeywords={Event Cameras, Stereo, Depth Estimation, 3D reconstruction}
}
\usepackage[absolute]{textpos}
\else
\usepackage[breaklinks,colorlinks,allcolors=cvprblue]{hyperref} %
\fi

\title{EventHub: Data Factory for Generalizable Event-Based Stereo Networks without Active Sensors}

\author{Luca Bartolomei$^{1,2,3}$%
\hspace{1.5cm} Fabio Tosi$^2$ \hspace{1.5cm} Matteo Poggi$^{1,2}$  \\ Stefano Mattoccia$^{1,2}$ \hspace{1.5cm} Guillermo Gallego$^{3}$ \smallskip \\
\begin{tabular}{ccc}
    \notsosmall $^1$Advanced Research Center on Electronic System (ARCES) & \hspace{1cm} & \notsosmall $^3$ \notsosmall TU Berlin, Robotics Institute Germany, \vspace{-0.1cm}\\
    \notsosmall $^2$Department of Computer Science and Engineering (DISI) & & \notsosmall Einstein Center Digital Future, \vspace{-0.1cm}\\
    \notsosmall University of Bologna, Italy & & \notsosmall SCIoI Excellence Cluster, Germany\\ %
\end{tabular} \\[1.2ex]
{Project page: \url{https://bartn8.github.io/eventhub}}
}

\begin{document}
\ifarxiv
\definecolor{somegray}{gray}{0.5}
\newcommand{\darkgrayed}[1]{\textcolor{somegray}{#1}}
\begin{textblock}{11.5}(2.25, 0.8)  %
\begin{center}
\darkgrayed{This paper has been accepted for publication at the\\
IEEE Conference on Computer Vision and Pattern Recognition (CVPR), Denver, 2026.
\copyright IEEE}
\end{center}
\end{textblock}
\fi

\twocolumn[{
\renewcommand\twocolumn[1][]{#1}
\maketitle
\begin{center} 
    \vspace{-.8cm}
    \renewcommand{\tabcolsep}{1pt}
    \resizebox{\textwidth}{!}{
    \begin{tabular}{ccccccccc}
         &\hspace{1.2cm}\rotatebox[origin=l]{90}{\centering\scriptsize\textbf{EventHub Train Data}} & \begin{overpic}[width=0.19\linewidth]{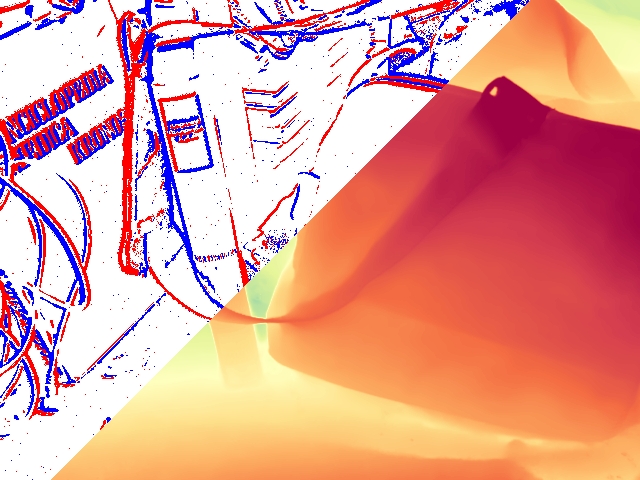}
         \setlength{\fboxsep}{0.5pt}
         \setlength{\fboxrule}{0.5pt}
         \put(25,68){\colorbox{white}{\color{black} \small \textbf{NeRFSt~\cite{tosi2023nerf}}}}
         \end{overpic} & 
         \begin{overpic}[width=0.19\linewidth]{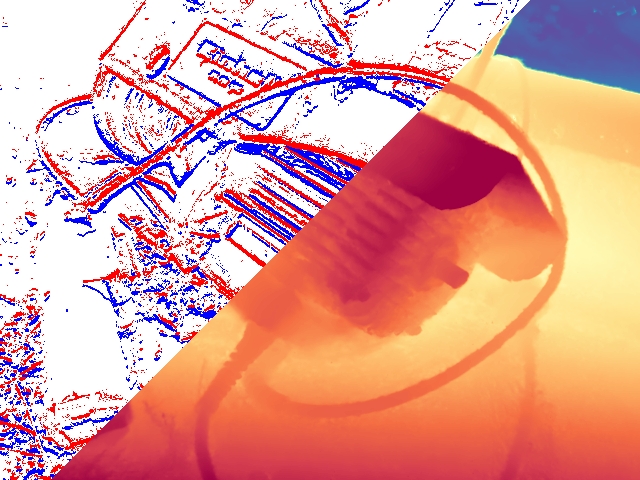} 
         \setlength{\fboxsep}{0.5pt}
         \setlength{\fboxrule}{0.5pt}
         \put(25,68){\colorbox{white}{\color{black} \small \textbf{NeRFSt~\cite{tosi2023nerf}}}}
         \end{overpic} & \;\;\; &
         \begin{overpic}[width=0.19\linewidth]{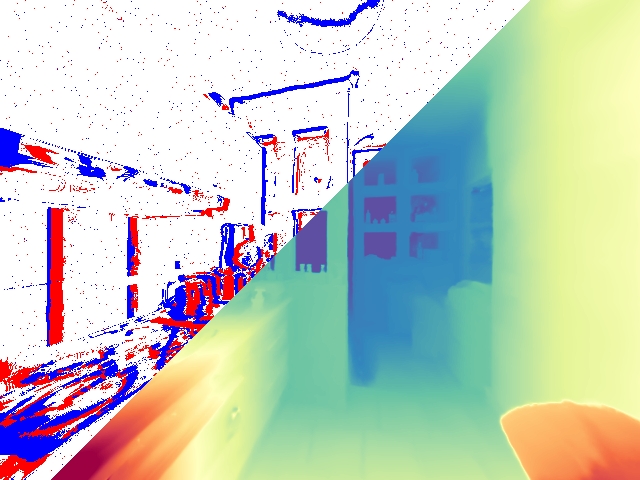} 
         \setlength{\fboxsep}{0.5pt}
         \setlength{\fboxrule}{0.5pt}
         \put(20,68){\colorbox{white}{\color{black} \small \textbf{ScanNet++~\cite{yeshwanth2023scannet}}}}
         \end{overpic} & 
         \begin{overpic}[width=0.19\linewidth]{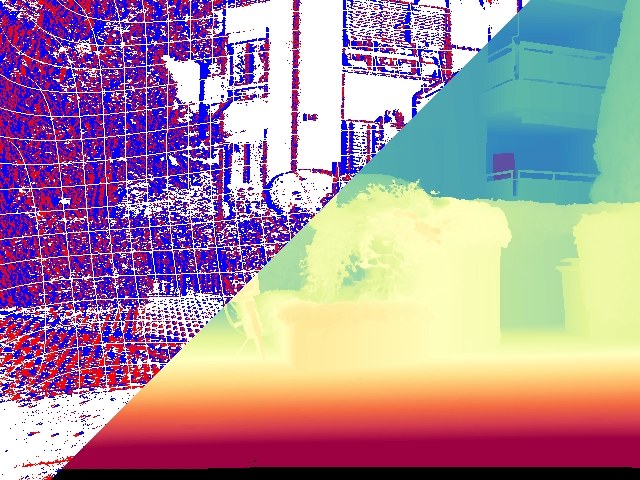} 
         \setlength{\fboxsep}{0.5pt}
         \setlength{\fboxrule}{0.5pt}
         \put(27,68){\colorbox{white}{\color{black} \small \textbf{DSEC~\cite{gehrig2021dsec}}}}
         \end{overpic} & \\ %
        \midrule [1pt] 

         \raisebox{-.58\height}[0pt][0pt]{\rotatebox[origin=l]{90}{\centering\scriptsize\textbf{Test on DSEC \cite{gehrig2021dsec}}}} & \includegraphics[width=0.08\linewidth]{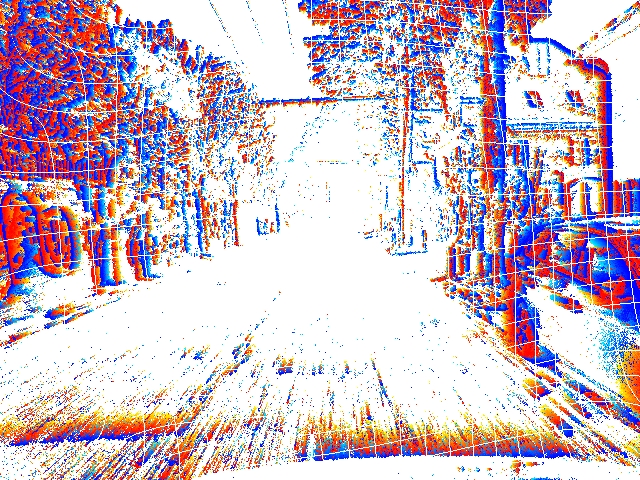} 
         & 
         \raisebox{-.58\height}[0pt][0pt]{
         \begin{overpic}[width=0.19\linewidth]{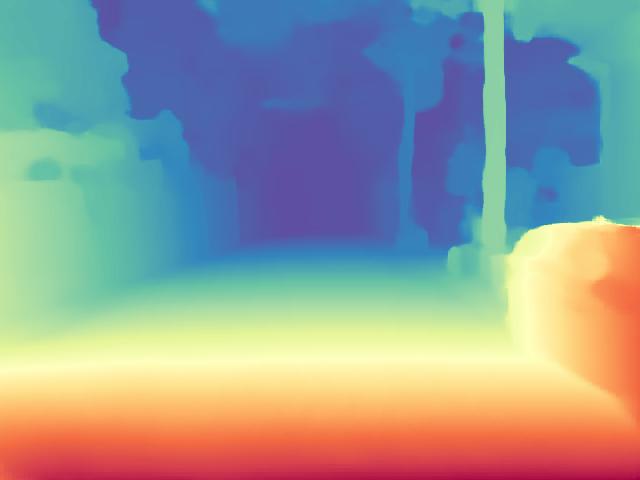}
         {
         \setlength{\fboxsep}{0.5pt}
         \setlength{\fboxrule}{0.5pt}
         \put(25,68){\colorbox{white}{\color{black} \small \textbf{EMatch~\cite{zhang2025ematch}}}}
         \put(0,2){{\color{white} \footnotesize \textbf{MAE 0.95px}}}%
         }
         \put(85,0){\Huge \textbf{\color{black} \cmark}}
         \put(85.5,0.25){\huge \textbf{\color{green} \cmark}}
         \end{overpic}
         }& 
         \raisebox{-.58\height}[0pt][0pt]{
         \begin{overpic}[width=0.19\linewidth]{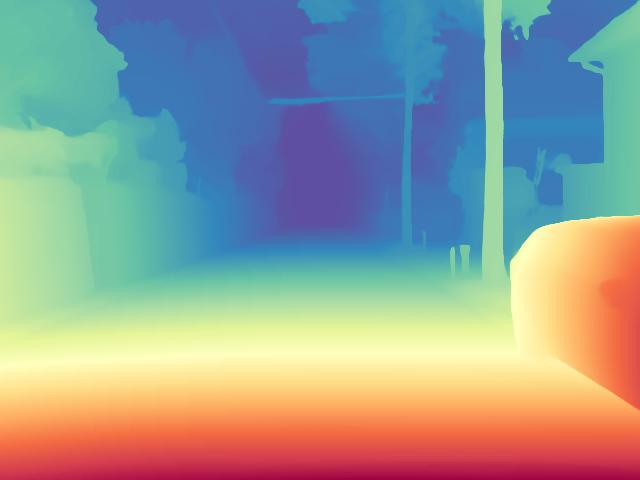}
         {
         \setlength{\fboxsep}{1pt}
         \setlength{\fboxrule}{1pt}
         \put(5,68.75){\colorbox{white}{\color{black} \scriptsize \textbf{E-FoundationStereo (Ours)}}}
         \put(0,2){{\color{white} \footnotesize \textbf{MAE 0.89px}}}%
         }
         \put(85,0){\Huge \textbf{\color{black} \cmark}}
         \put(85.5,0.25){\huge \textbf{\color{green} \cmark}}
         \end{overpic}
         } & \;\;\; &
         \raisebox{-.58\height}[0pt][0pt]{
         \begin{overpic}[width=0.19\linewidth]{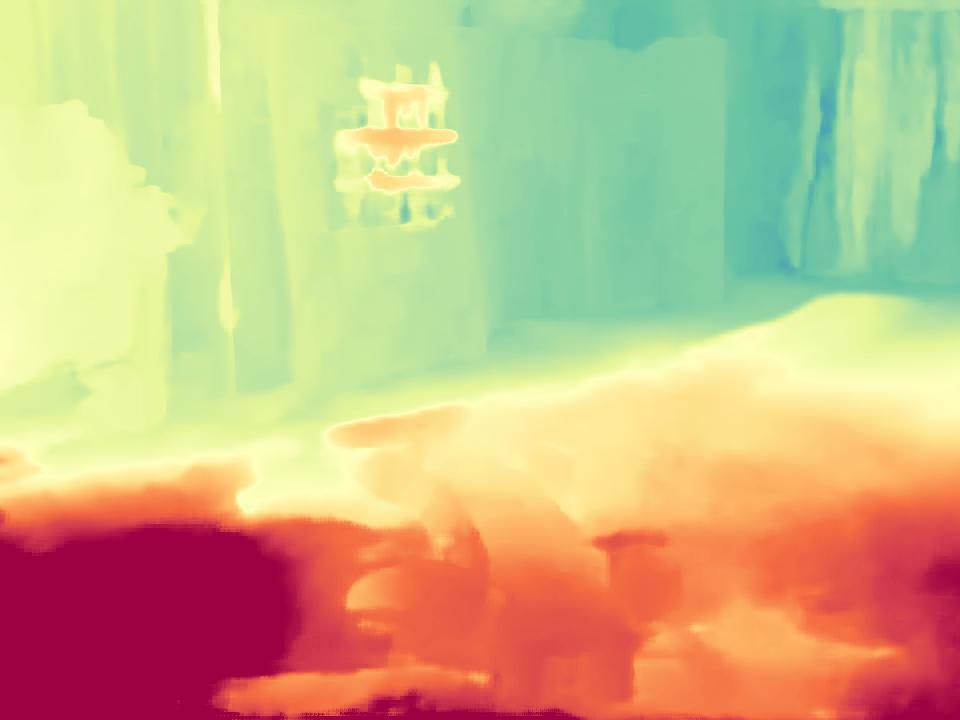}
         {
         \setlength{\fboxsep}{0.5pt}
         \setlength{\fboxrule}{0.5pt}
         \put(25,68){\colorbox{white}{\color{black} \small \textbf{EMatch~\cite{zhang2025ematch}}}}
         \put(0,2){{\color{white} \footnotesize \textbf{MAE 4.05px}}}%
         }
         \put(85,0){\Huge \textbf{\color{black} \xmark}}
         \put(86.25,1.5){\huge \textbf{\color{red} \xmark}}
         \end{overpic}
         }& 
         \raisebox{-.58\height}[0pt][0pt]{
         \begin{overpic}[width=0.19\linewidth]{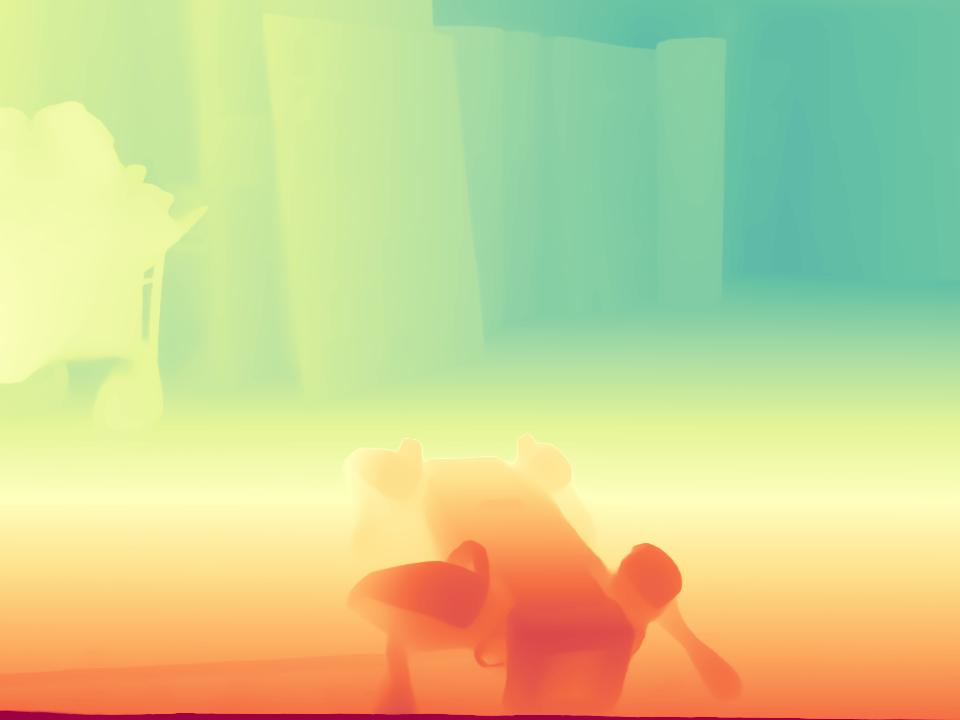}
         {
         \setlength{\fboxsep}{1pt}
         \setlength{\fboxrule}{1pt}
         \put(5,68.75){\colorbox{white}{\color{black} \scriptsize \textbf{E-FoundationStereo (Ours)}}}
         \put(0,2){{\color{white} \footnotesize \textbf{MAE 2.53px}}}%
         }         
         \put(85,0){\Huge \textbf{\color{black} \cmark}}
         \put(85.5,0.25){\huge \textbf{\color{green} \cmark}}
         \end{overpic} 
         } & \includegraphics[width=0.08\linewidth]{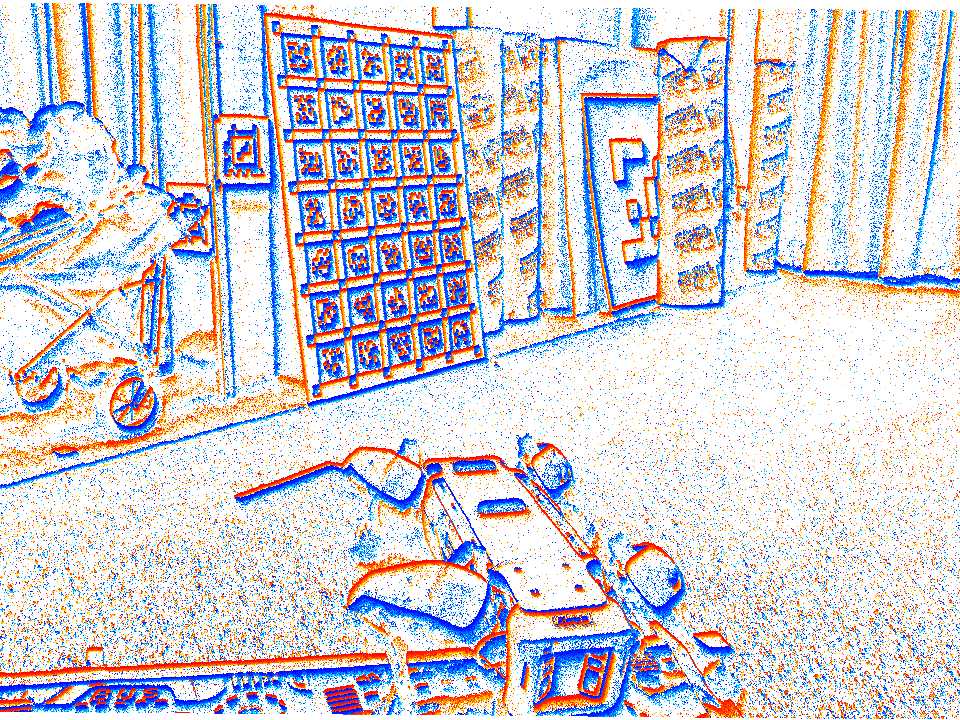}  \\[-1pt]     
         & \includegraphics[width=0.08\linewidth]{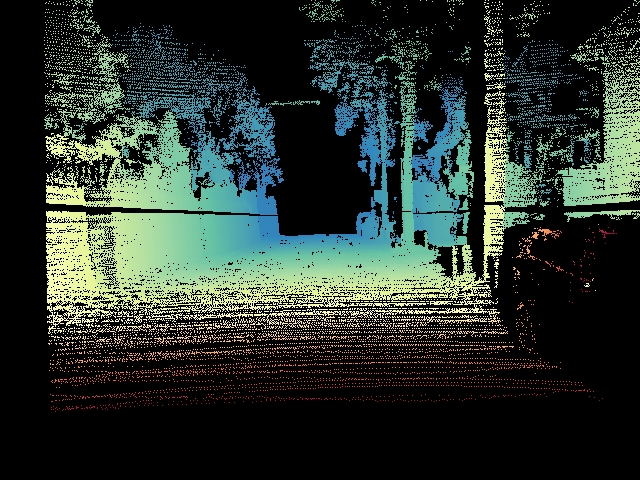} &&&&&& \includegraphics[width=0.08\linewidth]{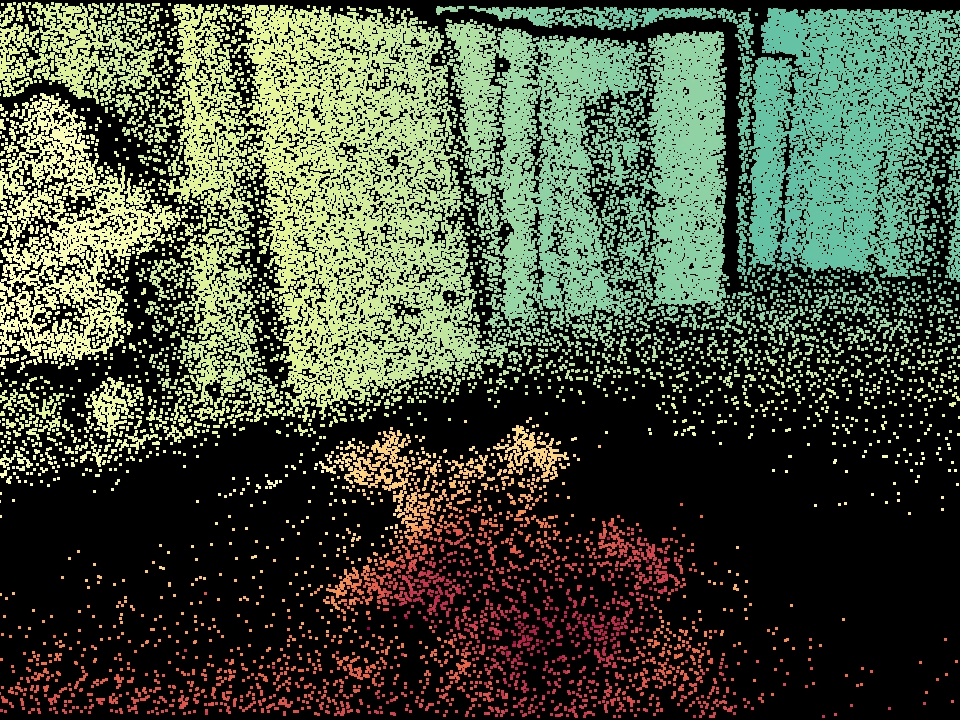} & \raisebox{-.58\height}[0pt][0pt]{\rotatebox[origin=r]{270}{\hspace{-2.5cm}\centering\scriptsize\textbf{Test on M3ED \cite{chaney2023m3ed}}}}\\
         
    \end{tabular}
    }
    \vspace{-0.1cm}
    \captionof{figure}{\textbf{EventHub: LiDAR-free proxy data for robust event stereo.} 
    Our factory generates training data from multiple sources~\cite{tosi2023nerf,yeshwanth2023scannet,gehrig2021dsec} (top), allowing our E-FoundationStereo to match EMatch~\cite{zhang2025ematch} in-domain~\cite{gehrig2021dsec} and outperform it in generalization~\cite{chaney2023m3ed} (bottom).}
    \label{fig:teaser}%
\end{center}
}]

\begin{abstract}

We propose EventHub, a novel framework for training deep-event stereo networks without ground truth annotations from costly active sensors, relying instead on standard color images. 
From these images, we derive either proxy annotations and proxy events through state-of-the-art novel view synthesis techniques, or simply proxy annotations when images are already paired with event data. 
Using the training set generated by our data factory, we repurpose state-of-the-art stereo models from RGB literature to process event data, obtaining new event stereo models with unprecedented generalization capabilities.
Experiments on widely used event stereo datasets support the effectiveness of EventHub and show how the same data distillation mechanism can improve the accuracy of RGB stereo foundation models in challenging conditions such as nighttime scenes.

\end{abstract}
    
\section{Introduction}
\label{sec:intro}

Now nearing its fiftieth anniversary~\cite{tosi2025survey}, stereo matching has undergone rapid evolution over the past decade thanks to deep learning \cite{tosi2025survey}, thus enabling high-accuracy and high-resolution depth maps that are crucial for applications such as autonomous driving, 3D scene reconstruction, augmented reality, and robotic navigation.
Recent deep-based stereo models achieved remarkable performance \cite{wen2025foundationstereo,bartolomei2025stereo}, also in a zero-shot manner, thanks to a large quantity of labeled data -- i.e., millions of labeled synthetic and real images.
The acquisition of those images required years of incredible efforts from the community: starting from sophisticated active setups \cite{scharstein2014high} to achieve high-accuracy real datasets, to the usage of large computing resources to render large-scale photo-realistic synthetic datasets \cite{wen2025foundationstereo}.

Recently, the introduction of the first commercial event cameras \cite{lichtsteiner2008128} led to the creation of a novel branch of stereo literature that aims to estimate depth from a pair of synchronized event cameras \cite{ghosh2025event}.
These sensors capture asynchronous per-pixel brightness changes occurring in the scene, so-called ``events'' \cite{gallego2020event,Posch14ieee}. 
An event is characterized by a pixel coordinate (the location where the change occurred), a timestamp (when it occurred), and a polarity ($\pm 1$, indicating whether brightness increased or decreased).
The asynchronous working principle enables these sensors to capture information at microsecond resolution, allowing them to surpass traditional frame-based cameras in challenging scenarios, such as fast motion (resulting in no motion blur) and high dynamic (resulting in no over/under-exposure).
As a drawback, adapting the large body of image-based computer vision algorithms to event cameras is not trivial due to the very sparse nature of events \cite{gallego2020event}.

Despite the growing interest in event-based stereo matching, the availability of labeled datasets remains very limited compared to the traditional frame-based domain \cite{ghosh2025event}.
Capturing dense and accurate ground truth for asynchronous event streams is significantly more
challenging due to the still-emergent event community and the substantial deviation from traditional frame-based cameras.

In this paper, we aim to introduce a novel framework for training deep-based event stereo networks effortlessly and without any ground-truth.
By leveraging state-of-the-art novel view synthesis solutions \cite{sun2025sparse}, we can generate event stereo training data from image sequences collected with a single color camera, alongside with proxy depth labels.
In alternative, when paired RGB stereo images and event stereo data are available, we distill the knowledge of stereo foundation models processing to annotate the latter. 
This approach drastically reduces the need for complex data acquisition setups and large-scale manual labeling efforts, democratizing access to high-quality training data for event-based stereo (\cref{fig:teaser}). 
With our data, we then repurpose stereo foundation models to obtain a new generation of state-of-the-art, event stereo models with unparalleled generalization capabilities, which can be used in turn to further improve the original color models in challenging scenarios.

We summarize our main contributions as follows:

\begin{itemize}
    \item We propose \emph{EventHub}, the first framework combining neural rendering data generation and cross-modal distillation from RGB stereo foundation models to train event stereo networks without active sensor supervision.%
    \item We demonstrate superior out-of-domain generalization compared to LiDAR-supervised models, reducing error by up to 50\% on M3ED and MVSEC datasets.
    \item We establish bi-directional knowledge transfer between RGB and event modalities, enabling event models to improve the performance of RGB stereo foundation models in challenging nighttime conditions.
\end{itemize}

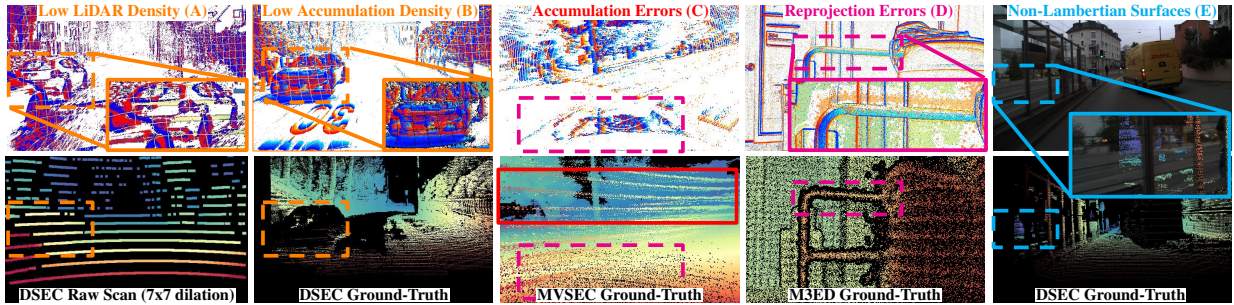
\begin{figure*}
    \centering
    \scalebox{0.62}{
    \begin{overpic}[abs,unit=1mm,scale=.25]{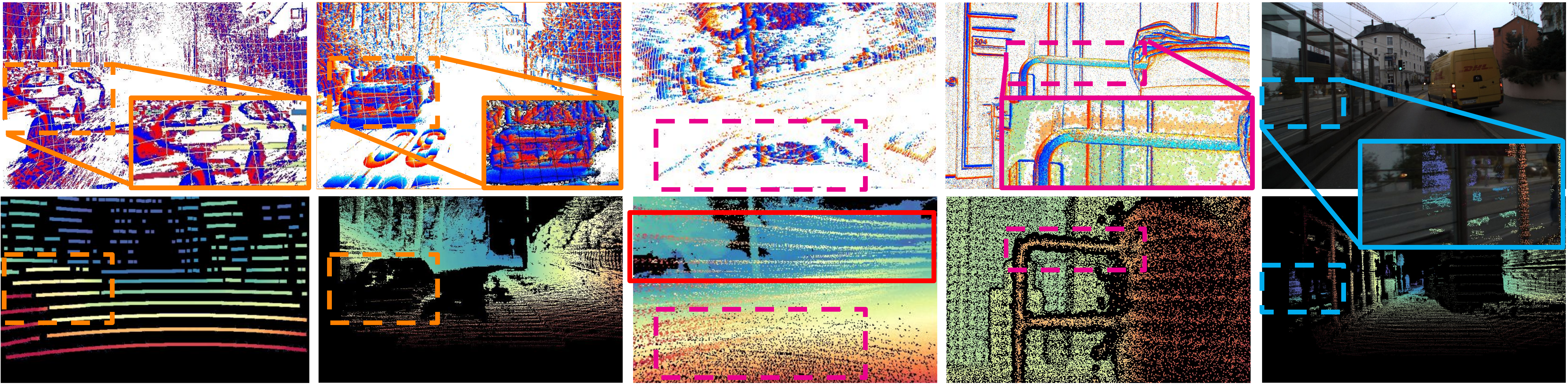}

        \setlength{\fboxsep}{0.5pt}
        \setlength{\fboxrule}{0.5pt}

        \put(7,62){\colorbox{white}{\color{orange}\textbf{Low LiDAR Density (A)}}}
        \put(55,62){\colorbox{white}{\color{orange}\textbf{Low Accumulation Density (B)}}}
        \put(113,62){\colorbox{white}{\color{red}\textbf{Accumulation Errors (C)}}}
        \put(167,62){\colorbox{white}{\color{magenta}\textbf{Reprojection Errors (D)}}}
        \put(215,62){\colorbox{white}{\color{cyan}\textbf{Non-Lambertian Surfaces (E)}}}
        
        \put(3,1){\colorbox{white}{\textbf{DSEC Raw Scan (7x7 dilation)}}}
        \put(63,1){\colorbox{white}{\textbf{DSEC Ground-Truth}}}
        \put(114,1){\colorbox{white}{\textbf{MVSEC Ground-Truth}}}
        \put(168,1){\colorbox{white}{\textbf{M3ED Ground-Truth}}}
        \put(221,1){\colorbox{white}{\textbf{DSEC Ground-Truth}}}
    \end{overpic}}\vspace{-0.2cm}
    \captionof{figure}{\textbf{Limitations of LiDAR-supervised real-world datasets.} 
    Despite their popularity \cite{gehrig2021dsec, chaney2023m3ed, zhu2018multivehicle}, LiDAR annotations remain sparse ({A}), poorly capture dynamic scenes ({B–C}), are prone to reprojection errors ({D}), and struggle on transparent or reflective surfaces ({E}).}\vspace{-0.3cm}
    \label{fig:real_world_datasets_errors}
\end{figure*}

\section{Related Work}
\label{sec:related}

\textbf{Frame-based Stereo.} 
Stereo depth estimation has transitioned from traditional hand-crafted approaches~\cite{scharstein2002taxonomy} to data-driven solutions~\cite{tosi2025survey,poggi2021synergies, laga2020survey}. 
Early learning-based methods~\cite{zbontar2015computing,luo2016efficient} focused on individual matching components, while later works~\cite{mayer2016large,kendall2017end_GC-NET,chang2018pyramid,xu2020aanet} introduced fully trainable pipelines combining feature extraction, cost aggregation, and disparity prediction, by leveraging the abundance of synthetic data \cite{mayer2016large,tartanair2020iros} for training. 
Building on optical flow principles, recurrent architectures~\cite{lipson2021raft,xu2023iterative,wang2024selective} performed iterative refinement over correlation volumes. 
Transformer models~\cite{guo2022context_CEST,Li_2021_ICCV_STTR,xu2023unifying} further enhanced matching through global attention mechanisms. 
Generalizing across different environments remains challenging. 
Solutions include learning domain-agnostic features~\cite{zhang2019domaininvariant,Zhang_2022_CVPR}, incorporating geometric constraints~\cite{aleotti2021neural,tosi2024neural}, self-supervised learning with photometric consistency~\cite{godard2017unsupervised,Wang_2019_CVPR,Poggi_2024_CVPR}, distillation from traditional methods~\cite{tonioni2017unsupervised,aleotti2020reversing,Chen_2021_ICCV} and radiance field supervision~\cite{tosi2023nerf,ling2024self}. 
Recently, foundation models trained on massive diverse datasets have demonstrated unprecedented zero-shot capabilities~\cite{wen2025foundationstereo,bartolomei2025stereo,jiang2025defom,cheng2025monster}, establishing a new state of the art. 
However, the scarcity of annotated stereo data in challenging conditions (e.g., night) limits their performance in such scenarios, leaving room for improvement. %

\textbf{Event-based Stereo.} 
While monocular event-based depth estimation~\cite{hidalgo2020learning,lee2025distilE2D,bartolomei2025depth,zhu2025depth} has been explored, we focus on stereo configurations exploiting binocular geometry \cite{ghosh2025event}.
Early stereo methods~\cite{schraml2007smartcam,kogler2011event} relied on temporal coincidence matching via frame accumulation or event-driven search, later enhanced with epipolar geometry and temporal-luminance constraints~\cite{rogister2011asynchronous,ieng2018neuromorphic}. 
Neuromorphic implementations~\cite{firouzi2016asynchronous,osswald2017spiking} deployed cooperative networks on specialized spiking hardware, while deep learning approaches~\cite{tulyakov2019learning,ahmed2021deep,zhang2022discrete} introduced learnable representations and spatio-temporal encoders. 
Recent works incorporate temporal context~\cite{cho2024temporal,nam2022stereo,Ghosh22aisy,Hitzges25neurips}, attention mechanisms~\cite{chen2024event} and unified architectures~\cite{zhang2025ematch} that handle both stereo and optical flow. 
Hybrid configurations combine events with frames: binocular setups (2E+2F)~\cite{mostafavi2021event,cho2022event,nam2022stereo} fuse modalities through recurrent networks or selection mechanisms, while asymmetric systems (1E+1F)~\cite{wang2021stereo,zhang2022data,lou2024zero} address cross-modal alignment challenges. 
Despite progress, event stereo development is severely constrained by the limited amount of annotated data \cite{ghosh2025event}. 
Existing datasets~\cite{gehrig2021dsec,zhu2018multivehicle,chaney2023m3ed} remain orders of magnitude smaller than frame-based ones. 
They also lack diversity, which constrains the ability of models to generalize beyond their training domains. 
This motivates us to seek alternative training strategies.

\textbf{Neural Rendering for Training Data Generation.} 
Neural radiance fields~\cite{mildenhall2021nerf} enable photorealistic novel view synthesis (NVS) from sparse images, facilitating synthetic training data generation. NeRF-supervised frameworks~\cite{tosi2023nerf} rendered stereo pairs from monocular sequences, using synthesized images and depth as proxy supervision for stereo networks.
Similar approaches emerged for optical flow~\cite{ling2024self}, with confidence-based filtering to remove unreliable proxy labels. 
Beyond depth estimation, neural rendering has been exploited for object detection~\cite{ge2022neural}, learning dense descriptors~\cite{yen2022nerf}, semantic labeling~\cite{zhi2021place}, 6D pose estimation~\cite{legrand2024domain} and automated annotation in driving scenes~\cite{huo2025egsral}.
For event cameras, video-to-event simulators~\cite{gehrig2020video} recycle existing datasets into synthetic streams, while specialized engines~\cite{li2023blinkflow} generate event data with optical flow labels. 
Concurrent work GS2E~\cite{li2025gs2e} generates multi-view event data from sparse RGB images via 3D Gaussian Splatting~\cite{kerbl20233d}, and it is mildly tested for NVS and image deblurring. 
However, no prior work has explored neural rendering for generating event stereo training data. 
Leveraging efficient radiance field rendering~\cite{mueller2022instant,kerbl20233d,sun2025sparse,lin2026depth}, we propose the first framework to synthesize stereo event streams from monocular RGB data, enabling large-scale training without active sensors, such as LiDAR.

\begin{figure*}
    \centering
    \resizebox{1.0\textwidth}{!}{
    \begin{overpic}[abs,unit=1mm,scale=.25]{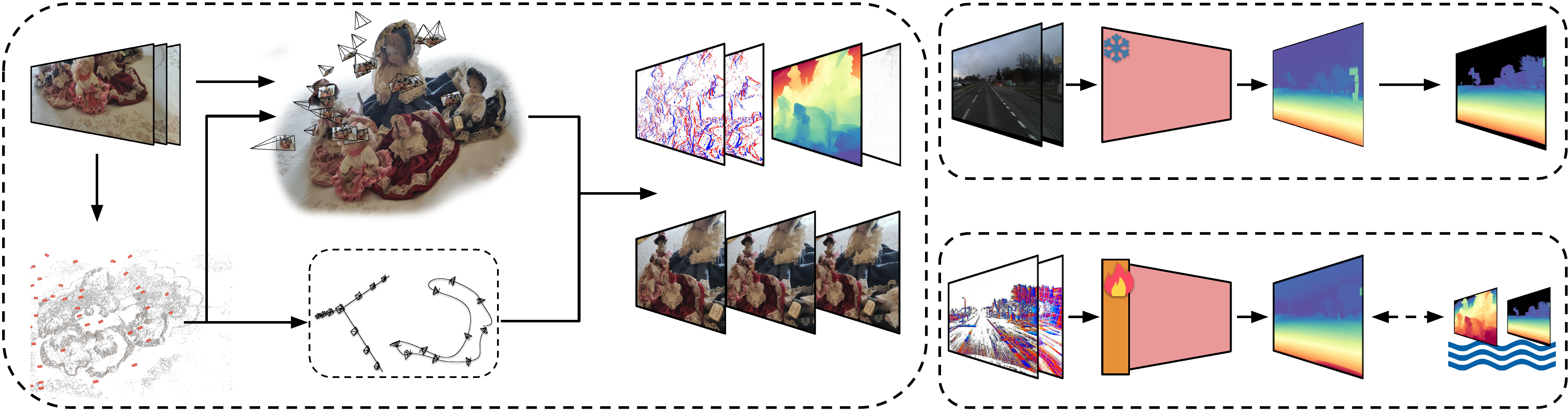}

        \setlength{\fboxsep}{0.2pt}
        \setlength{\fboxrule}{0.2pt}

        \put(3,2){\scriptsize(i)}
        \put(95.5,38){\scriptsize(ii)}
        \put(95.5,15){\scriptsize(iii)}
        
        \put(2.5,37.5){\tiny Multi-View Images (\ref{sub:imagecapture})}
        \put(3.5,17){\tiny COLMAP (\ref{sub:imagecapture})}
        \put(24.5,18){\tiny Regularized Dense 3D Optimization (\ref{sub:svrastertraining})}
        
        \put(27,1.5){\tiny Virtual Trajectory Construction (\ref{sub:virtualtrajectory})}
        \put(32,14){\tiny Trj. Local}
        \put(38,11){\tiny $\Gamma_x$}
        \put(34.5,4.5){\tiny $\Gamma_y$}
        \put(31.25,7.25){\tiny $\Gamma_z$}
        \put(41,14){\tiny Trj. Global}
        \put(44.5,8.5){\tiny $\Omega$}

        \put(64.5,5.5){\tiny Motion-Adaptive Stereo Rendering}
        \put(74,3.5){\tiny (\ref{sub:trinocularrendering})}
        \put(66.5,39){\tiny Rendered}
        \put(65,37.25){\tiny Stereo Events}
        \put(80,39){\tiny Rendered}
        \put(77,37.25){\tiny Depth \& Confidence}
        \put(70,21){\tiny Rendered RGB Triplet}

        \put(97,24.5){\tiny Stereo RGB}
        \put(112,32.25){\tiny \colorbox{White}{\textbf{Teacher-SFM}}}
        \put(115,26.25){\tiny (\ref{sec:method:distillation})}
        \put(111,24.5){\tiny Proxy Estimation}
        \put(126,24.5){\tiny Misaligned Proxy}
        \put(137,34.125){\tiny Reprojection}
        \put(146,24.5){\tiny Aligned Proxy}

        \put(96,1.5){\tiny Stereo Events}
        \put(115,9.4){\tiny \colorbox{White}{\textbf{(\ref{sec:method:adapting_vfm})}}}
        \put(112,7.8){\tiny \colorbox{White}{\textbf{Adapted-SFM}}}
        \put(113,1.5){\tiny Training}
        \put(127.5,1.5){\tiny Disparity Map}
        \put(137,11){\tiny Supervision}
        \put(147,1.5){\tiny EventHub}
    \end{overpic}}\vspace{-0.2cm}
    \caption{\textbf{Framework Overview}: We obtain training data through two complementary approaches: (i) \textbf{Event Data Factory}: SVRaster~\cite{sun2025sparse} generates synthetic event stereo pairs and depth labels from sparse RGB images via virtual camera trajectories (left); (ii) \textbf{Stereo Cross-Modal Distillation}: existing RGB stereo models produce proxy depth labels for real event data in calibrated RGB-Event stereo setups (top right). (iii) Both data sources are combined in EventHub to train/adapt event stereo networks (bottom right).}  
    \label{fig:architecture}
    \vspace{-0.1cm}
\end{figure*}
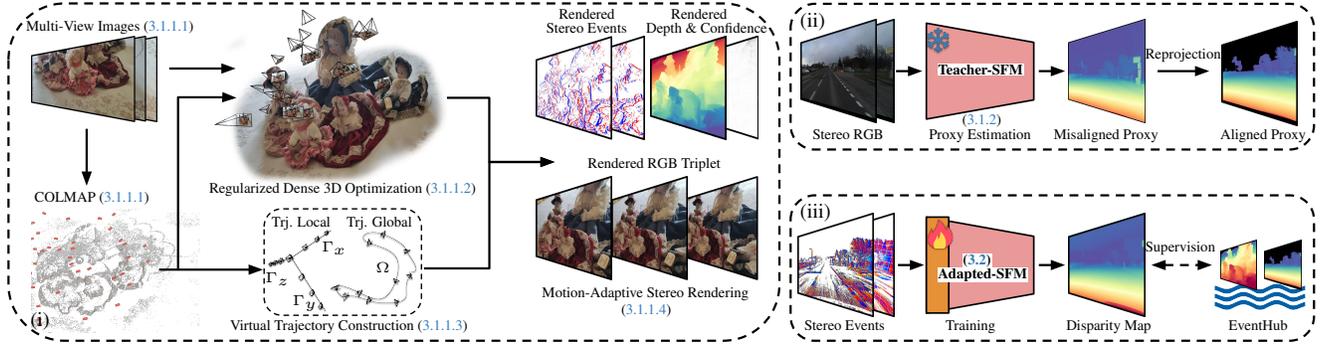

\section{Method}
\label{sec:method}

Sourcing accurate depth labels is costly and time-consuming, as it typically requires the use of active sensors such as LiDARs which, despite their high accuracy, provide very sparse data (\cref{fig:real_world_datasets_errors} (A)).
This limitation is partially mitigated by temporal accumulation; however, this strategy is ineffective in dealing with moving objects --\cref{fig:real_world_datasets_errors} (B) or yields noise due to the motion of dynamic entities-- \cref{fig:real_world_datasets_errors} (C). %
Finally, imprecise calibration or non-Lambertian surfaces also harm the quality of the annotations --\cref{fig:real_world_datasets_errors} (D,E).

Aiming to remove the dependency on noisy labeled data captured with costly LiDAR-based setups, 
we turn to a much cheaper data modality, simpler to obtain and already available in abundance: color images. Through the lens of RGB cameras, we can exploit state-of-the-art depth estimation techniques to annotate data with \emph{proxy labels}, having accuracy not far from the one of LiDAR sensors. 
Most available color images, however, are collected by a single RGB camera, usually navigating through the scene, unpaired with any event camera counterpart. 
In this setting, besides proxy labels, we also need to generate \emph{proxy events}. 
Conversely, when color images are paired with event data collected within the same environment, we can exploit camera calibration and multi-view geometry to annotate the real events, without the need to generate proxy events. 

The overview of our framework is shown in \cref{fig:architecture}.
We develop techniques to extract proxy labels in the two above-mentioned settings: 
(i) one based on NVS frameworks,in which a modified SVRaster~\cite{sun2025sparse} is used to generate both proxy events and proxy labels from RGB sequences (\cref{sec:method:nvsdata}), 
and (ii) one leveraging robust RGB stereo matching in dual RGB–Event stereo setups, where the RGB pair offers the proxy-supervision through state-of-the-art models, such as FoundationStereo~\cite{wen2025foundationstereo} (\cref{sec:method:distillation}).
Moreover, to exploit the knowledge already available in the color image domain, we take a step further by exploring how to adapt pre-trained, robust RGB-based stereo matching networks~\cite{bartolomei2025stereo,wen2025foundationstereo} to the event domain, thereby minimizing the need for labeled event data (\cref{sec:method:adapting_vfm}).

\subsection{EventHub: Data Generation}

\subsubsection{Synthetic Generation via Novel View Synthesis}
\label{sec:method:nvsdata}

Given sparse RGB images of a static scene, NVS frameworks~\cite{mildenhall2021nerf,kerbl20233d} reconstruct high-fidelity digital representations
that can be rendered from arbitrary viewpoints. %
While NeRF-based data factories for RGB stereo exist~\cite{tosi2023nerf}, an equivalent pipeline for the event-based stereo domain remains unexplored: NVS frameworks typically output static frames rather than events, requiring fast rendering to match the event camera's temporal resolution, plus additional components to handle frames-to-events generation and motion trajectories.
Therefore, we propose a novel pipeline for event stereo data generation by leveraging SVRaster~\cite{sun2025sparse} as the NVS foundation framework. 
We now describe the proposed pipeline step by step.

\sssubsection[sub:imagecapture]{Image Capture and Camera Calibration.}
After collecting $N$ multi-view RGB images $\hat{\mathbf{I}}_i$ of a static scene, we follow~\cite{tosi2023nerf} and deploy COLMAP~\cite{schonberger2016structure} to recover intrinsics $\hat{\mathbf{K}}\in\mathbb{R}^{3\times3}$ and $N$ camera poses $[\hat{\mathbf{R}}_i|\hat{\mathbf{t}}_i]=\hat{\mathbf{T}}_i \in \mathbb{SE}(3)$.%

\sssubsection[sub:svrastertraining]{Regularized Dense 3D Optimization.}
Next, for each captured scene, we fed $\hat{\mathbf{I}}_i$, $\hat{\mathbf{K}}$, and $\hat{\mathbf{T}}_i$ to SVRaster's training pipeline, obtaining a radiance representation of the scene. 
We follow~\cite{sun2025sparse} and use both MSE and SSIM to optimize the rendered image.
The rendered color $\mathbf{I}$ and corresponding depth $\mathbf{Z}$ along each camera ray are defined as:
\begin{equation}
    \mathbf{I} = \sum_{i=1}^N T_i \alpha_i \mathbf{c}_i, 
    \; \mathbf{Z} = \sum_{i=1}^N T_i \alpha_i z_i, 
    \; T_i = \prod_{j=1}^{i-1} (1-\alpha_j),
    \label{eq:alpha_rendering}
\end{equation}
where $\alpha_i\in [0,1]$, $T_i\in [0,1]$, $\mathbf{c}_i\in [0,1]^3$, and $z_i > 0$ are the opacity, the transmittance~\cite{mildenhall2021nerf}, the color, and the depth of the $i$-th voxel, respectively.

To further improve depth quality, we applied several regularizers during training: among these, 
(i) $\mathcal{L}_{N-\text{mean}}$ and $\mathcal{L}_{N-\text{med}}$ enforce self-consistency between rendered depth and normals, respectively aggregated using mean and median~\cite{huang20242d}; 
(ii) $\mathcal{L}_\text{DAv2}$ enforces the rendered depth to be consistent with monocular predictions from DepthAnythingV2~\cite{yang2024depth}.
We studied additional regularizers $\mathcal{L}_\text{asc}$, $\mathcal{L}_\text{sparse}$, and $\mathcal{L}_\text{mast3r}$, with further details in the supplementary material.
Each regularizer's contribution is weighted inside a regularization loss 
$\mathcal{L}_\text{reg} \doteq \lambda_{{N-\text{mean}}}\mathcal{L}_{{N-\text{mean}}} + \lambda_{N-\text{med}}\mathcal{L}_{N-\text{med}} + \lambda_{\text{asc}}\mathcal{L}_{\text{asc}} + \lambda_{\text{sparse}}\mathcal{L}_{\text{sparse}} + \lambda_{\text{DAv2}}\mathcal{L}_{\text{DAv2}} + \lambda_{\text{mast3r}}\mathcal{L}_{\text{mast3r}}$, 
yielding the total loss:
\begin{equation}
    \mathcal{L} \doteq \mathcal{L}_\text{MSE} + \lambda_\text{SSIM}\mathcal{L}_\text{SSIM} + \mathcal{L}_\text{reg}.
    \label{eq:final_loss}
\end{equation}

\begin{figure*}[t]
    \centering
    \renewcommand{\tabcolsep}{1pt}
    \resizebox{1.0\textwidth}{!}{
    \begin{tabular}{>{\centering\arraybackslash}p{2.2ex}cccccccc}
    
         \rotatebox{90}{\makecell{\hspace{2pt}NeRF-Stereo~\cite{tosi2023nerf}}} &
         \includegraphics[width=0.20\linewidth]{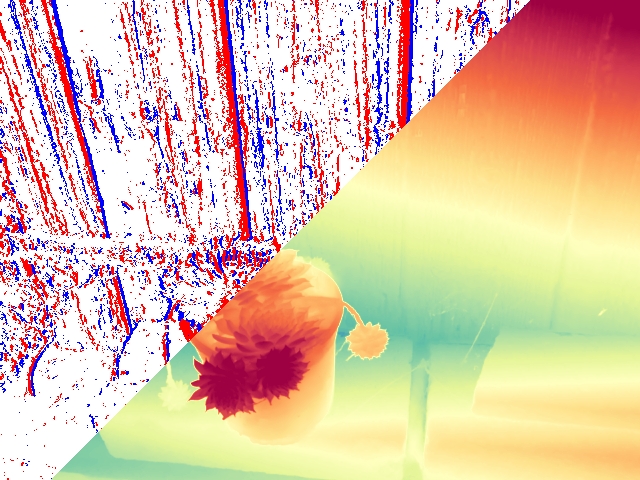} & 
         \includegraphics[width=0.20\linewidth]{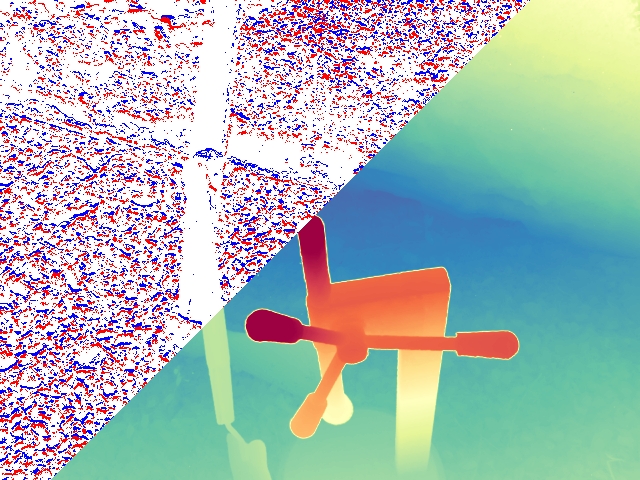} & 
         \includegraphics[width=0.20\linewidth]{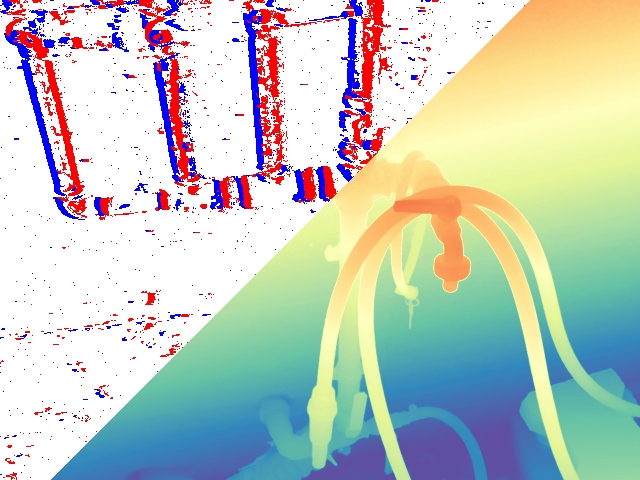} & 
         \includegraphics[width=0.20\linewidth]{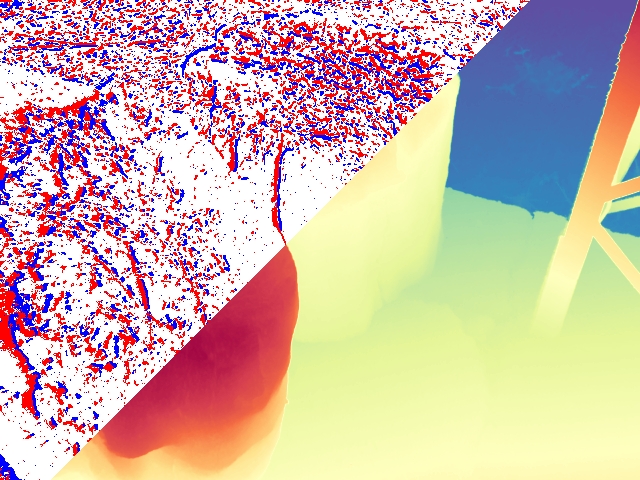} &

         \includegraphics[width=0.20\linewidth]{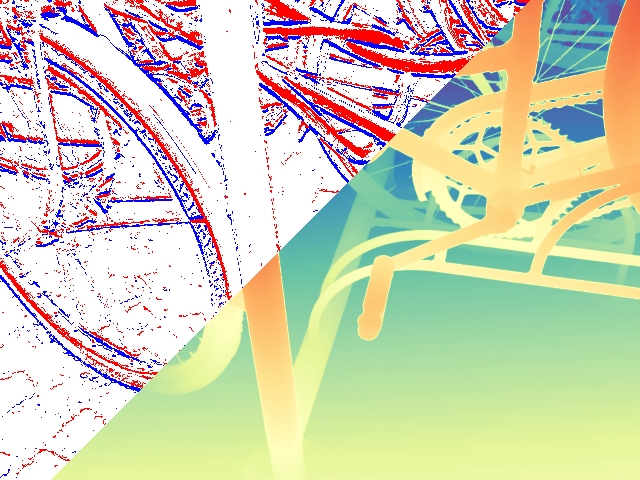} & 
         \includegraphics[width=0.20\linewidth]{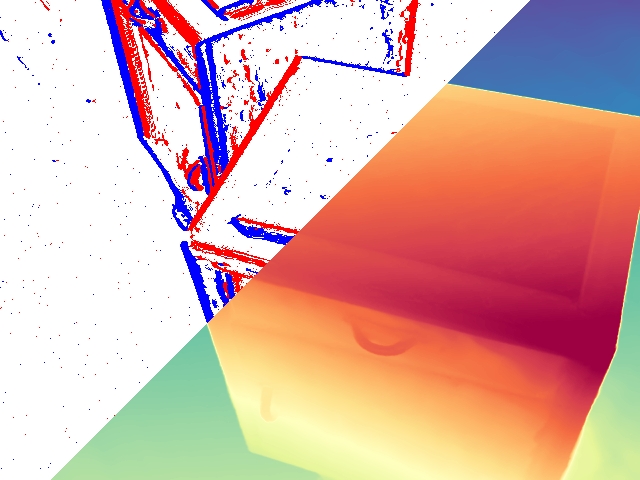} & 
         \includegraphics[width=0.20\linewidth]{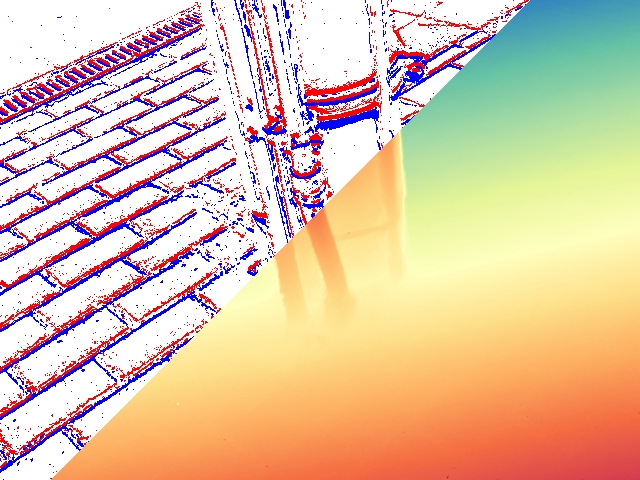} & 
         \includegraphics[width=0.20\linewidth]{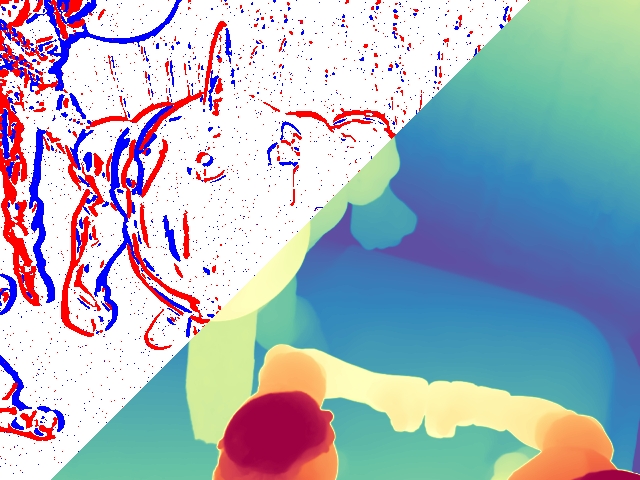} \\[-2pt]

         \rotatebox{90}{\makecell{\hspace{5pt}ScanNet++~\cite{yeshwanth2023scannet}}} &
         \includegraphics[width=0.20\linewidth]{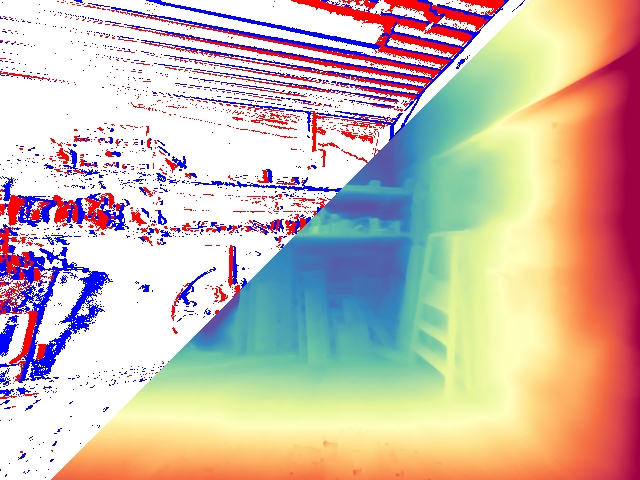} & 
         \includegraphics[width=0.20\linewidth]{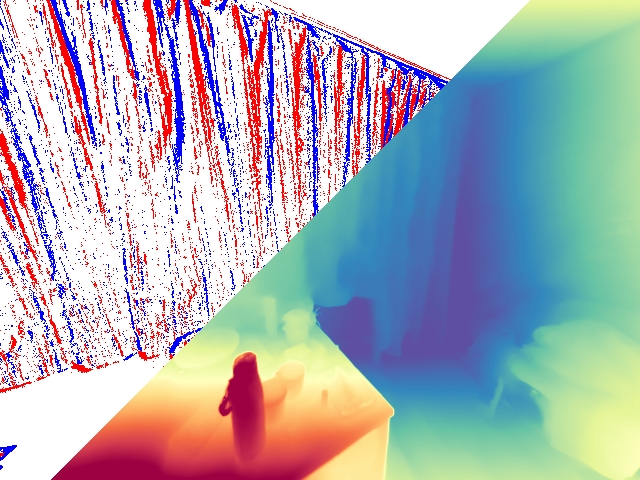} & 
         \includegraphics[width=0.20\linewidth]{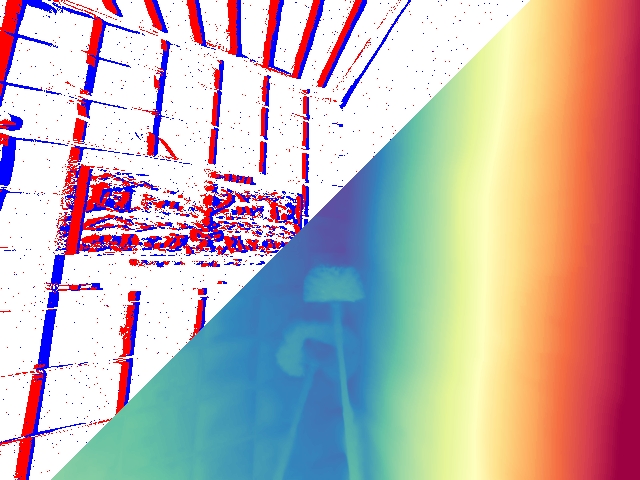} & 
         \includegraphics[width=0.20\linewidth]{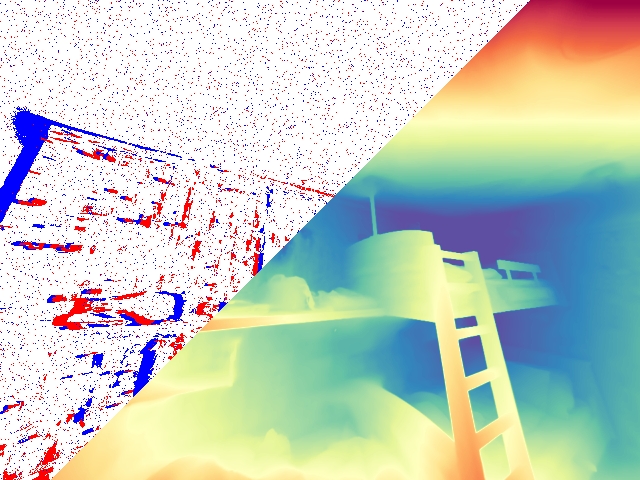} &

         \includegraphics[width=0.20\linewidth]{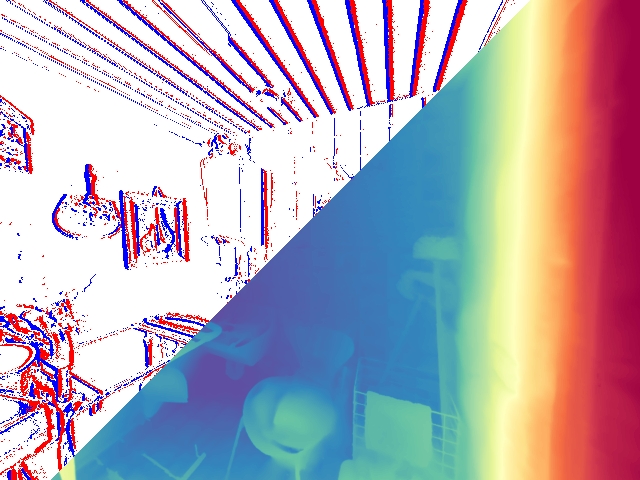} & 
         \includegraphics[width=0.20\linewidth]{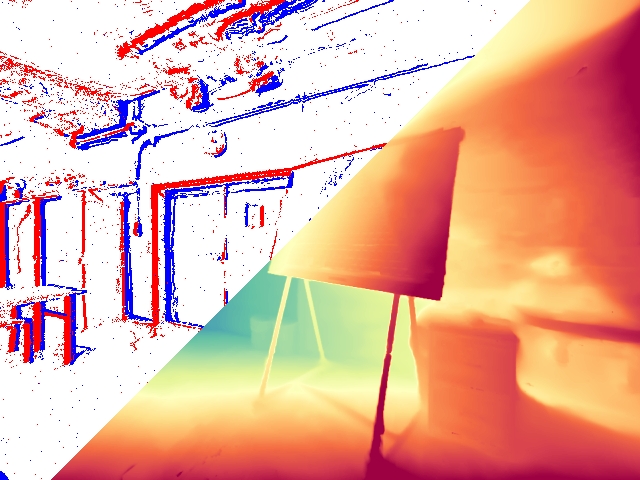} & 
         \includegraphics[width=0.20\linewidth]{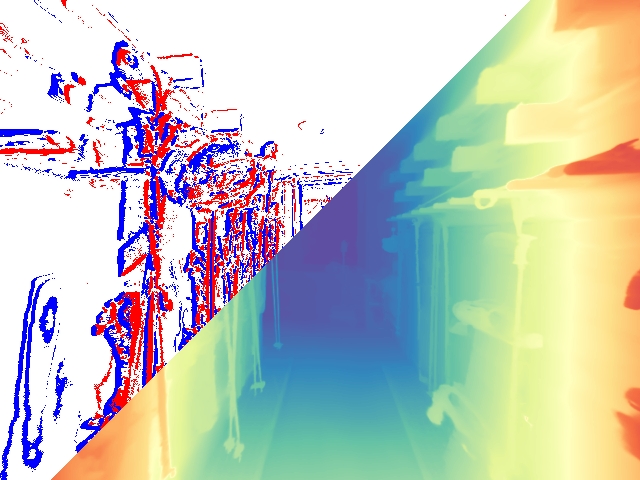} & 
         \includegraphics[width=0.20\linewidth]{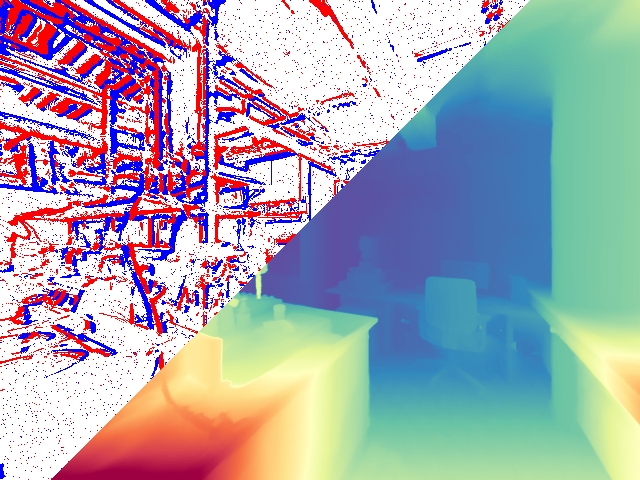} \\[-2pt]  

         \rotatebox{90}{\makecell{\hspace{16pt}DSEC~\cite{gehrig2021dsec}}} &
         \includegraphics[width=0.20\linewidth]{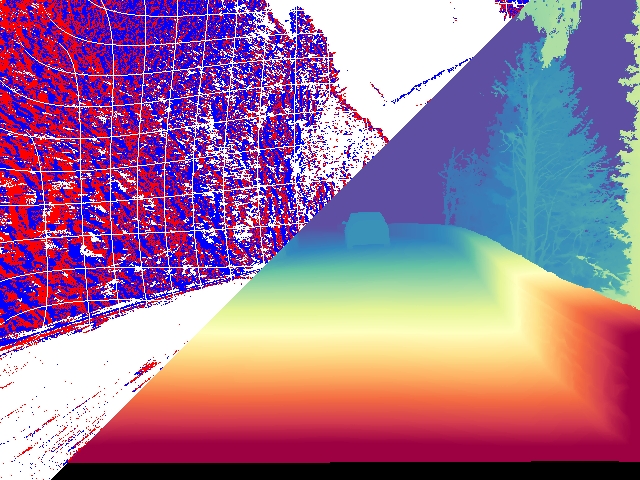} & 
         \includegraphics[width=0.20\linewidth]{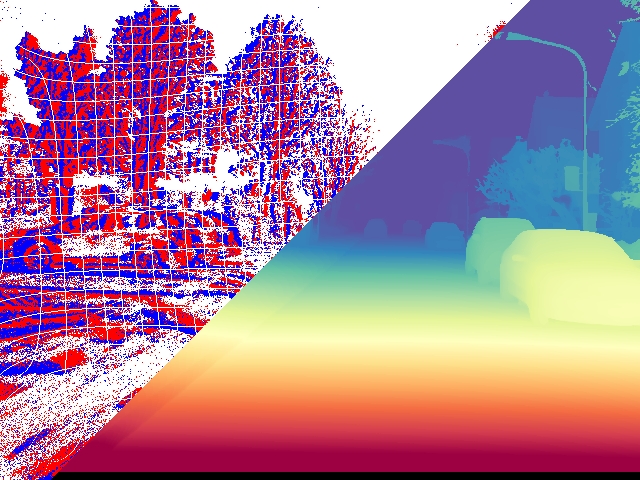} & 
         \includegraphics[width=0.20\linewidth]{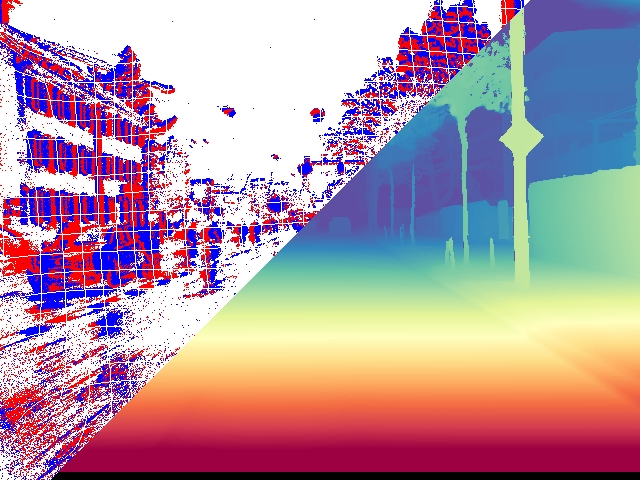} & 
         \includegraphics[width=0.20\linewidth]{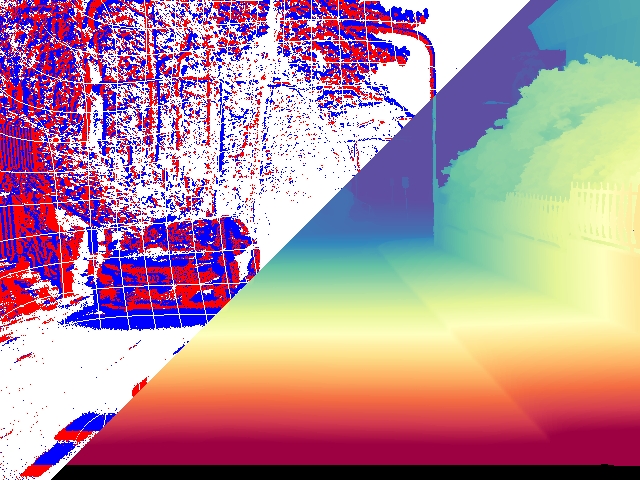} &

         \includegraphics[width=0.20\linewidth]{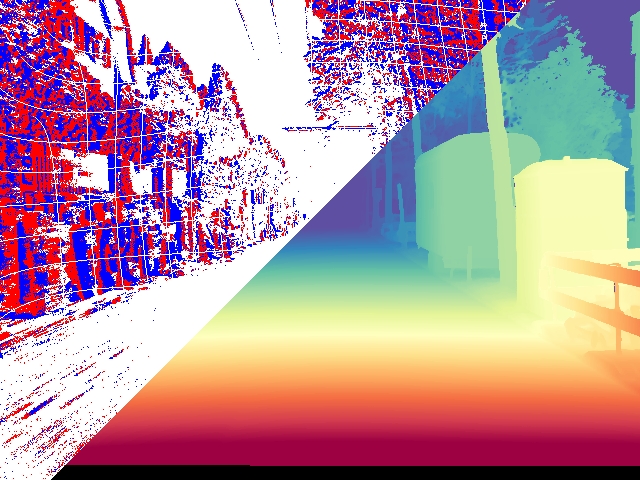} & 
         \includegraphics[width=0.20\linewidth]{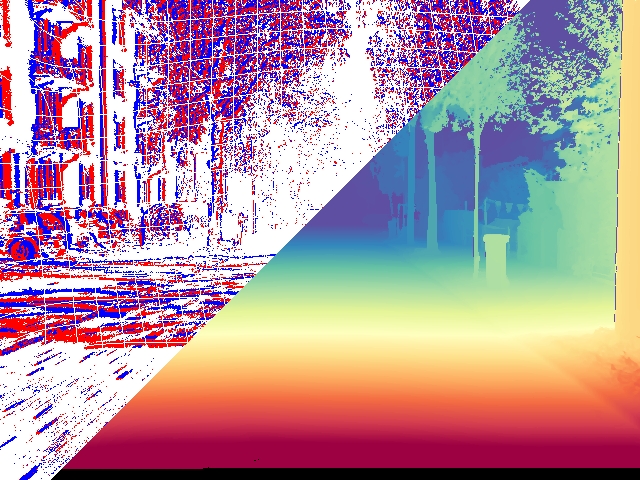} & 
         \includegraphics[width=0.20\linewidth]{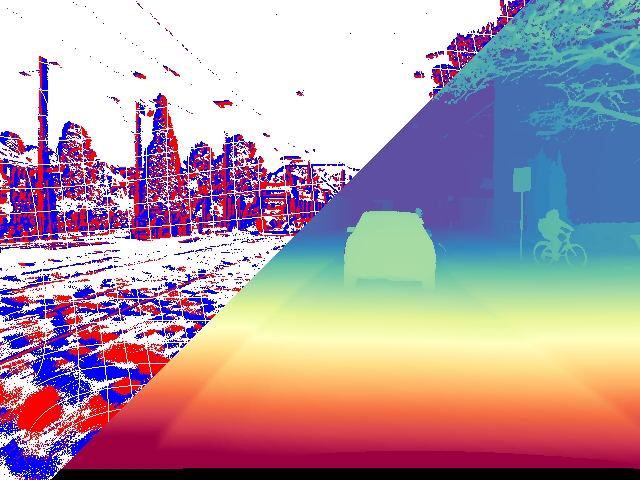} & 
         \includegraphics[width=0.20\linewidth]{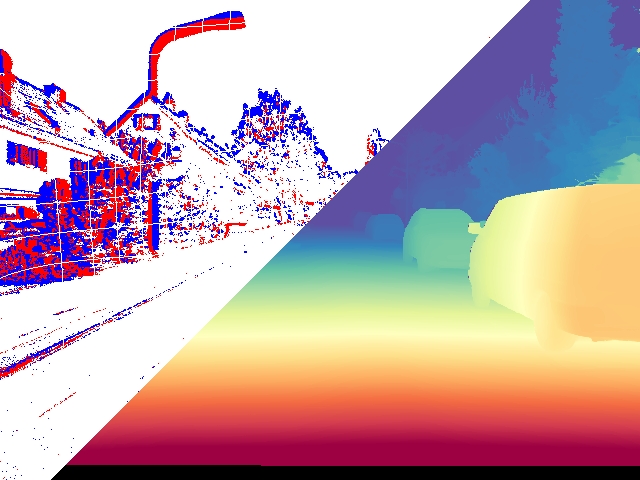} \\
         
    \end{tabular}
    }\vspace{-0.3cm}
    \caption{\textbf{Qualitative examples of events and proxy annotations by EventHub.} 
    From top to bottom, examples obtained from NeRF-Stereo \cite{tosi2023nerf}, ScanNet++ \cite{yeshwanth2023scannet} through novel view synthesis, and from DSEC \cite{gehrig2021dsec} through cross-modal distillation.}
    \label{fig:nsd_scannet_qualitives}\vspace{-0.3cm}
\end{figure*}

\sssubsection[sub:virtualtrajectory]{Virtual Trajectory Construction.}
Given that the captured scene is static and an event camera triggers events only when the logarithmic intensity changes exceed a threshold, we emulate such variations by simulating virtual camera egomotion.
We design two types of virtual trajectories: a local trajectory $\Gamma(\tau)$ and a global trajectory $\Omega(\tau)$.
Both $\Gamma(\tau)$ and $\Omega(\tau)$ are continuous functions mapping a virtual time instant $\tau\in[0,1]$ into a virtual pose $\mathbf{T}_\tau\in \mathbb{SE}(3)$. %
Given an initial camera pose $\hat{\mathbf{T}}_i$ (previously estimated using COLMAP), 
$\Gamma(\tau)=[\hat{\mathbf{R}}_i|\hat{\mathbf{t}}_i+\tau\mathbf{r}]$ applies a $\tau$-weighted translation $\mathbf{r}$ along an arbitrary axis, \eg, $\mathbf{r} = (0\ 1\ 0)^\top$.
This simple setup is well-suited for object-centric captured scenes~\cite{tosi2023nerf}, where the quality of novel views tends to degrade as the rendering pose moves farther from those observed during training.

Conversely, the global trajectory $\Omega(\tau)$ is obtained by performing a least-squares fit of three cubic splines to a subset (typically one-half or one-third) of the estimated camera poses, producing smooth and continuous motion.
Although a single cubic spline suffices to model the translation component $\mathbf{t}_\tau$, two additional splines followed by a re-orthogonalization ensure that $\mathbf{R}_\tau\in\mathbb{SO}(3)$.
This configuration enables the synthesis of complex trajectories involving large rotations. 
However, to maintain meaningful viewpoints (i.e., camera orientations directed toward observed scene regions), it is generally preferable to employ this type of trajectory for indoor recordings~\cite{yeshwanth2023scannet}.

\sssubsection[sub:trinocularrendering]{Motion-Adaptive Stereo Rendering.}
After defining a virtual stereo baseline $b$ and recovering the focal length $f$ from the intrinsic matrix $\mathbf{K}$ of the virtual camera, we render the disparity map $\mathbf{D}=(b\cdot f)/{\mathbf{Z}}$ used for the supervision of stereo networks.
Although depth regularization improves stability, the rendered depth maps still exhibit noise. %
To mitigate this, \cite{tosi2023nerf} proposed to extract trinocular images $\mathbf{I}_{LL}$, $\mathbf{I}_L$, $\mathbf{I}_R$ (where $\mathbf{I}_{LL}$ and $\mathbf{I}_R$ are rendered after applying respectively stereo translations $(b\ 0\ 0)^\top$ and $(-b\ 0\ 0)^\top$ to $\mathbf{t}_\tau$), balancing $\mathbf{D}$ with a trinocular photometric loss using Ambient Occlusion $\mathbf{C}_\text{AO}$ as the weighting term (more details in the supplementary material).
To improve confidence estimation, we design a voxel-based confidence measure:%
\begin{equation}
\textstyle
    \mathbf{C}_\text{Vsize} = \text{norm}\left(\sum_{i=1}^N T_i s_i\right) \odot \text{norm}\left(\sum_{i=1}^N T_i \alpha_i\right),
    \label{eq:cvsize}
\end{equation}
where $s_i$ denotes the size of the $i$-th voxel along the camera ray, $\odot$ represent the Hadamard product, and $\text{norm}(\cdot)$ normalizes the input to the range $[0,1]$.

Given that SVRaster~\cite{sun2025sparse} does not natively support event generation, we leverage ESIM~\cite{gehrig2020video} to simulate stereo events from rendered frames: given two consecutive virtual frames $\mathbf{I}_L(\tau)$ and $\mathbf{I}_L(\tau+\Delta \tau)$ (or $\mathbf{I}_R(\tau)$ and $\mathbf{I}_R(\tau+\Delta \tau)$ if $(-b\ 0\ 0)^\top$ is applied), ESIM simulates the event stream along the virtual motion, assuming $\Delta \tau$ being enough small.
Since both $\Gamma(\tau)$ and $\Omega(\tau)$ are continuous functions, we can render frames with arbitrary $\Delta \tau$, avoiding frame interpolation~\cite{gehrig2020video}; however, choosing $\Delta \tau$ is not trivial: too large values introduce simulation artifacts, while too small values leads to redundant computation.
Starting from a conservative value -- \eg, $\Delta \tau=\frac{1}{32}$ -- we dynamically adapt this value using pixel motion:
given the depth map $\mathbf{Z}$ and the camera poses $\mathbf{T}_\tau$ and $\mathbf{T}_{\tau+\Delta\tau}$, we compute the optical flow by projecting 3D points reconstructed from $\mathbf{Z}$ into the next view using the known relative motion $\mathbf{T}_{\tau \rightarrow \tau+\Delta\tau}$ and intrinsics $\mathbf{K}$. The flow field $\mathbf{F}$ is then obtained as the displacement between corresponding pixel projections across the two frames.
To ensure bounded pixel motion and prevent event artifacts, we set the number of intermediate renderings between $\tau$ and $\tau+\Delta\tau$ to $2^{n}$ with $n=\max(\lceil \log_2(|\mathbf{F}|_{\max}) \rceil, 0)$.

\begin{table}[t]
    \centering
    \resizebox{0.8\linewidth}{!}{
    \begin{tabular}{lcccc}
    \toprule
        \textbf{Dataset} & \textbf{MIX 1} & \textbf{MIX 2} & \textbf{MIX 3} & \textbf{MIX 4} \\
        \midrule
        NeRF-Stereo~\cite{tosi2023nerf} & \checkmark & \checkmark & & \checkmark \\
        ScanNet++~\cite{yeshwanth2023scannet} & & \checkmark & & \checkmark \\
        DSEC~\cite{gehrig2021dsec} & & & \checkmark & \checkmark \\
        \bottomrule
    \end{tabular}}\vspace{-0.3cm}
    \caption{\textbf{Combinations of datasets used by EventHub.} Proxy labels are applied to annotate each dataset.}\vspace{-0.3cm}
    \label{tab:mixes}
\end{table}

\begin{table*}[t]
\centering
\resizebox{1.0\textwidth}{!}{
\begin{tabular}{cl|cccc|cccc|cccc|cccc|c}
\toprule
& \multirow{2}{*}{\textbf{Training Method}} & \multicolumn{4}{c|}{\textbf{SE-CFF \cite{nam2022stereo}}} & \multicolumn{4}{c|}{\textbf{EMatch \cite{zhang2025ematch}}} & \multicolumn{4}{c|}{\textbf{E-FoundationStereo}} & \multicolumn{4}{c|}{\textbf{E-StereoAnywhere}} & \multirow{2}{*}{\textbf{Avg Rank}} \\
  & & \textbf{1PE} & \textbf{2PE} & \textbf{3PE} & \textbf{MAE} & \textbf{1PE} & \textbf{2PE} & \textbf{3PE} & \textbf{MAE} & \textbf{1PE} & \textbf{2PE} & \textbf{3PE} & \textbf{MAE} & \textbf{1PE} & \textbf{2PE} & \textbf{3PE} & \textbf{MAE} & \\
\midrule
(A) & Photometric \cite{tosi2023nerf} & 88.54 & 71.73 & 55.35 & 7.94 & 92.31 & 69.17 & 44.90 & 3.37 & 93.85 & 73.12 & 49.82 & 3.65 & 92.55 & 72.25 & 51.65 & 4.11 & 6.94 \\
(B) & EV-SceneFlow \cite{gehrig2020video,mayer2016large} & 66.30 & 50.18 & 41.47 & 3.50 & 71.64 & 53.67 & 42.02 & 3.56 & 61.80 & 48.04 & 41.68 & 3.10 & 64.97 & 49.54 & 41.74 & 3.21 & 6.06 \\
\midrule
\multirow{4}{*}{{(C)}} & MIX 1 & 45.61 & 25.15 & 16.54 & 1.87 & 47.17 & 23.23 & 13.58 & 1.70 & 38.70 & 15.63 & 8.83 & 1.39 & 35.86 & 15.13 & 9.03 & 1.36 & 5.00 \\
& MIX 2 & 38.52 & 18.17 & 11.02 & 1.56 & 41.06 & 17.81 & 9.97 & 1.45 & 27.20 & 9.72 & 5.53 & 1.04 & 31.49 & 12.45 & 7.32 & 1.23 & 4.00 \\
& MIX 3 & \snd{24.73} & \snd{8.58} & \snd{5.08} & \snd{1.01} & 31.30 & 11.14 & 5.90 & 1.15 & 20.99 & 6.82 & 4.10 & 0.89 & 24.35 & 8.22 & 4.76 & 0.99 & 2.75 \\
& MIX 4 & 27.31 & 9.69 & 5.66 & 1.07 & \snd{26.13} & \snd{8.55} & \snd{4.71} & \snd{0.99} & \snd{20.42} & \snd{6.53} & \snd{3.91} & \snd{0.87} & \snd{23.90} & \snd{7.97} & \snd{4.62} & \snd{0.96} & \snd 2.25 \\
\midrule
(D) & LiDAR (GT) & \fst{13.82} & \fst{4.05} & \fst{2.37} & \fst{0.66} & \fst{24.11} & \fst{7.80} & \fst{3.99} & \fst{0.89} & \fst{12.53} & \fst{3.48} & \fst{1.98} & \fst{0.60} & \fst{14.66} & \fst{4.32} & \fst{2.51} & \fst{0.69} & \fst 1.00 \\
\bottomrule
\end{tabular}
}\vspace{-0.3cm}
\caption{\textbf{In-domain experimental results -- DSEC \cite{gehrig2021dsec} dataset.} We train four event stereo models according to different training protocols, not exploiting LiDAR annotations (A,B,C), compared against in-domain training on DSEC with LiDAR labels (D). }\vspace{-0.2cm}
\label{tab:exp1_a}
\end{table*}

\subsubsection{Stereo Cross-Modal Distillation}
\label{sec:method:distillation}

In the second setting, we assume the availability of data collected in the same environment by two calibrated sensors: RGB and event cameras. 
In this case, we can obtain proxy labels from color images and transfer them to the events domain by exploiting multi-view geometry, if needed. 

More specifically, as we focus on the event stereo matching task, we assume the availability of paired color-event stereo pairs $(\mathbf{I}_L,\mathbf{I}_R)-(\mathbf{E}_L,\mathbf{E}_R)$, often true for the most popular event stereo datasets available in the literature \cite{zhu2018multivehicle,gehrig2021dsec,chaney2023m3ed}. 
Accordingly, we can use an off-the-shelf, state-of-the-art Stereo Foundation Model (SFM) \cite{bartolomei2025stereo,wen2025foundationstereo} $\Phi_c$ to predict proxy labels by processing a color stereo pair $(\mathbf{I}_L, \mathbf{I}_R)$. 
Then, by knowing the relative pose $\mathbf{T}_{c\rightarrow e}$ between the left color camera and the left event one, we can transfer the labels and annotate the event data.

Specifically, the disparity map predicted by $\Phi_c$ is converted into depth $\mathbf{Z}_c$ by knowing baseline $b_c$ and focal length $f_c$ of the color stereo camera
\begin{equation}
    \mathbf{Z}_c = %
    (b_c \cdot f_c) / \mathbf{D}_c,
    \quad\quad \text{with} \quad \mathbf{D}_c=\Phi_c(\mathbf{I}_L, \mathbf{I}_R).
\end{equation}
Then, being $\mathbf{u}_c$ a pixel in homogeneous coordinates on the color camera frame, we back-project it into a 3D point $\mathbf{p}_c$ according to depth $\mathbf{Z}_c(\mathbf{u}_c)$ and intrinsics $\mathbf{K}_c$. $\mathbf{p}_c$ is then expressed in the event camera reference system by applying the transformation $\mathbf{T}_{c\rightarrow e}$ between both, obtaining
\begin{equation}
    \mathbf{p}_e = {\mathbf{T}}_{c\rightarrow e} \mathbf{p}_c, \quad\quad \text{with}\quad \mathbf{p}_c = \mathbf{Z}_c(\mathbf{u}_c)\mathbf{K}_c^{-1} \mathbf{u}_c.
\end{equation}

Finally, we project the $z$ coordinate of $\mathbf{p}_e$ into the event camera frame according to intrinsics $\mathbf{K}_e$. Doing this for any pixel in $\mathbf{I}_L$ yields a depth map $\mathbf{Z}_e$ aligned with $\mathbf{E}_L$. Finally, we obtain the disparity map $\mathbf{D}_e$ through triangulation, thus obtaining proxy labels $\mathbf{D}_e={(b_e \cdot f_e)}/{\mathbf{Z}_e}$ for the event stereo pair $(\mathbf{E}_L,\mathbf{E}_R)$. 
In this way, we can distill the knowledge of state-of-the-art stereo models \cite{bartolomei2025stereo,wen2025foundationstereo} and reuse it in the events domain to pursue the same advances achieved in RGB stereo.
This procedure is not needed if a sensor such as the DAVIS camera is available, providing pixel aligned grayscale images and event streams \cite{zhu2018multivehicle}: 
in such a case, the initial disparity map $\mathbf{D}_c$ %
already coincides with $\mathbf{D}_e$.

\subsection{Repurposing RGB Stereo into Event Stereo}
\label{sec:method:adapting_vfm}

Besides exploiting RGB stereo models to distill proxy annotations for the event domain, we further benefit from the vast priors they learned from the abundant RGB stereo data available by repurposing the models themselves into event stereo models.
In other words, we design and train an event stereo model $\Phi_e$ having the same architecture as an RGB stereo one, starting from pre-trained weights $\Phi_c$ (i.e., those used to distill proxy labels).

To this aim, keeping the number of input channels unchanged across RGB and event stereo would avoid any modification to the original deep neural network model: 
purposely, we encode event streams into stacked tensors according to the 3-channel Tencode representation \cite{huang2023eventpoint} that are passed as inputs to $\Phi_e$, by sampling events backward in time based on a fixed number of events: %
\begin{equation}
    (x,y,t,p) \!\rightarrow\!
    \mathbf{S}(x,y) =
    \begin{cases}
        (1,\,\tfrac{t_{\max}-t}{\Delta t},\,0), & p>0\\
        (0,\,\tfrac{t_{\max}-t}{\Delta t},\,1), & p\le0,
    \end{cases}
\end{equation}
with $t_{\max}$ being the timestamp of the latest event occurred in the timelapse $\Delta t$ during which events are stacked. 

\section{Experiments}
\label{chapter:experiments}

\subsection{Implementation and Experimental Settings}
\label{subsec:implementation_details}

\textbf{EventHub Settings.}
We collect proxy data from multiple sources~\cite{tosi2023nerf,yeshwanth2023scannet,gehrig2021dsec}.
For datasets without ready-to-use events~\cite{tosi2023nerf,yeshwanth2023scannet}, we employ our NVS pipeline to synthesize both proxy events and depth, while for \cite{gehrig2021dsec} we apply our distillation pipeline to estimate proxy depth only.
Each NVS scene is optimized independently, setting $\lambda_\text{SSIM}=0.02$, $\lambda_{{N-\text{mean}}}=\lambda_{N-\text{med}}=0.0005$, $\lambda_{\text{DAv2}}=0.01$ and disabling all other regularizers. %
We use three local trajectories $\Gamma_x,\Gamma_y,\Gamma_z$ (one per axis) for \cite{tosi2023nerf}, and a global trajectory $\Omega$ for \cite{yeshwanth2023scannet} with additional processing described in the supplementary material.

For NeRF-Stereo~\cite{tosi2023nerf}, we render the 270 scenes three times (one for each baseline $b\in\{0.1,0.3,0.5\}$) at $640\times480$ px resolution, setting $\Delta\tau=0.03$.
For our 403 scenes selection of ScanNet++~\cite{yeshwanth2023scannet}, we render both $640\times480$ px and $1280\times720$ px resolutions, each with three baselines $b\in\{0.05,0.08,0.1\}$, setting $\Delta\tau=0.015$.
To filter noisy labels, we adopt curation pipeline \cite{wen2025foundationstereo}, training SE-CFF~\cite{nam2022stereo} paired with Tencode~\cite{huang2023eventpoint} on all the NVS data, discarding samples with excessive pixel errors, yielding $\sim70$k curated pairs.
For DSEC~\cite{gehrig2021dsec} proxy labeling, we retain the train split of \cite{bartolomei2024lidar} (excluding night sequences), generating proxy labels via FoundationStereo~\cite{wen2025foundationstereo} ViT-L, clipping depth between $[0.5,100]$m before reprojection, obtaining a total of $\sim30$k samples. 
Figure \ref{fig:nsd_scannet_qualitives} shows some annotated examples generated from the three datasets by EventHub,  
while \Cref{tab:mixes} reports different mixtures of training data derived from them and used in our experiments.

\textbf{Stereo Models and Training Settings.} 
We evaluate two event-based stereo networks~\cite{nam2022stereo,zhang2025ematch}, using Tencode~\cite{huang2023eventpoint} and VoxelGrid~\cite{zhu2019unsupervised} event representations, respectively, and two RGB-based models~\cite{wen2025foundationstereo,bartolomei2025stereo} adapted through our repurposing strategy and dubbed E-FoundationStereo and E-StereoAnywhere, respectively.
Event networks are trained from scratch with a learning rate (lr) of $5\cdot10^{-4}$, while RGB models are fine-tuned from the authors’ ViT-S checkpoints using $\text{lr}=5\cdot10^{-5}$ and freezing the DAv2-S prior only.
Training is performed in PyTorch with the AdamW optimizer, OneCycle learning rate scheduler, and data augmentations including random cropping at $576\times448$ px.
All models are trained for 10 epochs on a single A100 GPU with batch size 2.
On NVS data~\cite{tosi2023nerf,yeshwanth2023scannet} we use the NeRF-supervised loss \cite{tosi2023nerf}, while on distilled data \cite{gehrig2021dsec} and non-EventHub data
we use the original loss of each model. %
These settings are used for all experiments. %

\begin{figure}[t]
\centering
\renewcommand{\tabcolsep}{1pt}
\resizebox{0.45\textwidth}{!} {
	\begin{tabular}{ccc}
	\begin{overpic}[width=0.180\textwidth]{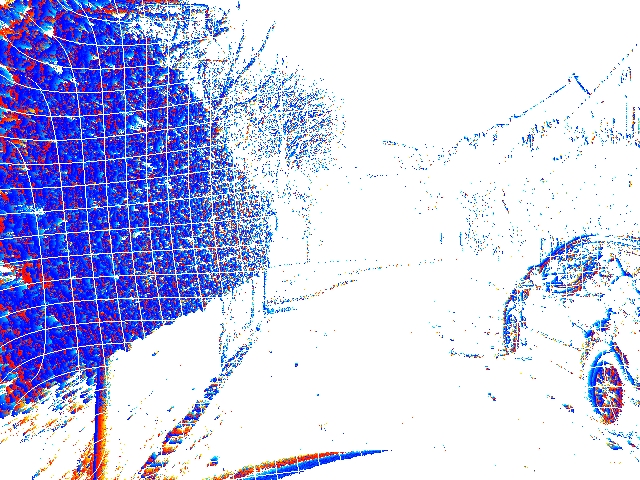}
    \setlength{\fboxsep}{1pt}
    \setlength{\fboxrule}{1pt}
    \put(30,66.5){\colorbox{white}{\small\textbf{Events}}}
    \end{overpic} &
    \begin{overpic}[width=0.180\textwidth]{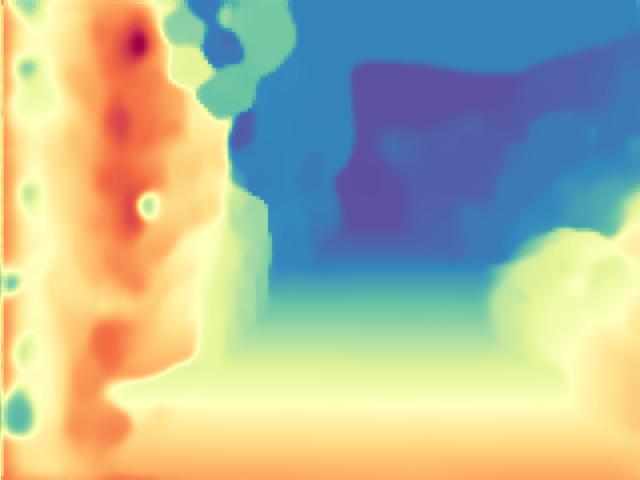} 
    \setlength{\fboxsep}{1pt}
    \setlength{\fboxrule}{1pt}
    \put(20,66.5){\colorbox{white}{\small\textbf{Photometric}}}
    \end{overpic} &
    \begin{overpic}[width=0.180\textwidth]{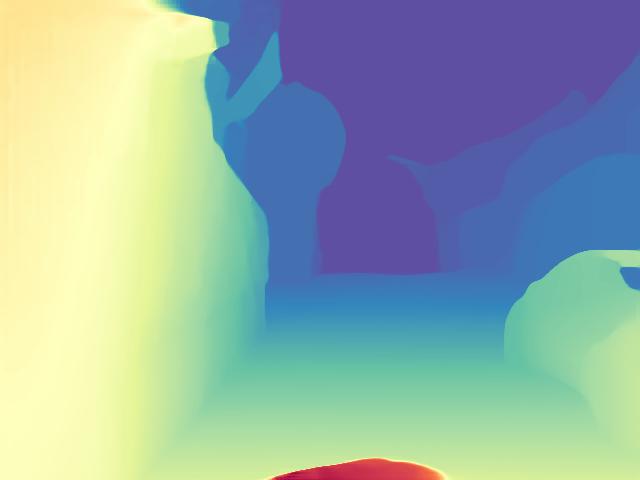}
    \setlength{\fboxsep}{1pt}
    \setlength{\fboxrule}{1pt}
    \put(15,66.5){\colorbox{white}{\small\textbf{EV-SceneFlow}}}
    \end{overpic}\\[-1pt]
    \begin{overpic}[width=0.180\textwidth]{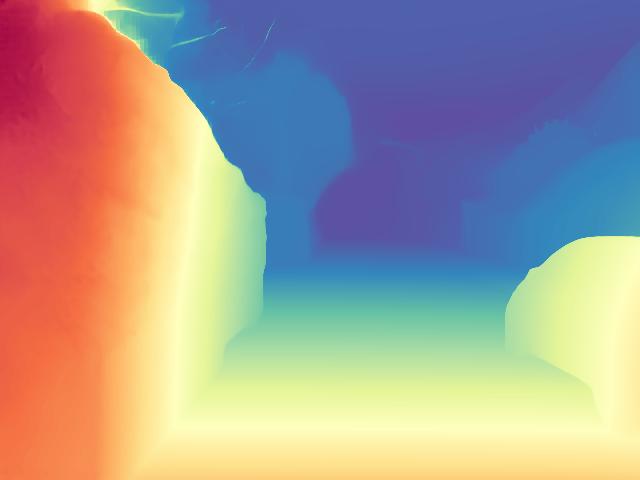}
    \setlength{\fboxsep}{1pt}
    \setlength{\fboxrule}{1pt}
    \put(17,66){\colorbox{white}{\small\textbf{MIX~3 (ours)}}}
    \end{overpic} &
    \begin{overpic}[width=0.180\textwidth]{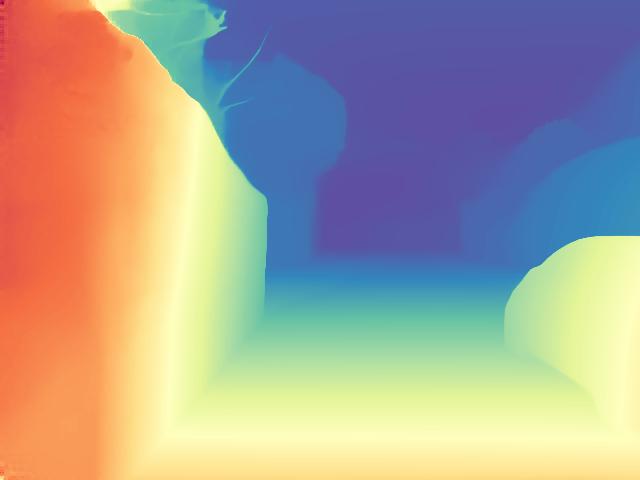}
    \setlength{\fboxsep}{1pt}
    \setlength{\fboxrule}{1pt}
    \put(17,66){\colorbox{white}{\small\textbf{MIX~4 (ours)}}}
    \end{overpic} &
    \begin{overpic}[width=0.180\textwidth]{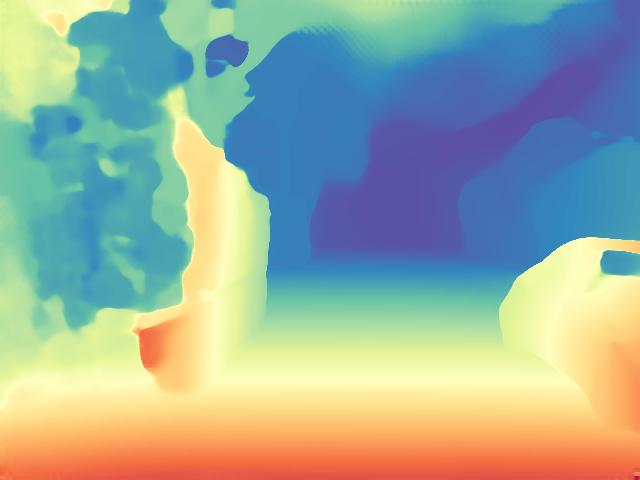}
    \setlength{\fboxsep}{1pt}
    \setlength{\fboxrule}{1pt}
    \put(20,66){\colorbox{white}{\small\textbf{LiDAR (GT)}}}
    \end{overpic}\\
	\end{tabular}
}\vspace{-0.3cm}
\caption{\textbf{Qualitative results on DSEC dataset \cite{gehrig2021dsec}.} Predictions by E-FoundationStereo trained according to different protocols.}\vspace{-0.3cm}
\label{fig:qualitative_dsec_zurich_city_06_a_00005}
\end{figure}

\begin{table*}[t]
\centering
\renewcommand{\tabcolsep}{8pt}
\resizebox{1.0\textwidth}{!}{
\begin{tabular}{l|l|cccc|cccc|cccc|c}
\toprule
\multirow{2}{*}{\textbf{Model}} & \multirow{2}{*}{\textbf{Training Method}} & \multicolumn{4}{c|}{\textbf{M3ED (Day)}} & \multicolumn{4}{c|}{\textbf{M3ED (Night)}} & \multicolumn{4}{c|}{\textbf{M3ED (Indoor)}} & \multirow{2}{*}{\textbf{Avg Rank}} \\
 &  & \textbf{1PE} & \textbf{2PE} & \textbf{3PE} & \textbf{MAE} & \textbf{1PE} & \textbf{2PE} & \textbf{3PE} & \textbf{MAE} & \textbf{1PE} & \textbf{2PE} & \textbf{3PE} & \textbf{MAE} & \\
\midrule
\multirow{3}{*}{SE-CFF \cite{nam2022stereo}} & MIX~3 & \snd{46.84} & \snd{25.31} & \snd{16.99} & \snd{2.81} & \snd{58.32} & \snd{36.07} & 24.45 & 3.50 & 51.03 & 31.43 & 23.84 & 4.52 & 2.50 \\
& MIX~4 & \fst{35.65} & \fst{15.18} & \fst{9.37} & \fst{1.22} & \fst{51.57} & \fst{26.84} & \fst{15.33} & \fst{1.70} & \snd{48.56} & \fst{27.36} & \fst{19.33} & \fst{2.95} & \fst 1.08 \\
& LiDAR (GT) & 58.82 & 41.41 & 32.93 & 3.05 & 58.94 & 36.78 & \snd{23.55} & \snd{2.20} & \fst{45.33} & \snd{28.39} & \snd{21.59} & \snd{4.48} & \snd 2.42 \\
\midrule
\multirow{3}{*}{EMatch \cite{zhang2025ematch}} & MIX~3 & 86.18 & 77.61 & 72.13 & 40.36 & 90.02 & 84.94 & 82.26 & 45.41 & 76.99 & 65.23 & 58.98 & 15.80 & 3.00 \\
& MIX~4 & \fst{43.99} & \fst{20.87} & \fst{12.93} & \fst{2.23} & \fst{63.69} & \fst{38.74} & \fst{26.38} & \fst{5.03} & \fst{58.42} & \fst{32.16} & \fst{21.48} & \fst{3.10} & \fst 1.00 \\
& LiDAR (GT) & \snd{83.16} & \snd{71.65} & \snd{62.81} & \snd{12.22} & \snd{83.36} & \snd{73.03} & \snd{66.06} & \snd{18.63} & \snd{64.34} & \snd{46.85} & \snd{38.80} & \snd{7.95} & \snd 2.00 \\
\midrule
\multirow{3}{*}{E-FoundationStereo} & MIX~3 & \snd{33.44} & \snd{19.20} & \snd{12.37} & \snd{1.49} & \snd{49.26} & \snd{26.09} & \snd{14.94} & \snd{1.84} & \snd{40.27} & 22.08 & \snd{15.73} & \fst{2.37} & \snd 2.00 \\
& MIX~4 & \fst{26.38} & \fst{11.57} & \fst{6.96} & \fst{0.98} & \fst{46.90} & \fst{23.09} & \fst{12.96} & \fst{1.54} & 40.74 & \fst{21.83} & \fst{15.61} & \snd{2.45} & \fst 1.25 \\
& LiDAR (GT) & 54.80 & 39.48 & 31.43 & 2.89 & 55.77 & 34.93 & 22.60 & 1.99 & \fst{38.93} & \snd{22.03} & 15.96 & 2.87 & 2.75 \\
\midrule
\multirow{3}{*}{E-StereoAnywhere} & MIX~3 & \snd{47.53} & \snd{27.85} & \snd{18.83} & 4.48 & \snd{59.21} & \snd{34.29} & \snd{21.21} & 3.64 & 46.02 & 26.71 & 19.55 & \snd{3.19} & \snd 2.42 \\
& MIX~4 & \fst{34.99} & \fst{13.01} & \fst{7.88} & \fst{1.12} & \fst{58.62} & \fst{27.85} & \fst{14.42} & \fst{1.71} & \fst{42.74} & \fst{23.79} & \fst{16.84} & \fst{2.58} & \fst 1.00 \\
& LiDAR (GT) & 63.70 & 43.33 & 33.90 & \snd{3.26} & 63.23 & 41.22 & 26.72 & \snd{2.78} & \snd{44.27} & \snd{25.81} & \snd{18.80} & 3.72 & 2.58 \\
\bottomrule
\end{tabular}
}\vspace{-0.3cm}
\caption{\textbf{Out-of-domain experimental results -- M3ED \cite{chaney2023m3ed} dataset.} 
We compare the generalization capability of the four event stereo models trained with MIX~3 and MIX~4 against their counterparts trained using DSEC LiDAR labels.}
\label{tab:exp2_a_s}
\end{table*}

\begin{table*}
\centering
\resizebox{1.0\textwidth}{!}{
\begin{tabular}{l|l|cccc|cccc|cccc|c}
\toprule
\multirow{2}{*}{\textbf{Model}} & \multirow{2}{*}{\textbf{Training Method}} & \multicolumn{4}{c|}{\textbf{MVSEC (Day)}} & \multicolumn{4}{c|}{\textbf{MVSEC (Night)}} & \multicolumn{4}{c|}{\textbf{MVSEC (Indoor)}} & \multirow{2}{*}{\textbf{Avg Rank}} \\
 &  & \textbf{1PE} & \textbf{2PE} & \textbf{3PE} & \textbf{MAE} & \textbf{1PE} & \textbf{2PE} & \textbf{3PE} & \textbf{MAE} & \textbf{1PE} & \textbf{2PE} & \textbf{3PE} & \textbf{MAE} & \\
\midrule
\multirow{3}{*}{SE-CFF \cite{nam2022stereo}} & MIX~3 & \snd{77.13} & \snd{53.87} & \snd{31.12} & \snd{3.21} & \snd{78.19} & \snd{54.75} & \snd{31.24} & \snd{3.64} & \snd{42.14} & \snd{24.53} & 18.04 & 3.30 & \snd 2.17 \\
 & MIX~4 & \fst{31.99} & \fst{12.00} & \fst{6.88} & \fst{1.11} & \fst{40.54} & \fst{19.00} & \fst{10.13} & \fst{1.45} & \fst{29.66} & \fst{12.34} & \fst{6.89} & \fst{1.39} & \fst 1.00 \\
 & LiDAR (GT) & 97.82 & 94.85 & 90.47 & 6.12 & 96.89 & 92.95 & 87.14 & 6.07 & 46.67 & 26.37 & \snd{15.43} & \snd{1.78} & 2.83 \\
\midrule
\multirow{3}{*}{EMatch \cite{zhang2025ematch}} & MIX~3 & \snd{93.80} & \snd{80.23} & \snd{59.05} & \snd{6.00} & \snd{91.74} & \snd{75.37} & \snd{49.51} & \snd{4.67} & \snd{58.44} & \snd{35.05} & \snd{24.24} & 3.02 & \snd 2.08 \\
 & MIX~4 & \fst{56.29} & \fst{21.61} & \fst{6.67} & \fst{1.39} & \fst{68.51} & \fst{40.48} & \fst{14.92} & \fst{1.81} & \fst{46.03} & \fst{21.40} & \fst{12.36} & \fst{1.93} & \fst 1.00 \\
 & LiDAR (GT) & 99.47 & 98.36 & 95.83 & 6.70 & 98.56 & 96.20 & 92.40 & 6.21 & 66.21 & 43.21 & 28.13 & \snd{2.60} & 2.92 \\
\midrule
\multirow{3}{*}{E-FoundationStereo} & MIX~3 & \snd{81.78} & \snd{54.75} & \snd{28.95} & \snd{2.75} & \snd{81.89} & \snd{58.54} & \snd{36.27} & \snd{2.73} & \snd{34.45} & \snd{18.51} & \snd{12.48} & 1.62 & \snd 2.08 \\
 & MIX~4 & \fst{45.94} & \fst{20.92} & \fst{9.45} & \fst{1.33} & \fst{58.15} & \fst{38.14} & \fst{18.02} & \fst{1.75} & \fst{24.55} & \fst{9.11} & \fst{5.29} & \fst{1.07} & \fst 1.00 \\
 & LiDAR (GT) & 97.91 & 94.65 & 89.45 & 6.04 & 97.32 & 94.15 & 89.74 & 6.26 & 40.19 & 21.27 & 12.62 & \snd{1.61} & 2.92 \\
\midrule
\multirow{3}{*}{E-StereoAnywhere} & MIX~3 & \snd{77.22} & \snd{60.04} & \snd{42.41} & \snd{4.37} & \snd{75.97} & \snd{57.26} & \snd{35.88} & \snd{3.47} & \snd{40.68} & \snd{24.74} & \snd{19.64} & 4.18 & \snd 2.08 \\
 & MIX~4 & \fst{68.18} & \fst{44.60} & \fst{20.92} & \fst{1.96} & \fst{72.21} & \fst{47.93} & \fst{20.39} & \fst{1.96} & \fst{21.27} & \fst{7.84} & \fst{4.48} & \fst{0.94} & \fst 1.00 \\
 & LiDAR (GT) & 98.39 & 95.68 & 90.43 & 7.85 & 97.59 & 94.77 & 89.27 & 8.12 & 49.80 & 31.71 & 21.39 & \snd{2.79} & 2.92 \\
\bottomrule
\end{tabular}
}\vspace{-0.3cm}
\caption{\textbf{Out-of-domain experimental results -- MVSEC \cite{zhu2018multivehicle} dataset.} 
We compare the generalization capability of the four event stereo models trained with MIX~3 and MIX~4 against their counterparts trained using DSEC LiDAR labels.}
\label{tab:exp3_a_s}
\end{table*}

\begin{figure*}[t]
\centering
\renewcommand{\tabcolsep}{1pt}
\scalebox{0.8} {
	\hspace{-10pt}\begin{tabular}{cccccccc}
	& \scriptsize\textbf{Events \& Ground Truth} & \;\;&  & \scriptsize\textbf{SE-CFF \cite{nam2022stereo}} & \scriptsize\textbf{EMatch \cite{zhang2025ematch}} & \scriptsize\textbf{E-StereoAnywhere} & \scriptsize\textbf{E-FoundationStereo} \\
	\multirow{2}{*}{ \rotatebox[origin=l]{90}{\hspace{1.5em}\centering\scriptsize\textbf{M3ED \cite{chaney2023m3ed}}}} & \includegraphics[width=0.225\textwidth]{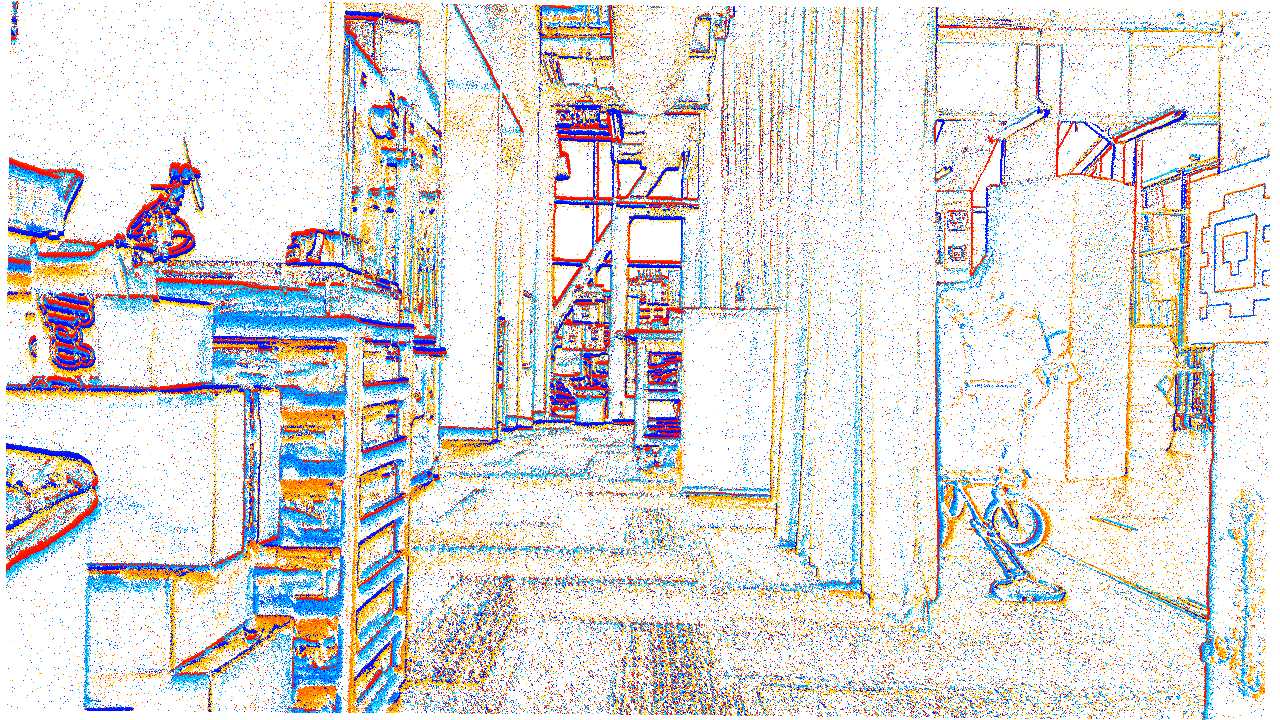} & \;\;& \rotatebox[origin=l]{90}{\hspace{1.25em}\centering\scriptsize\textbf{LiDAR (GT)}} & \includegraphics[width=0.225\textwidth]{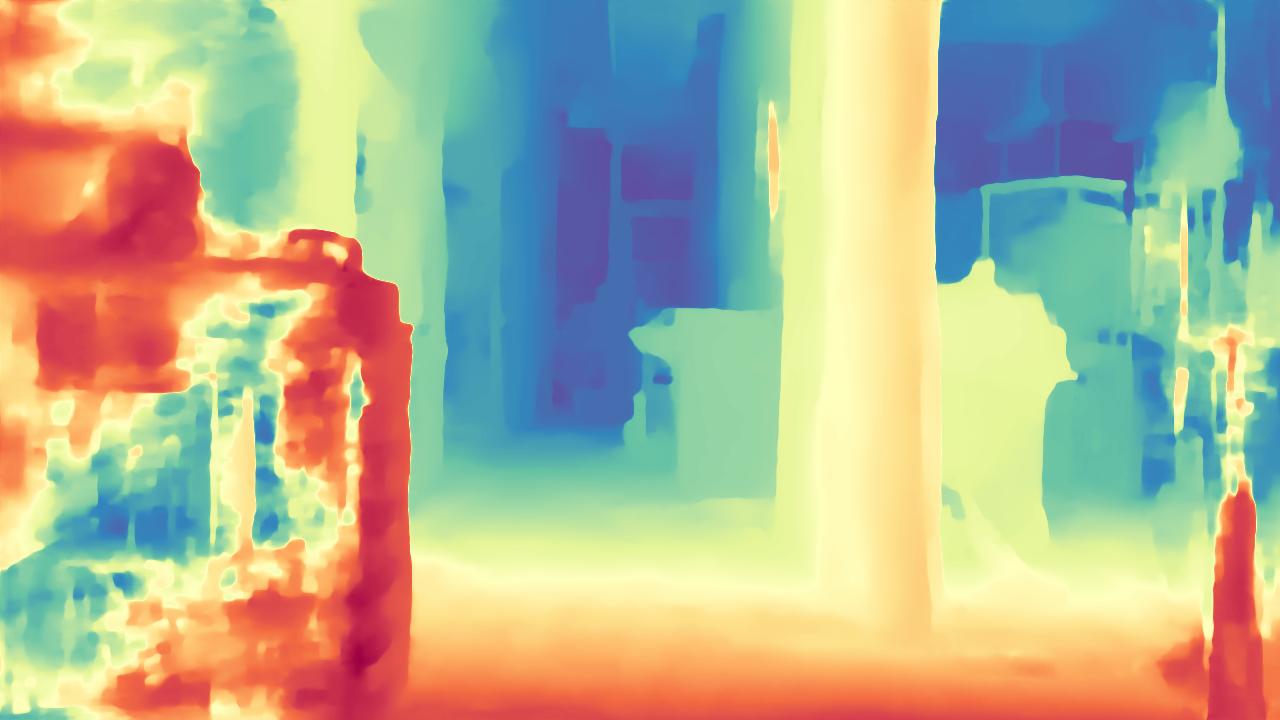} & \includegraphics[width=0.225\textwidth]{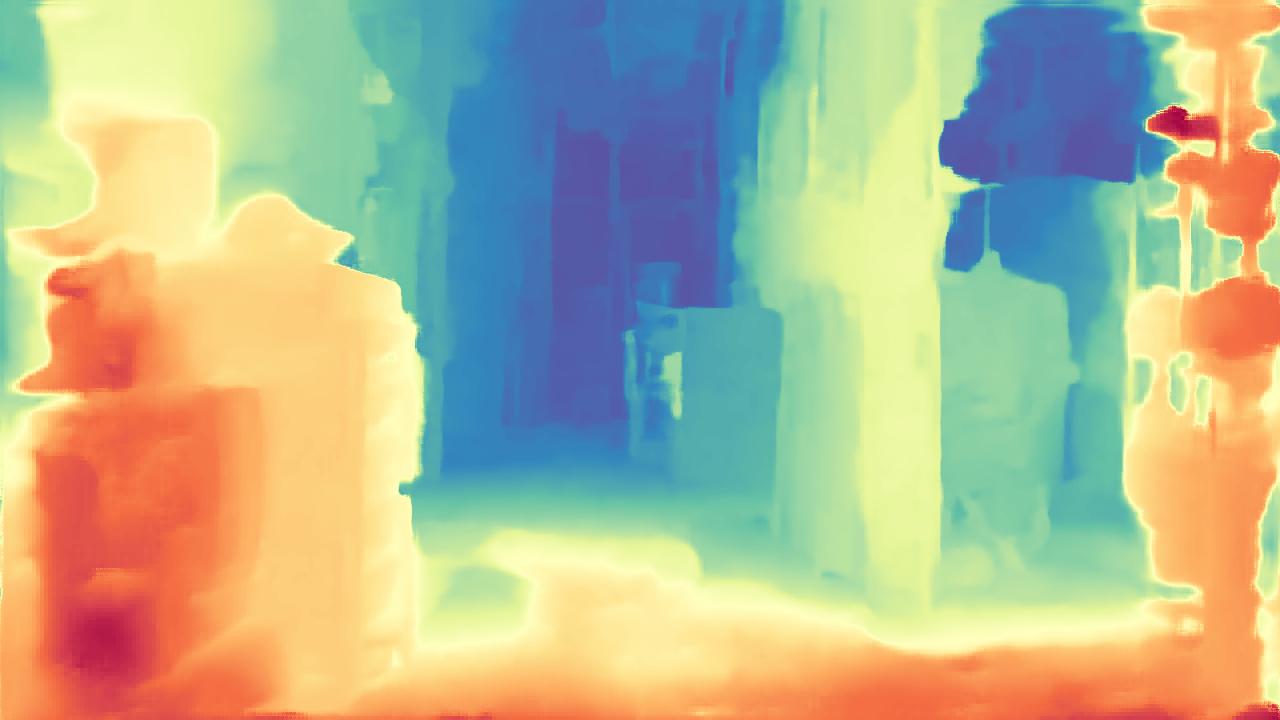} & \includegraphics[width=0.225\textwidth]{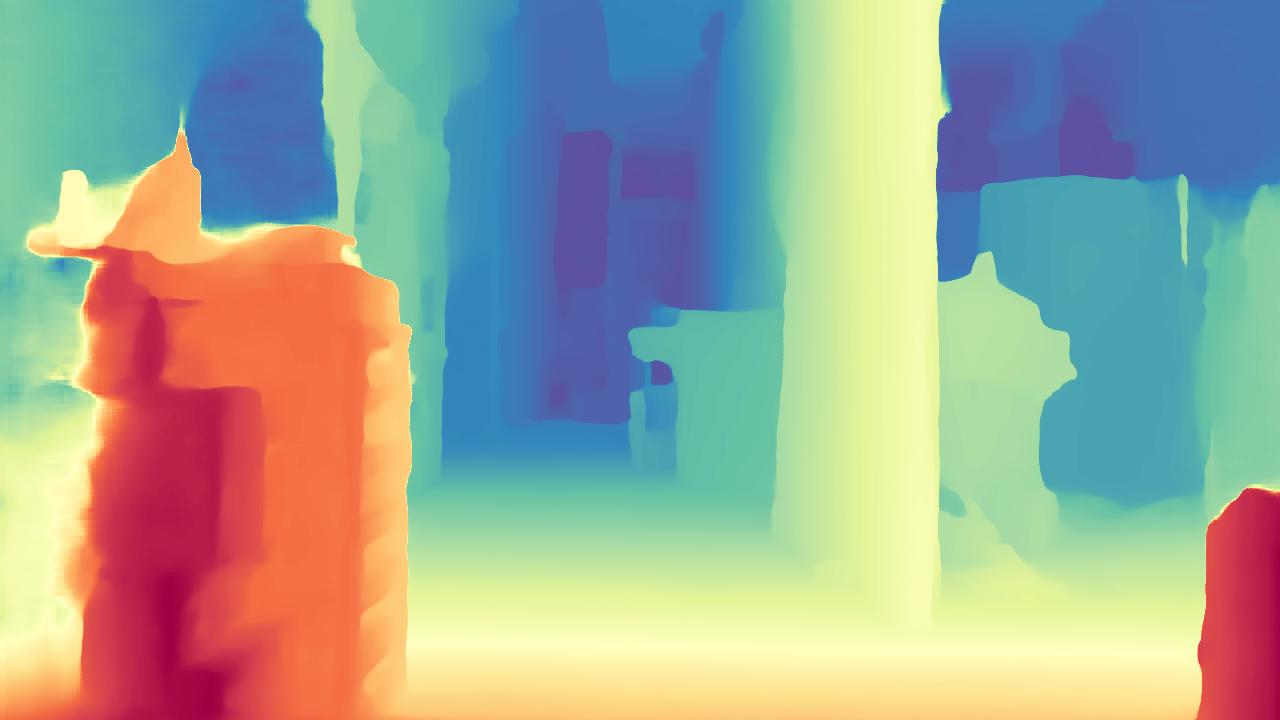} & \includegraphics[width=0.225\textwidth]{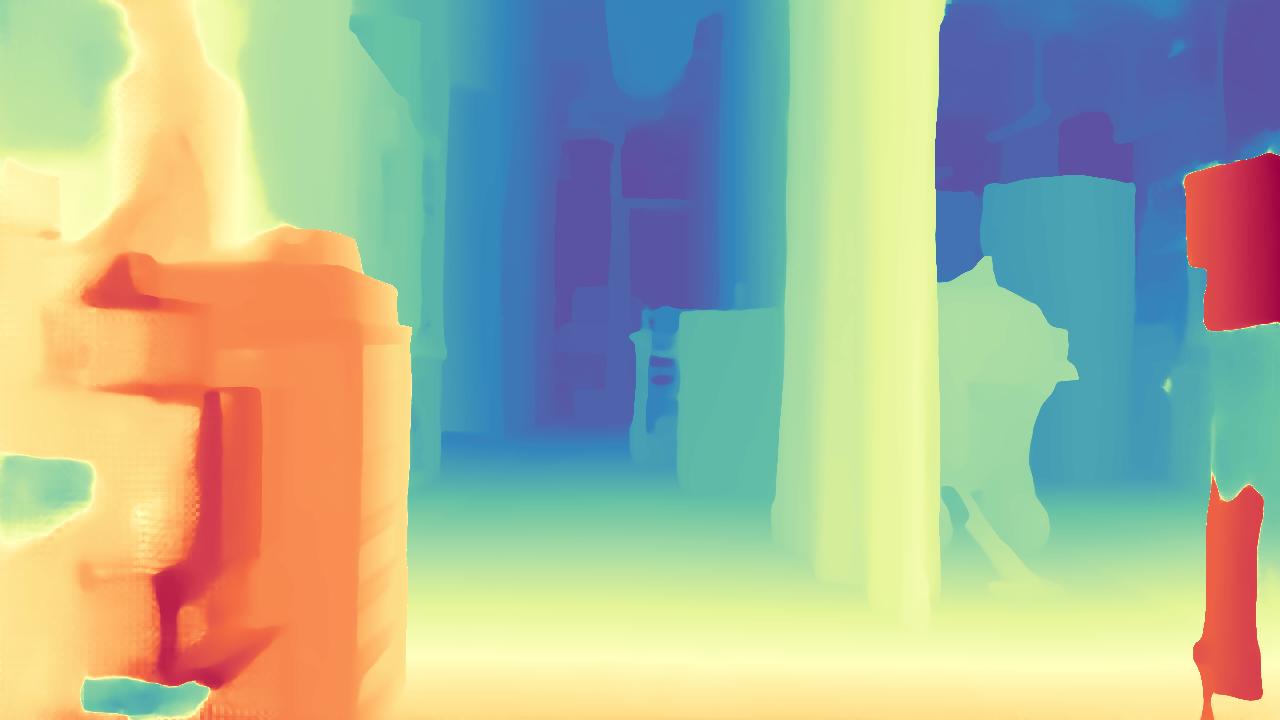} \\[-2pt]
	& \includegraphics[width=0.225\textwidth]{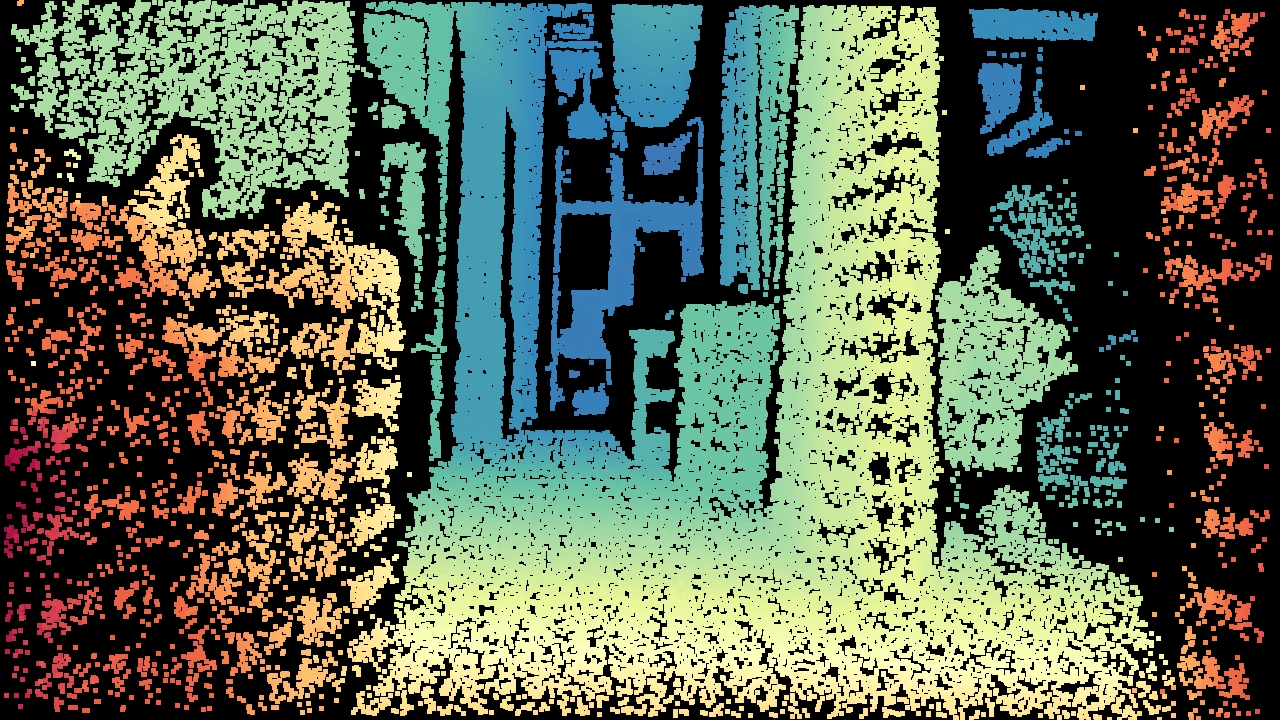} & \;\;& \rotatebox[origin=l]{90}{\hspace{2em}\centering\scriptsize\textbf{MIX~4}} & \includegraphics[width=0.225\textwidth]{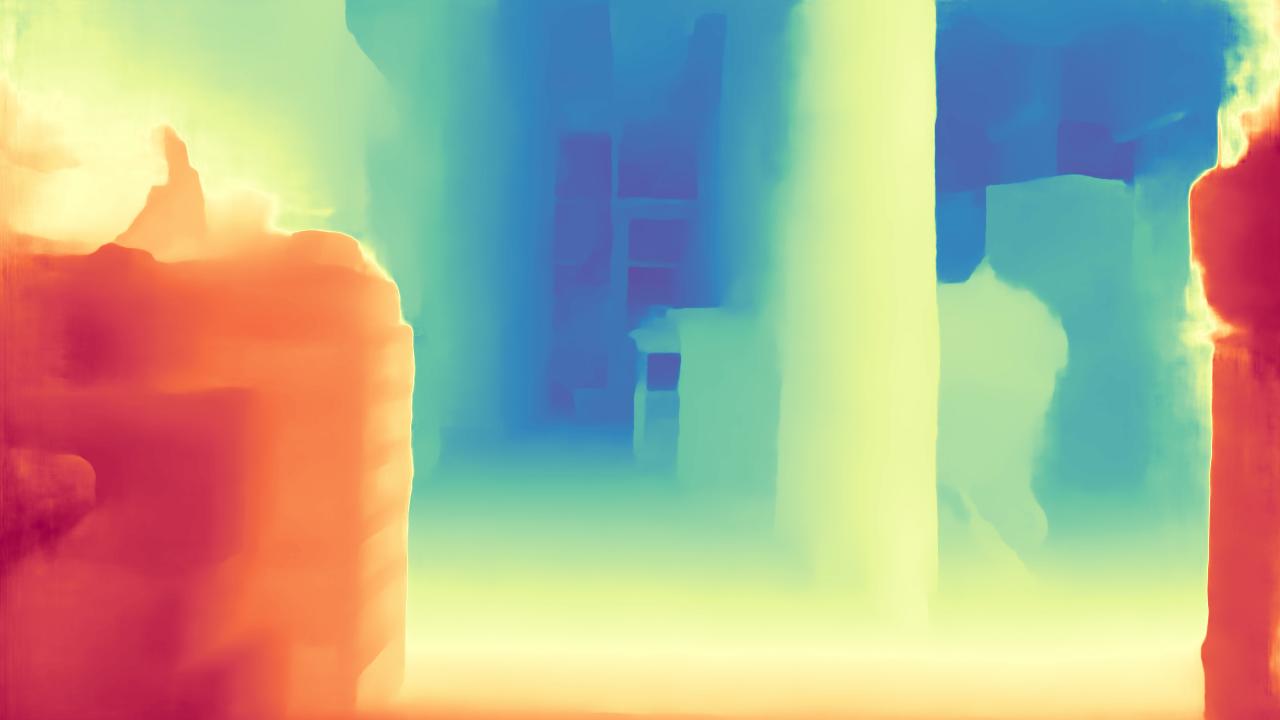} & \includegraphics[width=0.225\textwidth]{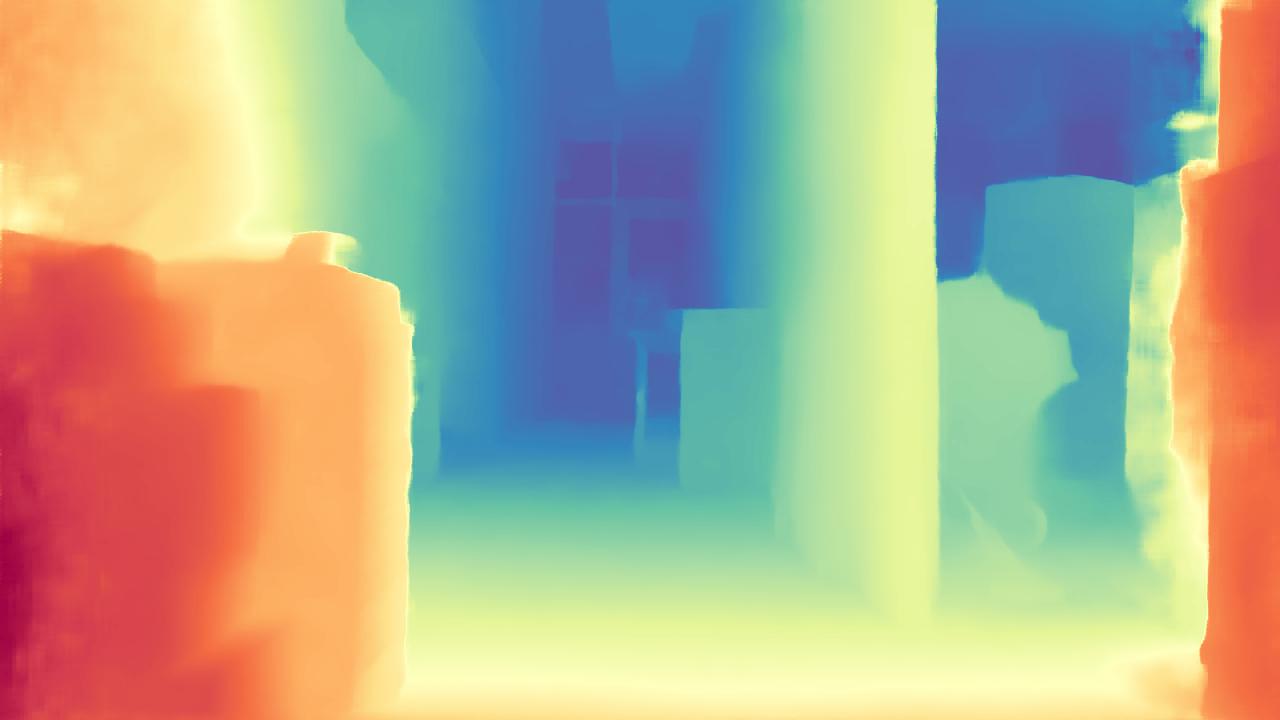} & \includegraphics[width=0.225\textwidth]{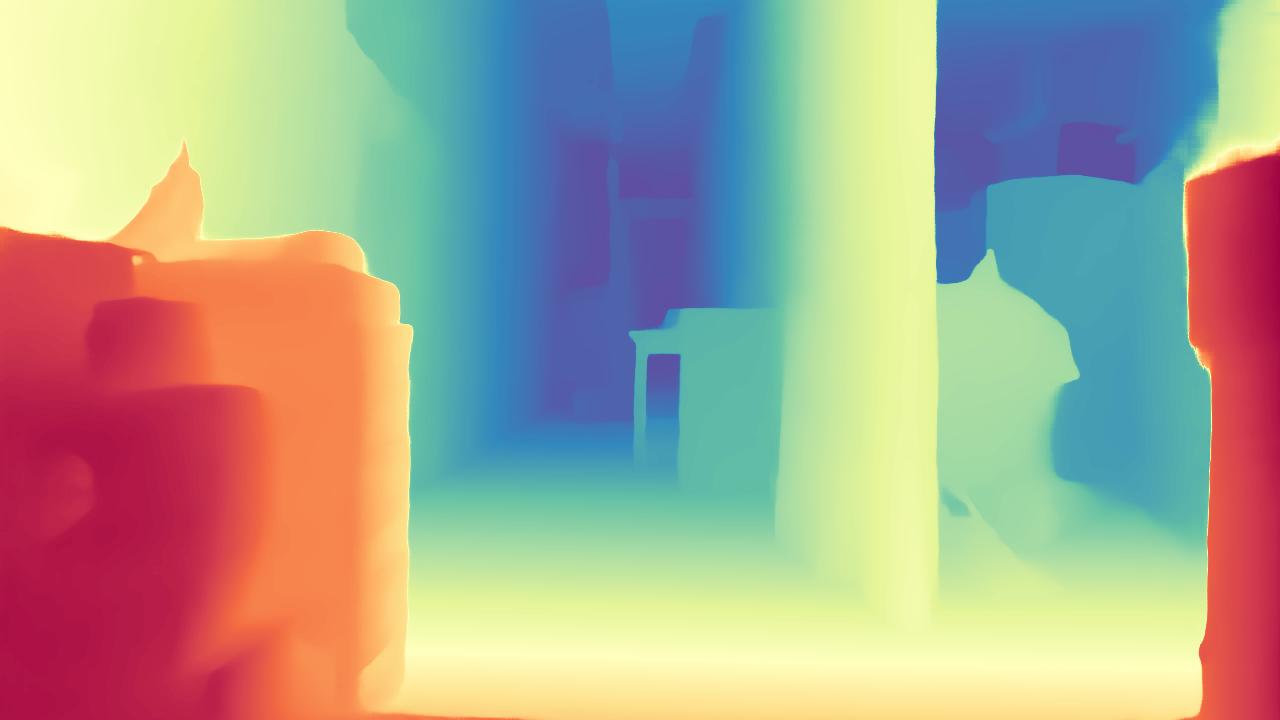} & \includegraphics[width=0.225\textwidth]{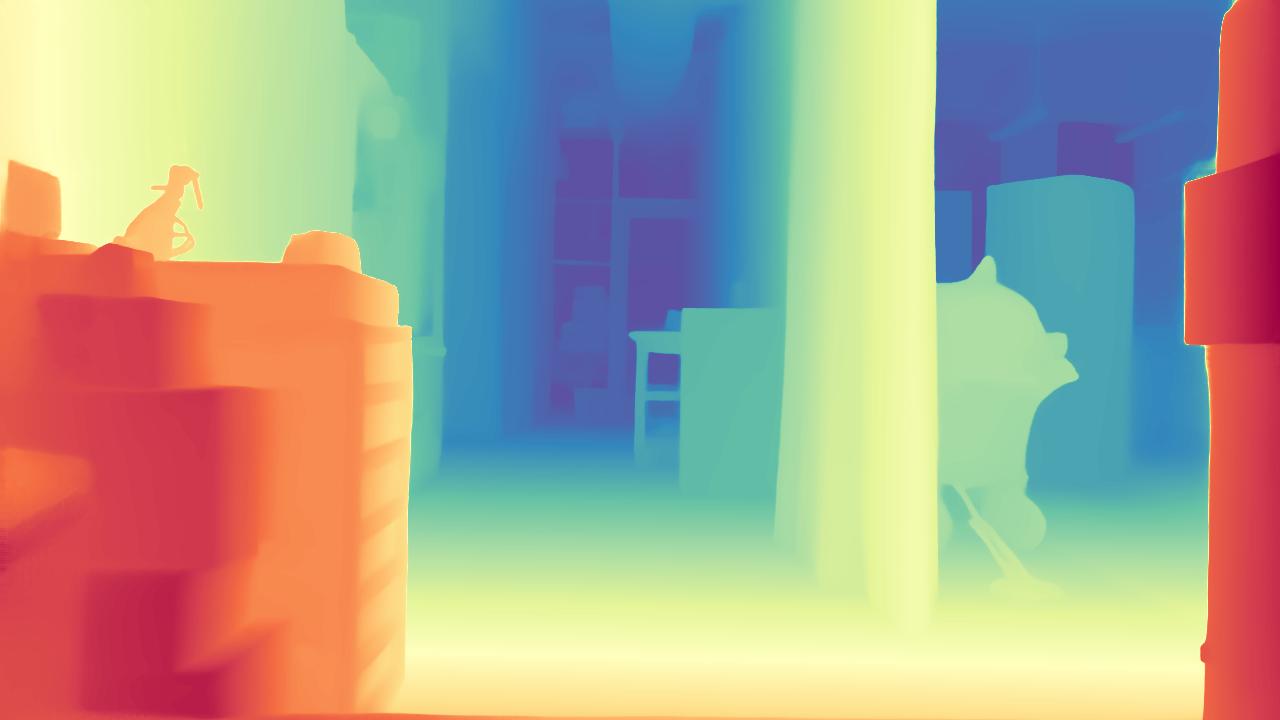} \\[-2pt]
    \multirow{2}{*}{ \rotatebox[origin=l]{90}{\hspace{1.5em}\centering\scriptsize\textbf{MVSEC \cite{zhu2018multivehicle}}}} & \includegraphics[clip,trim=0cm 0.5cm 0cm 1.5cm,width=0.225\textwidth]{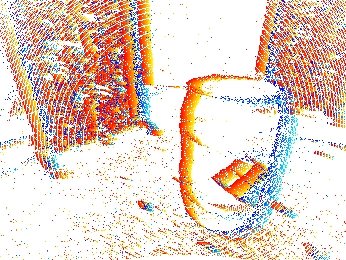} & \;\;& \rotatebox[origin=l]{90}{\hspace{1.25em}\centering\scriptsize\textbf{LiDAR (GT)}} & \includegraphics[clip,trim=0cm 0.5cm 0cm 1cm,width=0.225\textwidth]{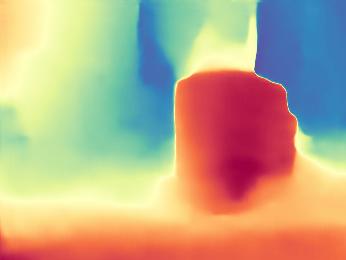} & \includegraphics[clip,trim=0cm 0.5cm 0cm 1cm,width=0.225\textwidth]{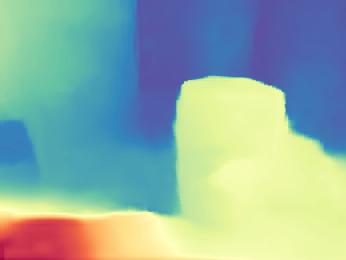} & \includegraphics[clip,trim=0cm 0.5cm 0cm 1cm,width=0.225\textwidth]{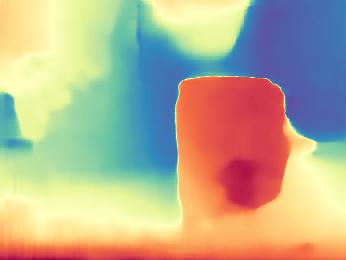} & \includegraphics[clip,trim=0cm 0.5cm 0cm 1cm,width=0.225\textwidth]{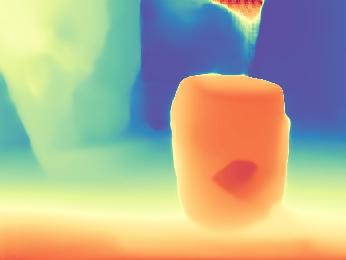} \\[-2pt]
	& \includegraphics[clip,trim=0cm 0.5cm 0cm 1cm,width=0.225\textwidth]{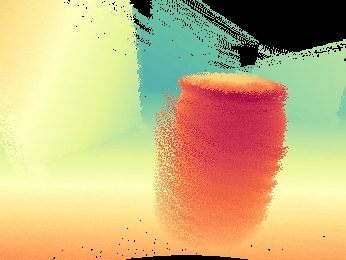} & \;\;& \rotatebox[origin=l]{90}{\hspace{2em}\centering\scriptsize\textbf{MIX~4}} & \includegraphics[clip,trim=0cm 0.5cm 0cm 1cm,width=0.225\textwidth]{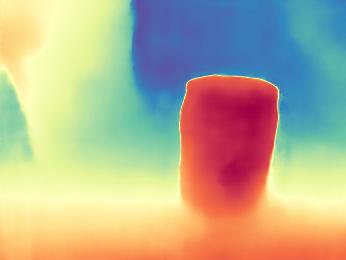} & \includegraphics[clip,trim=0cm 0.5cm 0cm 1cm,width=0.225\textwidth]{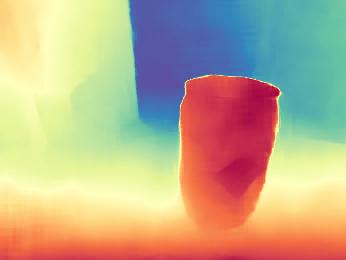} & \includegraphics[clip,trim=0cm 0.5cm 0cm 1cm,width=0.225\textwidth]{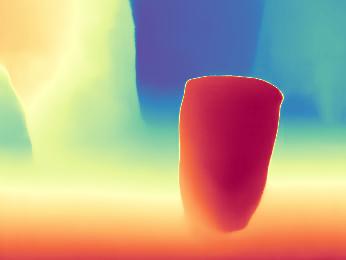} & \includegraphics[clip,trim=0cm 0.5cm 0cm 1cm,width=0.225\textwidth]{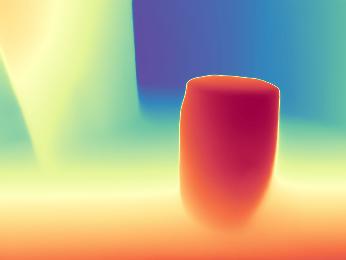} \\
	\end{tabular}
}\vspace{-0.3cm}
\caption{\textbf{Qualitative results on M3ED \cite{chaney2023m3ed} and MVSEC \cite{zhu2018multivehicle}.} Predictions by the four models trained with LiDAR labels or MIX~4.}\vspace{-0.3cm}
\label{fig:qualitative_m3ed_mvsec}
\end{figure*}

\subsection{Evaluation Datasets \& Protocol}

\textbf{Datasets.} We run our evaluation on three main datasets: DSEC \cite{gehrig2021dsec}, M3ED \cite{chaney2023m3ed} and MVSEC \cite{zhu2018multivehicle}. 
DSEC features $640\times480$ px event stereo pairs captured by Prophesee Gen3.1 sensors, with  ground truth depth annotations obtained from a 32-line LiDAR whose scans are accumulated and post-processed. 
We use the validation split proposed in \cite{bartolomei2024lidar} to evaluate in-domain performance. 
M3ED and MVSEC are instead used to evaluate the generalization performance of models trained  under different paradigms (no data from these datasets is used for training). 
M3ED contains $1280\times720$ px event stereo pairs captured by Prophesee IMX636 sensors and annotated by a 64-line LiDAR. MVSEC provides $346\times260$ px event stereo pairs captured by DAVIS346B sensors, annotated with a 16-line LiDAR accumulated via LOAM \cite{zhang2014loam}.

\textbf{Evaluation Metrics.} 
We evaluate the networks using two main disparity metrics: the \emph{Mean-Absolute-Error} (MAE) in pixels, 
and the percentage of pixels having an absolute disparity error larger than a specific threshold, set to 1, 2, and 3 pixels (namely, 1PE, 2PE, and 3PE).

We highlight the \colorbox{firstcolor}{\textbf{best}} and \colorbox{secondcolor}{second-best} scores.

\subsection{In-Domain Evaluation}

We first assess how the different mixtures of data generated by EventHub impact the accuracy of trained models, as well as comparing our training strategies with existing LiDAR-free alternatives as well as with LiDAR supervision. 
\Cref{tab:exp1_a} collects the outcome of this evaluation, carried out by training four event stereo models according to four main strategies. 
The first two rows report results obtained by training the models: (A) using photometric loss between DSEC's RGB stereo images projected into event frame, or (B) using a synthetic event dataset derived from SceneFlow \cite{mayer2016large} via \cite{gehrig2020video}, which provides perfect ground truth disparities, yet proxy event data. 
Both approaches are scarcely effective, with MAE never dropping below 3 pixels for any model. 

Then, we report the results achieved by training on the data and annotation produced by EventHub (C), involving different mixtures of data. 
Notably, MIX~1 already yields largely lower error values, benefiting from the stronger supervision of the proxy labels rendered by SVRaster.
Adding data from ScanNet (MIX~2) yields moderate improvements. 
Training on MIX~3 further boosts performance across all stereo models, which is not surprising since it involves training data from the same domain used in the evaluation.
Nonetheless, combining all data sources (MIX~4) produces the best overall performance for EMatch \cite{zhang2025ematch}, \mbox{E-FoundationStereo} and \mbox{E-StereoAwywhere}.
The bottom row reports the accuracy obtained by running in-domain supervised training using LiDAR ground truth (D), which unsurprisingly yields the lowest errors, yet proving how models trained with MIX~4 get very close to this upper bound despite not using any ground truth annotation from LiDAR.

Despite the thoroughness of \cref{tab:exp1_a}, the shortcomings of LiDAR data for both training and evaluation are fully not apparent.
\Cref{fig:qualitative_dsec_zurich_city_06_a_00005} unveils how the sparse nature of LiDAR annotations hampers the network's ability to produce dense and accurate disparity maps. 
In contrast, training with MIX~3 already avoids most of the artifacts introduced by LiDAR supervision, even though this is not reflected in the error metrics, which are based on LiDAR data.

\begin{table}[t]
    \centering
    \resizebox{\linewidth}{!}{
    \begin{tabular}{l|l|cccc}
    \toprule
     \multirow{2}{*}{\textbf{Model}} & \multirow{2}{*}{\textbf{Training Method}} &  \multicolumn{4}{c}{\textbf{DSEC (Night)}}\\
      &  & \textbf{1PE} & \textbf{2PE} & \textbf{3PE} & \textbf{MAE}\\
     \midrule
     \multirow{2}{*}{{FoundationStereo (ViT-S)}}  & Author's Weights \cite{wen2025foundationstereo} & 68.40 & 47.69 & 35.80 & 3.89 \\
      & MIX~3 (Night) & \fst 24.75 & \fst 8.09 & \fst 4.25 & \fst 1.01 \\ \hline
      \multirow{2}{*}{{FoundationStereo (ViT-L)}} & Author's Weights \cite{wen2025foundationstereo} & 30.06 & 14.87 & 10.88 & 1.87 \\
      & MIX~3 (Night) & \fst 25.33 & \fst 8.48 & \fst 4.56 & \fst 1.06 \\
     \midrule
     \multirow{2}{*}{{StereoAnywhere (ViT-S)}} & Author's Weights \cite{bartolomei2025stereo} & 33.01 & 14.24 & 8.93 & 1.61 \\ 
     & MIX~3 (Night) & \fst 30.61 & \fst 10.81 & \fst 5.72 & \fst 1.22 \\ \hline
     \multirow{2}{*}{{StereoAnywhere (ViT-L)}} & Author's Weights \cite{bartolomei2025stereo} & \fst 31.34 & 13.20 & 8.27 & 1.52 \\
     & MIX~3 (Night) & 32.42 & \fst 11.45 & \fst 5.93 & \fst 1.23 \\
     \bottomrule
    \end{tabular}}\vspace{-0.2cm}
    \caption{\textbf{Experimental results on DSEC night images \cite{gehrig2021dsec} -- RGB stereo models.} 
    Fine-tuning SFMs on proxy labels derived from our event models yields improvements on nighttime images.}\vspace{-0.3cm} 
    \label{tab:proxy_night}
\end{table}

\subsection{Out-of-Domain Evaluation}

We now extend our evaluation beyond the single domain represented by DSEC, moving to M3ED and MVSEC. 
These cover both indoor and outdoor scenarios and are collected by sensors with very different properties, thus representing a significant domain shift with respect to DSEC.   

\Cref{tab:exp2_a_s} presents the results achieved by the four models on M3ED, each supervised with MIX~3, MIX~4 and LiDAR  training strategies, 
i.e., the same models trained on DSEC  and transferred to M3ED without any additional fine-tuning. 
Since MIX~1 and MIX~2 perform worse than MIX~3 and MIX~4, they are omitted from here onward.
We report results averaged over three main subdomains: {\small\texttt{Day}}, {\small\texttt{Night}} and {\small\texttt{Indoor}} scenes. 
We observe that models trained with MIX~4 largely outperform their counterparts trained with LiDAR annotations in terms of generalization. 
This confirms the sub-optimality of supervision provided by sparse and noisy LiDAR measurements, which is often surpassed even by MIX~3 alone (which just replaces LiDAR annotations with proxy labels, without additional training data).

\Cref{tab:exp3_a_s} reports the outcome of the evaluation on MVSEC, using the same four models trained with MIX~3, MIX~4, or LiDAR labels, averaging results over the same three subdomains as before. 
Once again, MIX~4 emerges as the absolute winner in terms of generalization, while MIX~3 consistently achieves the second-best results, except in a few cases.
Finally, \cref{fig:qualitative_m3ed_mvsec} shows examples from M3ED and MVSEC datasets, highlighting that any model produces much sharper and more accurate disparity maps when trained with MIX~4 than with LiDAR labels.

\subsection{Closing the Loop: Improving SFMs at Night}

Finally, we investigate whether the event-based stereo models trained with proxy labels can, in turn, serve as sources of new proxy labels for annotating color images in scenarios where conventional SFMs struggle, such as nighttime conditions.
To this end, we generate proxy labels from stereo pairs $(\mathbf{E}_L,\mathbf{E}_R)$ and transfer them to $(\mathbf{I}_L,\mathbf{I}_R)$, reversing the procedure described in \cref{sec:method:distillation}. 

\Cref{tab:proxy_night} presents the results of this experiment, showing that both FoundationStereo and StereoAnywhere perform poorly on nighttime images. 
After fine-tuning them for 10 epochs on the proxy labels predicted by their \mbox{E-FoundationStereo} and \mbox{E-StereoAnywhere} couterparts, their accuracy improves substantially, effectively closing the loop across modalities.

\section{Conclusion}

We presented \emph{EventHub}, a paradigm for supervising deep event-stereo networks that does not rely on expensive, yet noisy, annotations from LiDAR sensors. 
\emph{EventHub} leverages novel view synthesis and knowledge distillation to obtain proxy labels (and proxy events, when needed) directly from conventional color image collections. 
Models trained with \emph{EventHub} achieve superior generalization, outperforming those trained with LiDAR labels in cross-domain scenarios.
These models can be used to close the loop across modalities, yielding proxy labels to improve RGB stereo models in scenarios where they struggle.

\ifarxiv

\onecolumn
{
    \clearpage
    \centering
    \Large
    \textbf{\thetitle}\\
    \vspace{0.5em}Supplementary Material \\
    \vspace{1.0em}
}

This document reports additional material related to the CVPR paper %
``EventHub: Data Factory for Generalizable Event-Based Stereo Networks without Active Sensors". %

\begin{itemize}
    \item First, we present an extended description of our Novel View Synthesis (NVS) pipeline in \Cref{sec:method}, including details about the depth regularizers used to improve depth estimation (\cref{subsec:regularizers}), and the novel voxel-based confidence $\mathbf{C}_\text{Vsize}$ (\cref{subsec:confidence}).
    
    \item Next, we include additional implementation details, in particular, regarding the global trajectory $\Omega(\tau)$ (\Cref{subsec:global_traj_scannet}), the datasets splits (\cref{subsec:datasets_splits}), and the stereo model losses (\cref{subsec:custom_stereo_losses}).
    
    \item Finally, we present extensive qualitative results regarding both generated data from \cite{tosi2023nerf,yeshwanth2023scannet,gehrig2021dsec} using our EventHub pipeline, and disparity estimation from our trained event stereo networks, using the three evaluation datasets \cite{zhu2018multivehicle,gehrig2021dsec,chaney2023m3ed}.
\end{itemize}

\section{Method Overview: Additional Details}

In this section, we include an extended description of our EventHub pipeline.

\subsection{Depth Regularizers}
\label{subsec:regularizers}

To improve the quality of our NVS generation pipeline, we rely on a subset of the following regularization strategies:

\begin{itemize}
    \item $\mathcal{L}_{N-\text{mean}}$ and $\mathcal{L}_{N-\text{med}}$: both losses encourage agreement between depth and normal renderings, obtained through mean and median aggregation, respectively~\cite{huang20242d};
    \item $\mathcal{L}_\text{DAv2}$ promotes consistency between the rendered depth and the monocular predictions from DepthAnythingV2~\cite{yang2024depth};
    \item $\mathcal{L}_\text{asc}$ encourages density to increase monotonically along the ray direction;
    
    \item $\mathcal{L}_\text{sparse}$ fosters depth regularization using COLMAP~\cite{schonberger2016structure} sparse 3D points;
    
    \item $\mathcal{L}_\text{MASt3R}$ guides the depth regularization following MASt3R predictions.
\end{itemize}

\textbf{Ablation study and metrics for NVS.} To assess the contribution of each depth regularizer, we conducted an ablation experiment on ScanNet++~\cite{yeshwanth2023scannet}, which provides ground-truth depth.
In particular, we selected a small dataset split and evaluated the contribution of each regularizer using two metrics for NVS image quality (\ie, PSNR and SSIM) and two metrics for depth evaluation (\ie, MAE and $\delta\leq\rho$):
\begin{itemize}
    \item \textbf{Peak Signal-to-Noise Ratio (PSNR)}. 
    For color images $\mathbf{I}$ and ground-truth $\mathbf{I}^*$, PSNR is defined based on the mean squared error (MSE):
    \begin{equation}
        \text{MSE} = \frac{1}{N} \sum_i (\mathbf{I}_i - \mathbf{I}_i^*)^2, \qquad
        \text{PSNR} = -10 \log_{10}(\text{MSE}),    
        \label{eq:psnr}
    \end{equation}
    where $N$ is the number of pixels, and $\mathbf{I}_i$ and $\mathbf{I}_i^*$ are the RGB values of the $i$-th pixel in the rendered and ground-truth images, respectively. 
    Higher PSNR indicates better agreement with the ground truth.
    
    \item \textbf{Structural Similarity Index (SSIM)}. 
    SSIM measures similarity between predicted and ground-truth color images $\mathbf{I}$ and $\mathbf{I}^*$ by comparing local windows $\mathbf{X}\in\mathcal{N}_\mathbf{I}$ and $\mathbf{Y}\in\mathcal{N}_\mathbf{I^*}$:
    \begin{equation}
        \text{SSIM}(\mathbf{I}, \mathbf{I}^*) = \frac{1}{M} \sum_{\mathbf{X}\in\mathcal{N}_\mathbf{I}\ \mathbf{Y}\in\mathcal{N}_\mathbf{I^*}}
        \frac{(2\mu_\mathbf{X}\mu_{\mathbf{Y}} + C_1)(2\sigma_{\mathbf{X}\mathbf{Y}} + C_2)}
            {(\mu_\mathbf{X}^2 + \mu_{\mathbf{Y}}^2 + C_1)(\sigma_\mathbf{X}^2 + \sigma_{\mathbf{Y}}^2 + C_2)},
        \label{eq:ssim}
    \end{equation}
    where $M$ is the number of windows, $C_1$ and $C_2$ are constants, $\mu_\mathbf{X}, \mu_{\mathbf{Y}}$ are local means, $\sigma_\mathbf{X}^2, \sigma_{\mathbf{Y}}^2$ the local variances, and $\sigma_{\mathbf{X}\mathbf{Y}}$ the local covariance. 
    Higher values indicate better structural similarity.
    
    \item \textbf{Mean Absolute Error (MAE)}. 
    It measures the average magnitude of errors:
    \begin{equation}
        \text{MAE} = \frac{1}{N} \sum_i |\mathbf{Z}_i - \mathbf{Z}_i^*|,
        \label{eq:MAE}
    \end{equation}
    where $N$ is the number of pixels, $\mathbf{Z}_i$ and $\mathbf{Z}_i^*$ are the predicted and ground-truth depths of the $i$-th pixel, respectively.
    
    \item \textbf{Threshold Accuracy}. 
    It reports the percentage of predicted depths within a threshold $\rho$ (in our ablation experiment $\rho=1.25$) indicating the proportion of accurate predictions:
    \begin{equation}
        \text{Accuracy} = \frac{1}{N} \sum_i \chi\left(\max\Big(\frac{\mathbf{Z}_i}{\mathbf{Z}_i^*}, \frac{\mathbf{Z}_i^*}{\mathbf{Z}_i}\Big) \leq \rho \right) = \frac{1}{N} \sum_i \chi\left( \delta \leq \rho \right)
        \label{eq:delta_accuracy}
    \end{equation}
    where $N$ is the number of pixels, $\mathbf{Z}_i$ and $\mathbf{Z}_i^*$ are the predicted and ground-truth depths of the $i$-th pixel, respectively, and $\chi(\cdot)$ is the indicator function.
\end{itemize}

\begin{table*}[t]
    \centering
    \renewcommand{\tabcolsep}{8pt}
    \resizebox{\linewidth}{!}{
    \begin{tabular}{rcccccc|cccc}
    Row & $\lambda_{{N-\text{mean}}}$ & $\lambda_{N-\text{med}}$ & $\lambda_{\text{asc}}$ & $\lambda_{\text{sparse}}$ & $\lambda_{\text{DAv2}}$ & $\lambda_{\text{MASt3R}}$ & PSNR & SSIM($\times100$) &  MAE (cm) & $\delta \le 1.25$ (\%) \\
    \toprule
       1 & - & - & - & - & - & - & \fst 33.85 & \fst 87.36 & 8.81 & 93.51 \\
       2 & 0.001 & 0.001 & - & - & - & - & 33.25 & 86.77 & 9.03 & 92.43 \\
       3 & 0.001 & 0.001 & 0.01 & - & - & - & 33.25 & 86.77 & 9.03 & 92.47 \\
       4 & 0.001 & 0.001 & 0.01 & 0.01 & - & - & 33.25 & 86.77 & 8.99 & 92.52 \\
       5 & 0.001 & 0.001 & 0.01 & 0.01 & 0.01 & - & 33.19 & 86.73 & 6.61 & 96.23 \\
       6 & 0.001 & 0.001 & 0.01 & 0.01 & 0.01 & 0.01 & 26.86 & 79.64 & 38.71 & 67.31 \\
       7 & 0.001 & 0.001 & 0.01 & - & 0.01 & - & 33.20 & 86.73 & 6.58 & 96.25 \\
       8 & 0.001 & 0.001 & - & - & 0.01 & - & 33.19 & 86.73 & \snd 6.57 & \snd 96.29 \\
       9 & - & - & - & - & 0.01 & - & \snd 33.74 & \snd 87.24 & 7.15 & 95.89 \\
       10 & 0.0005 & 0.0005 & - & - & 0.01 & - & 33.37 & 86.91 &\fst 6.38 & \fst 96.44 \\
    \bottomrule
    \end{tabular}}\vspace{-0.3cm}
    \caption{\textbf{Depth regularization ablation.} 
    PSNR values are given in decibels.
    SSIM values are multiplied by a 100 factor. 
    MAE values are reported in centimeters. }
    \label{tab:svraster_ablation}
\end{table*}

\begin{table}[t]
    \renewcommand{\tabcolsep}{8pt}
    \centering
    \resizebox{0.9\linewidth}{!}{
        \begin{tabular}{lccccr}
            \toprule
            Model & PSNR $\uparrow$ & MAE (cm) $\downarrow$ & $\delta \le 1.25$ (\%) $\uparrow$ & Setup Time (min/scene) $\downarrow$ & FPS $\uparrow$ \\
            \midrule
            Depth Anything v3 \cite{lin2026depth} & 19.19 & 41.94 & 53.92 & \bf $\sim$1 & \bf 165 \\
            Instant-NGP \cite{mueller2022instant} & 29.21 & 24.68 & 82.88 & $\sim$8 & 5 \\
            3DGS \cite{kerbl20233d} & 32.51 & 22.36 & 76.23 & $\sim$20 & \bf 165 \\
            SVRaster \cite{sun2025sparse} & \bf 33.37 & \bf 6.38 & \bf 96.44 & $\sim$20 & 143\\
            \bottomrule
        \end{tabular}
    }
    \vspace{-0.3cm}
    \caption{\textbf{Comparison between different NVS engines.} SVRaster achieves the best trade-off between rendering quality, setup time and rendering speed. 
    }
    \label{tab:comparison} \vspace{-0.3cm}
\end{table}

\textbf{Ablation Analysis.} 
\Cref{tab:svraster_ablation} reports the results of our study on depth-guided regularization terms. 
The left columns indicate the weights $\lambda$ set for each regularizer, starting with the default values from \cite{sun2025sparse}. 
Without any regularization (first row), SVRaster achieves solid PSNR and SSIM but exhibits a relatively large depth error (MAE = 8.81 cm). 
Introducing the first four regularizers -- \ie, $\mathcal{L}_\text{N-mean}$ and $\mathcal{L}_\text{N-med}$ (row 2), $\mathcal{L}_\text{asc}$ (row 3), and $\mathcal{L}_\text{sparse}$ (row 4) -- yields no meaningful improvements, aside from a marginal SSIM gain.
In contrast, incorporating the monocular prior from DepthAnythingV2~\cite{yang2024depth} (row 5) produces a substantial reduction in depth error (25\% decrease in MAE) while preserving nearly unchanged image quality. 
Adding $\mathcal{L}_\text{MASt3R}$ on top of all other regularizers (row 6), however, severely degrades performance.
Given the strong influence of $\mathcal{L}_\text{DAv2}$, we perform additional ablations where the remaining regularizers are removed one at a time (rows 7, 8, and 9). 
This analysis shows minor contribution from $\mathcal{L}_\text{asc}$ and $\mathcal{L}_\text{sparse}$, but disabling $\lambda_{{N-\text{mean}}}$ and $\lambda_{N-\text{med}}$ leads to worse results than those of row 5. 
Therefore, we reintroduce these two terms with halved weights (row 10), which yields the best overall depth performance. 
We adopt this last configuration as the final set of depth-regularization weights for our NVS pipeline.

\textbf{Impact of the NVS engine.} To support our choice of using SVRaster to render both proxy labels and event streams, we report a comparison with other state-of-the-art novel view synthesis approaches in Table \ref{tab:comparison}. In addition to rendering quality, we also consider the setup time necessary to process each single scene before starting the rendering process, as well as the speed at which data is generated. Notably, Depth Anything v3 has the lowest setup time, as it directly predicts a 3DGS field in a feed-forward fashion rather than a per-scene optimization process. However, this speed is traded for a much lower rendering quality. Instant-NGP still requires a low setup time, yet features a very low rendering speed and sub-optimal rendering quality. Finally, although requiring the highest setup time, 3DGS and SVRaster yields the highest rendering quality: among the two, SVRaster shines thanks to the careful use of depth regularization.

\subsection{Novel Voxel-based Confidence}
\label{subsec:confidence}

Despite the added depth regularization, the resulting depth maps may still contain noticeable noise. 
To address this issue, \cite{tosi2023nerf} introduced a trinocular photometric loss:
\begin{equation}
    \mathcal{L}_\text{NS} = \lambda_\text{disp} \cdot \eta(\mathbf{C}_\text{AO};\mu_\text{AO}) \cdot \mathcal{L}_\text{disp} + \mathbf{M}_\text{auto} \cdot \lambda_\text{3p} \cdot (1-\eta(\mathbf{C}_\text{AO};\mu_\text{AO})) \cdot \mathcal{L}_\text{3p},
    \label{eq:nerf_loss}
\end{equation}
where $\mathcal{L}_\text{disp}$ is the disparity supervision loss with respect to the estimated disparity $\mathbf{D}_e$ (further details in \cref{subsec:custom_stereo_losses}), $\lambda_\text{disp}=1.0$ and $\lambda_\text{3p}=0.1$ are the loss weights set to the default values in \cite{tosi2023nerf}, $\eta(\mathbf{C}_\text{AO};\mu_\text{AO})$ is the truncation function that truncates confidence $\mathbf{C}_\text{AO}$ using the threshold $\mu_\text{AO}=0.5$:
\begin{equation}
    \eta(\mathbf{C};\mu) = \begin{cases}
        0 & \text{if}\ \mathbf{C} \leq \mu \\
        \mathbf{C} & \text{otherwise}
    \end{cases}, 
    \qquad \mathbf{C}_\text{AO} = \text{norm}\left(\sum_{i=1}^N T_i \alpha_i^2\right), 
    \qquad \text{norm}(\mathbf{X})=\frac{\mathbf{X}-\min(\mathbf{X})}{\max(\mathbf{X})-\min(\mathbf{X})},
    \label{eq:conf_ao}
\end{equation}
and given the three rendered images $\mathbf{I}_{LL}$, $\mathbf{I}_L$, $\mathbf{I}_R$ -- where $\mathbf{I}_{LL}$ and $\mathbf{I}_R$ are rendered after applying respective stereo translations $(b\ 0\ 0)^\top$ and $(-b\ 0\ 0)^\top$ to the translation component $\mathbf{t}_\tau$ of the virtual trajectory $\Gamma(\tau)$ or $\Omega(\tau)$ -- we can define the trinocular photometric loss $\mathcal{L}_\text{3p}$ as follow:
\begin{equation}
    \mathcal{L}_\text{3p}(\mathbf{I}_{LL},\mathbf{I}_{L},\mathbf{I}_{R}) 
    = \min\Bigl( \mathcal{L}_\text{2p}\bigl(\mathbf{I}_L,\mathcal{W}(\mathbf{I}_{LL},\mathbf{D}_e)\bigr), \mathcal{L}_\text{2p}\bigl(\mathbf{I}_L,\mathcal{W}(\mathbf{I}_{R},-\mathbf{D}_e)\bigr) \Bigr),
    \label{eq:l_3p}
\end{equation}
where $\mathcal{L}_\text{2p}$ is the standard photometric loss, $\mathcal{W}(\cdot,\cdot)$ is the backward warping function using the estimated disparity $\mathbf{D}_e$ from the event stereo model, and $\mathbf{M}_\text{auto}$ is the automasking term that removes untextured regions.
The standard photometric loss $\mathcal{L}_\text{2p}$ and the automasking term $\mathbf{M}_\text{auto}$ are defined, respectively, as follow:
\begin{equation}
    \mathcal{L}_\text{2p}(\textbf{I},\textbf{I}^\mathcal{W}) = \beta \,\frac{1-\text{SSIM}(\textbf{I},\textbf{I}^\mathcal{W})}{2}+(1-\beta) \left|\textbf{I}-\textbf{I}^\mathcal{W}\right|,
    \label{eq:photometric}
\end{equation}
\begin{equation}
    \mathbf{M}_\text{auto} = \chi\left(\min\mathcal{L}_\text{3p}\left(\mathcal{W}(\mathbf{I}_{LL},\mathbf{D}_e),\mathbf{I}_{L},\mathcal{W}(\mathbf{I}_{R},-\mathbf{D}_e)\right)<\min\mathcal{L}_\text{3p}\left(\mathbf{I}_{LL},\mathbf{I}_{L},\mathbf{I}_{R}\right)\right).
    \label{eq:automask}
\end{equation}

\begin{table*}[t]
    \centering
    \renewcommand{\tabcolsep}{8pt}
    \resizebox{0.6\linewidth}{!}{
    \begin{tabular}{lc|ccc}
        Confidence & Threshold & MAE (cm) & $\delta \le 1.25$ (\%) & Density (\%) \\
        \toprule
        - & - & 6.38 & 96.44 & 100.00 \\
        \hline
        $\mathbf{C}_{\text{AO}}$ & 0.35 & \snd 6.26 & \fst 96.60 & \snd 95.45 \\
        $\mathbf{C}_{\text{Vsize}}$ & 0.75 & \fst 6.23 & \snd 96.57 & \fst 97.42 \\
        \hline
        $\mathbf{C}_{\text{AO}}$ & 0.40 & \snd 6.22 & \fst 96.66 & \snd 92.04 \\
        $\mathbf{C}_{\text{Vsize}}$ & 0.80 & \fst 6.15 & \snd 96.61 & \fst 95.56 \\
        \hline
        $\mathbf{C}_{\text{AO}}$ & 0.45 & \snd 6.20 & \fst 96.72 & \snd 87.38 \\
        $\mathbf{C}_{\text{Vsize}}$ & 0.85 & \fst 6.03 & \snd 96.69 & \fst 91.63 \\
        \bottomrule
    \end{tabular}}
    \caption{\textbf{Confidence threshold study.} Comparison between ambient occlusion confidence $\mathbf{C}_\text{AO}$~\cite{tosi2023nerf} and our voxel-based confidence $\mathbf{C}_\text{Vsize}$ on ScanNet++~\cite{yeshwanth2023scannet}. MAE values are reported in centimeters.}
    \label{tab:conf_ablation}
\end{table*}

We studied a replacement for $\mathbf{C}_\text{AO}$ that exploits the properties peculiar to the underlying NVS engine~\cite{sun2025sparse}, and introduced $\mathbf{C}_\text{Vsize}$ using the voxel size as a confidence measure. 
Indeed, voxel sizes are defined during scene optimization %
and encouraged to be smaller for voxels seen from multiple viewpoints (\ie, those points in the scene that are more constrained by multi-view geometry).
With reference to \Cref{eq:cvsize}, we include additional details:
\begin{equation}
    \mathbf{C}_\text{Vsize} = \text{norm}\left(\sum_{i=1}^N T_i s_i\right) \odot \text{norm}\left(\sum_{i=1}^N T_i \alpha_i\right)=\mathbf{C}'_\text{Vsize}\odot\mathbf{C}_\text{hole},
    \label{eq:cvsize_clone}
\end{equation}
where $\mathbf{C}'_\text{Vsize}$ returns high confidence to pixels whose rays intersect small voxels, and $\mathbf{C}_\text{hole}$ is the hole confidence that gives low confidence to pixels whose rays intersect empty space.
We conducted an ablation experiment to compare the performance of our novel voxel-based confidence $\mathbf{C}_\text{Vsize}$ against the ambient occlusion confidence $\mathbf{C}_\text{AO}$ from \cite{tosi2023nerf}.
We evaluated both approaches using different truncation thresholds on a small ScanNet++~\cite{yeshwanth2023scannet} subset (\ie, \texttt{\footnotesize 07f5b601ee}, \texttt{\footnotesize 08bbbdcc3d}, \texttt{\footnotesize 0c5385e84b}, \texttt{\footnotesize 210f741378}, \texttt{\footnotesize 25aa952aa3}, \texttt{\footnotesize 39f36da05b}, \texttt{\footnotesize 56a0ec536c}, \texttt{\footnotesize 5a269ba6fe}, \texttt{\footnotesize a1d9da703c}, \texttt{\footnotesize bc2fce1d81}, \texttt{\footnotesize be0ed6b33c}, \texttt{\footnotesize daffc70503}, \texttt{\footnotesize dc263dfbf0}, \texttt{\footnotesize ef18cf0708}, \texttt{\footnotesize fb564c935d}), reporting depth estimation results in \Cref{tab:conf_ablation}.
Notably, our $\mathbf{C}_\text{Vsize}$ consistently achieves lower MAE while maintaining a higher density if compared to $\mathbf{C}_\text{AO}$.
We selected $\mu_{\text{Vsize}}=0.75$ as the final truncation threshold for our voxel-based confidence.%

\section{Additional Experiments}

We now report further, focused experiments. 

\textbf{Further in-domain comparisons.} Table \ref{tab:exp1_updated} reports some additional experiments on DSEC, aimed at assessing the impact of rendering quality on the accuracy of the trained stereo models. We conduct this further evaluation over two axis: on top, we compare the results achieved by replacing SVRaster as the rendering engine of our pipeline with the feed-forward model Depth Anything v3 \cite{lin2026depth}. Despite the much faster data generation process enabled by this latter, we can observe a significant drop in the accuracy of the trained models; at the bottom, we extend the amount of synthetic data used to generate proxy events with E2VID \cite{gehrig2020video}, specifically by including TartanAir together with Sceneflow. Despite the improvement enabled by the larger amount of initial data, we can still notice a consistent gap between models trained on this kind of data with respect to ours.
Importantly, we emphasize that event data generated from synthetic RGB datasets are not direct competitors to our EventHub framework; rather, the two sources could be combined to enhance performance further.

\textbf{Efficiency Analysis.} In Table \ref{tab:hardware_analysis}, we report the complexity of each of the stereo backbones involved in our experiments, detailing the number of parameters, FLOPs, the runtime and the peak memory usage. SE-CFF stands as the least complex architectures, although achieving the worse results in our evaluation. On the contrary, E-StereoAnywhere and E-FoundationStereo stand as the most computationally intense architectures. 

\textbf{Convergence Analysis.} By fixing the amount of epochs across the different dataset to 10, as described in the main paper, we obtain different amounts of total training steps, possibly biasing the evaluation of the trained models. However, as shown in Figure \ref{fig:plot}, we can appreciate how the models converge pretty soon to stable results, with marginal or no improvements being achieved by extending the training for more iterations, as occurs when using larger data splits such as MIX2 and MIX4.

\begin{table}[t]
\renewcommand{\tabcolsep}{8pt}
\centering
\resizebox{0.9\linewidth}{!}{
    \begin{tabular}{l|rrrr|rrrr}
    \toprule
    \multirow{2}{*}{{Training Method}} & \multicolumn{4}{c|}{{SE-CFF}} & \multicolumn{4}{c}{{E-FoundationStereo}} \\
    & {1PE $\downarrow$} & {2PE $\downarrow$} & {3PE $\downarrow$} & {MAE $\downarrow$} & {1PE $\downarrow$} & {2PE $\downarrow$} & {3PE $\downarrow$} & {MAE $\downarrow$}\\
    \midrule
    MIX 3 (SVRaster \cite{sun2025sparse}) & \bf {24.73} & \bf {8.58} & \bf {5.08} & \bf {1.01} &\bf  20.99 & \bf 6.82 & \bf 4.10 & \bf 0.89 \\
    MIX 3 (Depth Anything v3 \cite{lin2026depth}) & 74.35 & 47.82 & 30.63 & 3.18 & 71.37 & 41.01 & 23.99 & 2.54 \\
    \midrule
    EV-SceneFlow \cite{gehrig2020video,mayer2016large} & 66.30 & 50.18 & 41.47 & 3.50 & 61.80 & 48.04 & 41.68 & 3.10 \\
    EV-(SceneFlow+TartanAir) \cite{gehrig2020video,mayer2016large,tartanair2020iros} & 57.78 & 33.01 & 19.75 & 2.17 & 41.86 & 23.13 & 16.58 & 1.76 \\    
    \bottomrule
    \end{tabular}
    }
\vspace{-0.3cm}
\caption{{\textbf{Further in-domain experimental results -- DSEC dataset \cite{gehrig2021dsec}.} On top: comparison between SVRaster and Depth Anything v3 generated data. At the bottom: results by extending the synthetic data used to generate proxy events.}}
\label{tab:exp1_updated} 
\end{table}

\begin{table}[t]
    \renewcommand{\tabcolsep}{15pt}
    \centering
    \resizebox{1.0\linewidth}{!}{
    \begin{tabular}{lcccc}
        \hline
        Model & Parameters (M) & FLOPs (G) & Runtime (ms) & Peak Memory (MB) \\
        \hline
        SE-CFF \cite{nam2022stereo} & 2.97 & 85.98 & 46.27  & 379.13 \\
        EMatch \cite{zhang2025ematch} & 6.71 & 501.95 & 115.20  & 3090.49 \\
        E-StereoAnywhere & 39.96 & 1566.58 & 219.81 & 1479.82 \\
        E-FoundationStereo & 60.09 & 4445.51 & 280.11  & 1525.13 \\
        \hline
    \end{tabular}
    }
    \vspace{-0.3cm}
    \caption{\textbf{Hardware analysis on DSEC dataset \cite{gehrig2021dsec}.} Measurements taken on a X GPU.}
    \label{tab:hardware_analysis}
\end{table}

\begin{figure}
\renewcommand{\tabcolsep}{1pt}
    \begin{tabular}{cc}
         \includegraphics[width=0.5\textwidth]{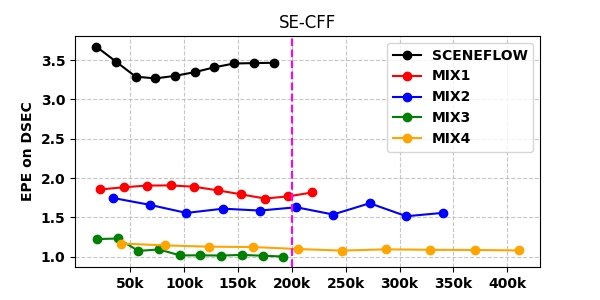} & 
         \includegraphics[width=0.5\textwidth]{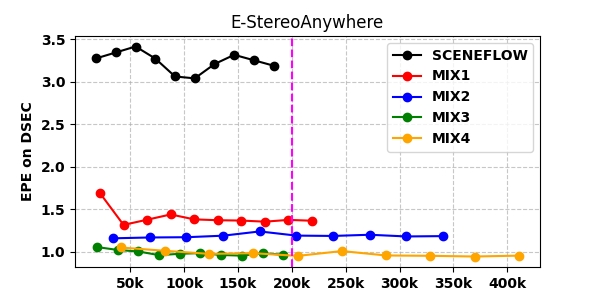} \\ %
    \end{tabular}\vspace{-0.4cm}
    \caption{\textbf{Evaluation on DSEC after different numbers of training steps.}}
    \label{fig:plot}
\end{figure}

\section{Additional Details Concerning Implementation and Experimental Settings}
\label{sec:additional_implementation_details}

In this section, we include additional implementation details, in particular an extended overview of our global trajectory $\Omega(\tau)$, the datasets splits used for both training \cite{tosi2023nerf,yeshwanth2023scannet,gehrig2021dsec} and evaluation \cite{gehrig2021dsec,chaney2023m3ed,zhu2018multivehicle}, and the adaptation of $\mathcal{L}_\text{NS}$ loss for each stereo architecture, \ie, SE-CFF~\cite{nam2022stereo}, EMatch~\cite{zhang2025ematch}, E-StereoAnywhere~\cite{bartolomei2025stereo}, and E-FoundationStereo~\cite{wen2025foundationstereo}.

\subsection{Global Trajectory Implementation}
\label{subsec:global_traj_scannet}

For each selected ScanNet++ scene, we gather all COLMAP training poses $[\hat{\mathbf{R}}_i|\hat{\mathbf{t}}_i]=\hat{\mathbf{T}}_i \in \mathbb{SE}(3)$ and project $\hat{\mathbf{t}}_i=(\hat{x}_i,\hat{y}_i,\hat{z}_i)^\top$ onto a 2D top-view by discarding the last $\hat{z}_i$ component.
We then compute the corresponding $\alpha$-shape, yielding an obstacle-avoiding 2D circular path.
The resulting 2D curve is lifted back to 3D via a nearest-neighbor search, which we used to optimize the three splines using least squares (implemented using the SciPy package). As detailed in the main paper, one spline provides a continuous representation of the translation component $\mathbf{t}_\tau$, while the splines $\mathbf{r}(\tau)$ and $\mathbf{l}(\tau)$ parametrize the rotation component $\mathbf{R}_\tau$:
\begin{equation}
    \mathbf{R}_\tau =
    \begin{bmatrix}
        \mathbf{d}(\tau) \times \mathbf{l}(\tau) & 
        \mathbf{d}(\tau) & 
        \mathbf{l}(\tau)
    \end{bmatrix},\quad \mathbf{d}(\tau) = \mathbf{l}(\tau) \times \mathbf{r}(\tau).
    \label{eq:recover_r_tau}
\end{equation}
However, given the ScanNet++ randomness of pose orientation, which causes unnatural camera egomotion, we re-estimate camera rotations $\hat{\mathbf{R}}_i$ to align them with the direction of motion.
Specifically, we approximate the motion direction $\nabla\mathbf{t}_\tau$ using finite differences, and construct the updated orientation:
\begin{equation}
    \mathbf{R}'_\tau = \begin{bmatrix}
        \mathbf{r}'(\tau) & \mathbf{d}'(\tau) & \nabla\mathbf{t}_\tau
    \end{bmatrix},\quad \mathbf{r}'(\tau) = \mathbf{g}\times\nabla\mathbf{t}_\tau,\quad \mathbf{d}'(\tau) = \nabla\mathbf{t}_\tau\times\mathbf{r}'(\tau),
    \label{eq:motion_orientation}
\end{equation}
where $\mathbf{g}=\left(0\ 0\ 1\right)^\top$ denotes the ScanNet++ gravity vector.
Finally, we clamp the $\hat{z}_i$ translation component to its $[45,55]$-th percentile range to suppress strong vertical oscillations.
This procedure yields a ``human-like" walking trajectory through the scene, as shown in \Cref{fig:global_traj} (right).

\begin{figure}
    \centering
    \begin{tabular}{cc}
        \includegraphics[width=0.45\textwidth]{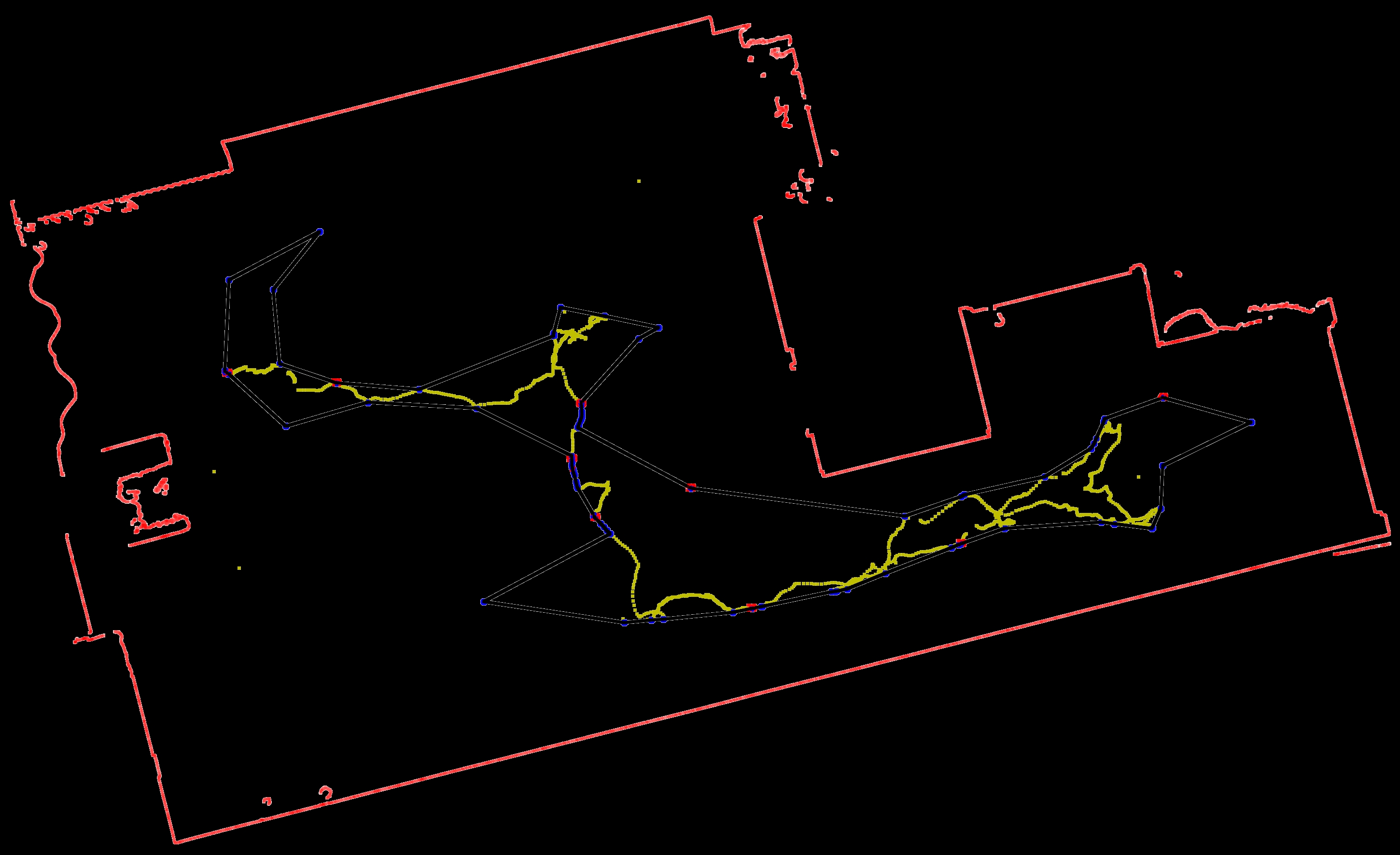} & \includegraphics[width=0.45\textwidth]{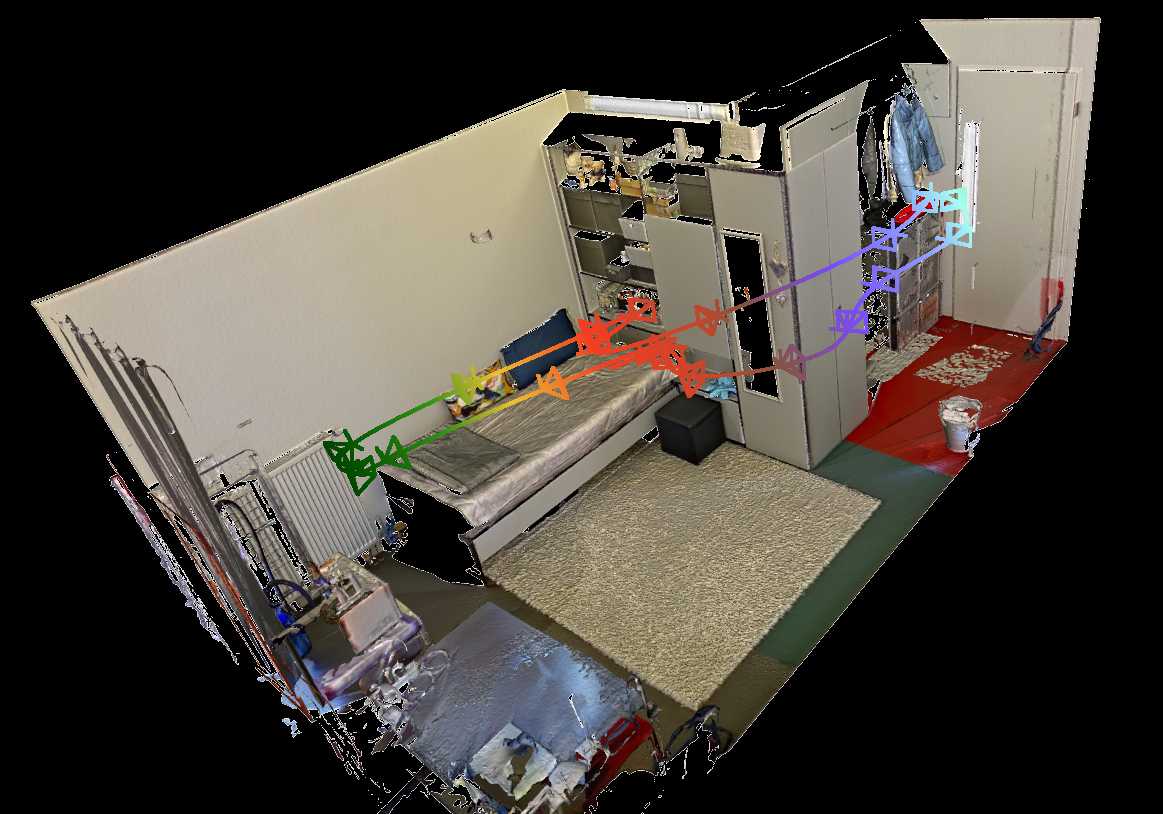} \\
        2D top-view showing the $\alpha$-shape (black) & Final 3D global trajectory inside the scene
    \end{tabular}
    \caption{\textbf{Global trajectory construction.} }
    \label{fig:global_traj}
\end{figure}

\subsection{ScanNet++ Scenes Used for NVS}
\label{subsec:datasets_splits}

We enrich \Cref{subsec:implementation_details} with further information regarding the dataset used for event data generation \cite{tosi2023nerf,gehrig2021dsec,yeshwanth2023scannet}. %
For event data generation from Novel View Synthesis, we collect 30 samples for each scene, where each sample is composed of the stereo streams $\mathbf{E}_L$ and $\mathbf{E}_R$, the intrinsic $\mathbf{K}$, the baseline $b$, the RGB triplet $\mathbf{I}_{LL}$, $\mathbf{I}_{L}$, and $\mathbf{I}_{R}$, the depth $\mathbf{Z}$ and the confidence $\mathbf{C}_\text{Vsize}$.
The maximum number of events for the event stereo streams $\mathbf{E}_L$ and $\mathbf{E}_R$ is limited to $650\, 000$ and $1\, 000\, 000$ events, respectively, for the samples at resolutions $640\times480$ px and $1280\times720$ px.  
Furthermore, we randomize the contrast threshold using a uniform distribution $\mathcal{U}(0.15,0.25)$.
We used all 270 scenes from the NeRF Stereo Dataset~\cite{tosi2023nerf} -- \ie, starting from scene $0000$ up to scene $0269$ -- while we selected the following 403 scenes from ScanNet++\cite{yeshwanth2023scannet}: \texttt{\footnotesize 00777c41d4}, \texttt{\footnotesize 0271889ec0}, \texttt{\footnotesize 02c2ddee2a}, \texttt{\footnotesize 036bce3393}, \texttt{\footnotesize 0452249a1e}, \texttt{\footnotesize 04d0dc245b}, \texttt{\footnotesize 04df8734b7}, \texttt{\footnotesize 052d72e137}, \texttt{\footnotesize 0658da5bc0}, \texttt{\footnotesize 068ba2946c}, \texttt{\footnotesize 06b5863f73}, \texttt{\footnotesize 06bc6d1b24}, \texttt{\footnotesize 076c822ecc}, \texttt{\footnotesize 079a326597}, \texttt{\footnotesize 07f5b601ee}, \texttt{\footnotesize 08bbbdcc3d}, \texttt{\footnotesize 09a6767fc2}, \texttt{\footnotesize 09bced689e}, \texttt{\footnotesize 0a5c013435}, \texttt{\footnotesize 0c5385e84b}, \texttt{\footnotesize 0c6c7145ba}, \texttt{\footnotesize 0c7962bd64}, \texttt{\footnotesize 0caa1ae59a}, \texttt{\footnotesize 0d8ead0038}, \texttt{\footnotesize 0e100756bf}, \texttt{\footnotesize 0e350246d3}, \texttt{\footnotesize 0e900bcc5c}, \texttt{\footnotesize 0f0191b10b}, \texttt{\footnotesize 0f25f24a4f}, \texttt{\footnotesize 0f3474b837}, \texttt{\footnotesize 10242d1eaf}, \texttt{\footnotesize 10c8ab99f4}, \texttt{\footnotesize 1117299565}, \texttt{\footnotesize 1204e08f17}, \texttt{\footnotesize 124a6e789b}, \texttt{\footnotesize 12c0f7a7da}, \texttt{\footnotesize 13285009a4}, \texttt{\footnotesize 132cb783ed}, \texttt{\footnotesize 13b4efaf62}, \texttt{\footnotesize 15c4aa5bbb}, \texttt{\footnotesize 16c9bd2e1e}, \texttt{\footnotesize 1730c7d709}, \texttt{\footnotesize 1841a0b525}, \texttt{\footnotesize 192ab15daf}, \texttt{\footnotesize 1a130d092a}, \texttt{\footnotesize 1a3100752b}, \texttt{\footnotesize 1a8e0d78c0}, \texttt{\footnotesize 1b9692f0c7}, \texttt{\footnotesize 1bb93d185e}, \texttt{\footnotesize 1c08823a41}, \texttt{\footnotesize 1c4b893630}, \texttt{\footnotesize 1c7a683c92}, \texttt{\footnotesize 1d003b07bd}, \texttt{\footnotesize 1eacc65607}, \texttt{\footnotesize 20871b98f3}, \texttt{\footnotesize 20ff72df6e}, \texttt{\footnotesize 210f741378}, \texttt{\footnotesize 216b9e55e8}, \texttt{\footnotesize 238b940049}, \texttt{\footnotesize 246fe09e98}, \texttt{\footnotesize 2489b7f4fe}, \texttt{\footnotesize 24b248e676}, \texttt{\footnotesize 251443268c}, \texttt{\footnotesize 25aa952aa3}, \texttt{\footnotesize 25bae29ab3}, \texttt{\footnotesize 25bde9e167}, \texttt{\footnotesize 260db9cf5a}, \texttt{\footnotesize 260fa55d50}, \texttt{\footnotesize 2634683a9f}, \texttt{\footnotesize 2748de13fb}, \texttt{\footnotesize 2779f8f9e2}, \texttt{\footnotesize 27dc178a3d}, \texttt{\footnotesize 281bc17764}, \texttt{\footnotesize 2970e95b65}, \texttt{\footnotesize 29c7afafed}, \texttt{\footnotesize 2a1b555966}, \texttt{\footnotesize 2a496183e1}, \texttt{\footnotesize 2b71155e0d}, \texttt{\footnotesize 2f5996ff01}, \texttt{\footnotesize 2f6f83ea1f}, \texttt{\footnotesize 302a7f6b67}, \texttt{\footnotesize 303745abc7}, \texttt{\footnotesize 30f4a2b44d}, \texttt{\footnotesize 320c3af000}, \texttt{\footnotesize 324d07a5b3}, \texttt{\footnotesize 3391ff8a71}, \texttt{\footnotesize 3423e509af}, \texttt{\footnotesize 35050f41c5}, \texttt{\footnotesize 355e5e32db}, \texttt{\footnotesize 364f01bc18}, \texttt{\footnotesize 37562e7f48}, \texttt{\footnotesize 3799bd47b3}, \texttt{\footnotesize 37c9538a2b}, \texttt{\footnotesize 38fcf02d0b}, \texttt{\footnotesize 390eda9157}, \texttt{\footnotesize 39580e2a43}, \texttt{\footnotesize 39e6ee46df}, \texttt{\footnotesize 39f36da05b}, \texttt{\footnotesize 3a3745a437}, \texttt{\footnotesize 3aa115e55e}, \texttt{\footnotesize 3b90310b1c}, \texttt{\footnotesize 3c8d535d49}, \texttt{\footnotesize 3caf4324fd}, \texttt{\footnotesize 3cbb18c391}, \texttt{\footnotesize 3ce6d36ab5}, \texttt{\footnotesize 3d838ee1ab}, \texttt{\footnotesize 3e7e4b07c4}, \texttt{\footnotesize 3e928dc2f6}, \texttt{\footnotesize 3ff873c77e}, \texttt{\footnotesize 413085a827}, \texttt{\footnotesize 41b00feddb}, \texttt{\footnotesize 4318f8bb3c}, \texttt{\footnotesize 4380e4646a}, \texttt{\footnotesize 43cd995c51}, \texttt{\footnotesize 4422722c49}, \texttt{\footnotesize 4423a61d09}, \texttt{\footnotesize 442b144761}, \texttt{\footnotesize 44c85584ae}, \texttt{\footnotesize 4517d988d8}, \texttt{\footnotesize 45d2e33be1}, \texttt{\footnotesize 46001f434d}, \texttt{\footnotesize 4610b2104c}, \texttt{\footnotesize 46638cfd0f}, \texttt{\footnotesize 47b37eb6f9}, \texttt{\footnotesize 4808c4a397}, \texttt{\footnotesize 480ddaadc0}, \texttt{\footnotesize 484ad681df}, \texttt{\footnotesize 48573f4c95}, \texttt{\footnotesize 48701abb21}, \texttt{\footnotesize 4897e95232}, \texttt{\footnotesize 49789448b8}, \texttt{\footnotesize 4aef651da7}, \texttt{\footnotesize 4c141d5b1b}, \texttt{\footnotesize 4c5c60fa76}, \texttt{\footnotesize 4d451d9c36}, \texttt{\footnotesize 4e0b8cbd33}, \texttt{\footnotesize 4ea827f5a1}, \texttt{\footnotesize 504cf57907}, \texttt{\footnotesize 511061232}, \texttt{\footnotesize 51bdbf173f}, \texttt{\footnotesize 523657b4d0}, \texttt{\footnotesize 5334a4164a}, \texttt{\footnotesize 53755e535e}, \texttt{\footnotesize 546292a9db}, \texttt{\footnotesize 54b005d19d}, \texttt{\footnotesize 55b2bf8036}, \texttt{\footnotesize 5654092cc2}, \texttt{\footnotesize 56669a70bc}, \texttt{\footnotesize 56a0ec536c}, \texttt{\footnotesize 58960ff105}, \texttt{\footnotesize 589f5c7c58}, \texttt{\footnotesize 58f6a5c5ec}, \texttt{\footnotesize 59e3f1ea37}, \texttt{\footnotesize 5a269ba6fe}, \texttt{\footnotesize 5a9cdde1ba}, \texttt{\footnotesize 5aeac3800a}, \texttt{\footnotesize 5bc6227191}, \texttt{\footnotesize 5c215ef3b0}, \texttt{\footnotesize 5d152fab1b}, \texttt{\footnotesize 5d902f1593}, \texttt{\footnotesize 5ea3e738c3}, \texttt{\footnotesize 5f0fb991a7}, \texttt{\footnotesize 6126572846}, \texttt{\footnotesize 612f70fe00}, \texttt{\footnotesize 617326da3e}, \texttt{\footnotesize 618310ed87}, \texttt{\footnotesize 6183f0657d}, \texttt{\footnotesize 61adeff7d5}, \texttt{\footnotesize 6248c6742d}, \texttt{\footnotesize 635852d56e}, \texttt{\footnotesize 639f2c4d5a}, \texttt{\footnotesize 6464461276}, \texttt{\footnotesize 64672b5bf5}, \texttt{\footnotesize 652d9cb0d7}, \texttt{\footnotesize 666d04a14a}, \texttt{\footnotesize 66ba53719a}, \texttt{\footnotesize 66c98f4a9b}, \texttt{\footnotesize 67d702f2e8}, \texttt{\footnotesize 696317583f}, \texttt{\footnotesize 69e56cf0f8}, \texttt{\footnotesize 69e5939669}, \texttt{\footnotesize 6ad6cef000}, \texttt{\footnotesize 6b19334aeb}, \texttt{\footnotesize 6b40d1a939}, \texttt{\footnotesize 6bd39ac392}, \texttt{\footnotesize 6da1d5ab04}, \texttt{\footnotesize 6f1848d1e3}, \texttt{\footnotesize 70945f435a}, \texttt{\footnotesize 709ab5bffe}, \texttt{\footnotesize 70f0e494b2}, \texttt{\footnotesize 712b9ae775}, \texttt{\footnotesize 724c40236c}, \texttt{\footnotesize 72f527a47c}, \texttt{\footnotesize 73f9370962}, \texttt{\footnotesize 75d29d69b8}, \texttt{\footnotesize 7739004a45}, \texttt{\footnotesize 77b40ce601}, \texttt{\footnotesize 785e7504b9}, \texttt{\footnotesize 791a5c253d}, \texttt{\footnotesize 7b04052ad0}, \texttt{\footnotesize 7b4a316aea}, \texttt{\footnotesize 7b4cb756d4}, \texttt{\footnotesize 7c0ba828a9}, \texttt{\footnotesize 7c31a42404}, \texttt{\footnotesize 7c31bccde5}, \texttt{\footnotesize 7d8d37ca38}, \texttt{\footnotesize 7e7d2e8640}, \texttt{\footnotesize 7f22d5ef1b}, \texttt{\footnotesize 7f68c514bd}, \texttt{\footnotesize 7f77abce34}, \texttt{\footnotesize 7fb8ff20e9}, \texttt{\footnotesize 8013901416}, \texttt{\footnotesize 80ffca8a48}, \texttt{\footnotesize 81a82c3618}, \texttt{\footnotesize 82f448db76}, \texttt{\footnotesize 82ff39b7ef}, \texttt{\footnotesize 85251de7d1}, \texttt{\footnotesize 85dc2702b7}, \texttt{\footnotesize 867d97cf3d}, \texttt{\footnotesize 871efc90fa}, \texttt{\footnotesize 8737a0d1ad}, \texttt{\footnotesize 88627b561e}, \texttt{\footnotesize 8890d0a267}, \texttt{\footnotesize 88f265fe25}, \texttt{\footnotesize 893fb90e89}, \texttt{\footnotesize 8be0cd3817}, \texttt{\footnotesize 8d0f714398}, \texttt{\footnotesize 8de35c04a3}, \texttt{\footnotesize 8e22c48c20}, \texttt{\footnotesize 8f82c394d6}, \texttt{\footnotesize 8fc40ba77b}, \texttt{\footnotesize 9084d4cd97}, \texttt{\footnotesize 909a9ea5fc}, \texttt{\footnotesize 91fc568d84}, \texttt{\footnotesize 9444b90aaa}, \texttt{\footnotesize 9471b8d485}, \texttt{\footnotesize 94b1acde81}, \texttt{\footnotesize 95748dd597}, \texttt{\footnotesize 95d525fbfd}, \texttt{\footnotesize 97e5512e91}, \texttt{\footnotesize 9816c49e97}, \texttt{\footnotesize 98b4ec142f}, \texttt{\footnotesize 98fe276aa8}, \texttt{\footnotesize 99010a8938}, \texttt{\footnotesize 9b74afd2d2}, \texttt{\footnotesize 9bfbc75700}, \texttt{\footnotesize 9c7b4394af}, \texttt{\footnotesize 9cfea269dd}, \texttt{\footnotesize 9d8fcc4215}, \texttt{\footnotesize 9dc5ad040f}, \texttt{\footnotesize 9ef5fc6271}, \texttt{\footnotesize a08d9a2476}, \texttt{\footnotesize a1d9da703c}, \texttt{\footnotesize a23f391ba9}, \texttt{\footnotesize a30646cae6}, \texttt{\footnotesize a31b2ef388}, \texttt{\footnotesize a492fe77aa}, \texttt{\footnotesize a4d48ea6b3}, \texttt{\footnotesize a4e227f506}, \texttt{\footnotesize a892730b61}, \texttt{\footnotesize a8f7f66985}, \texttt{\footnotesize a9e4791c7e}, \texttt{\footnotesize aa852f7871}, \texttt{\footnotesize aab83fd6f1}, \texttt{\footnotesize ab046f8faf}, \texttt{\footnotesize ab11145646}, \texttt{\footnotesize ab6983ae6c}, \texttt{\footnotesize abf29d2474}, \texttt{\footnotesize ac250f0ead}, \texttt{\footnotesize acd69a1746}, \texttt{\footnotesize ad2d07fd11}, \texttt{\footnotesize adf4ab4a53}, \texttt{\footnotesize aea84db0de}, \texttt{\footnotesize b068706ef0}, \texttt{\footnotesize b08a908f0f}, \texttt{\footnotesize b09431c547}, \texttt{\footnotesize b0b004c40f}, \texttt{\footnotesize b0f057c684}, \texttt{\footnotesize b0fe0c610f}, \texttt{\footnotesize b1d75ecd55}, \texttt{\footnotesize b20a261fdf}, \texttt{\footnotesize b24697b3a1}, \texttt{\footnotesize b2632b738a}, \texttt{\footnotesize b3ac0beef0}, \texttt{\footnotesize b4b39438f0}, \texttt{\footnotesize b5918e4637}, \texttt{\footnotesize b6d73041c8}, \texttt{\footnotesize b97261909e}, \texttt{\footnotesize bac7ee3b1b}, \texttt{\footnotesize bb05a0c48c}, \texttt{\footnotesize bb0ad8a081}, \texttt{\footnotesize bc2fce1d81}, \texttt{\footnotesize bc400d86e1}, \texttt{\footnotesize be05b26a38}, \texttt{\footnotesize be0ed6b33c}, \texttt{\footnotesize be8367fcbe}, \texttt{\footnotesize bf07750a0b}, \texttt{\footnotesize bf50f418ba}, \texttt{\footnotesize bfcfe53c6a}, \texttt{\footnotesize bfd3fd54d2}, \texttt{\footnotesize c026d108e0}, \texttt{\footnotesize c07c707449}, \texttt{\footnotesize c08d1d52b7}, \texttt{\footnotesize c0da8f4a4d}, \texttt{\footnotesize c0f5742640}, \texttt{\footnotesize c29b5e479c}, \texttt{\footnotesize c2d714d386}, \texttt{\footnotesize c31ebd4b22}, \texttt{\footnotesize c40466a844}, \texttt{\footnotesize c465f388d1}, \texttt{\footnotesize c47168fab2}, \texttt{\footnotesize c4aaedcfd1}, \texttt{\footnotesize c4d4cb61f6}, \texttt{\footnotesize c601466b77}, \texttt{\footnotesize c842edbdf5}, \texttt{\footnotesize c856c41c99}, \texttt{\footnotesize c8eeef6427}, \texttt{\footnotesize c8f2218ee2}, \texttt{\footnotesize c9a8357e8f}, \texttt{\footnotesize ca0c580422}, \texttt{\footnotesize cab239278a}, \texttt{\footnotesize cb7785f6ad}, \texttt{\footnotesize cc5ea8026c}, \texttt{\footnotesize ccfd3ed9c7}, \texttt{\footnotesize cd0b6082d2}, \texttt{\footnotesize ce12db9e81}, \texttt{\footnotesize cec8312f4e}, \texttt{\footnotesize d054227009}, \texttt{\footnotesize d1345a65c1}, \texttt{\footnotesize d1f82299d0}, \texttt{\footnotesize d240136ce4}, \texttt{\footnotesize d2f44bf242}, \texttt{\footnotesize d537ef1d41}, \texttt{\footnotesize d551dac194}, \texttt{\footnotesize d61691f945}, \texttt{\footnotesize d6a77f7c22}, \texttt{\footnotesize d6bb698875}, \texttt{\footnotesize d7abfc4b17}, \texttt{\footnotesize d7b871aaa8}, \texttt{\footnotesize d807fb583b}, \texttt{\footnotesize d918af9c5f}, \texttt{\footnotesize d986399f4c}, \texttt{\footnotesize daffc70503}, \texttt{\footnotesize db5293a870}, \texttt{\footnotesize dc263dfbf0}, \texttt{\footnotesize dd685be466}, \texttt{\footnotesize de3c77cecd}, \texttt{\footnotesize de5881aa12}, \texttt{\footnotesize deb1867829}, \texttt{\footnotesize dec0b11090}, \texttt{\footnotesize defd3457db}, \texttt{\footnotesize dfa70fb232}, \texttt{\footnotesize dfac5b38df}, \texttt{\footnotesize e050c15a8d}, \texttt{\footnotesize e0de253456}, \texttt{\footnotesize e1aa584dd5}, \texttt{\footnotesize e2caaaf5b5}, \texttt{\footnotesize e3ad7115db}, \texttt{\footnotesize e3b3b0d0c7}, \texttt{\footnotesize e3c1da58dd}, \texttt{\footnotesize e3e0617f98}, \texttt{\footnotesize e3ecd49e2b}, \texttt{\footnotesize e3ef8b690b}, \texttt{\footnotesize e4007ff6b5}, \texttt{\footnotesize e4e625a3e4}, \texttt{\footnotesize e4fb2a623b}, \texttt{\footnotesize e5a769dbf5}, \texttt{\footnotesize e667e09fe6}, \texttt{\footnotesize e69064f2f3}, \texttt{\footnotesize e7ccd75e5d}, \texttt{\footnotesize e81c8b3eec}, \texttt{\footnotesize e8e81396b6}, \texttt{\footnotesize e8ea9b4da8}, \texttt{\footnotesize e909f8213d}, \texttt{\footnotesize e9e16b6043}, \texttt{\footnotesize eaa6c90310}, \texttt{\footnotesize eaab7bcc15}, \texttt{\footnotesize eab5494dca}, \texttt{\footnotesize eb8ef9b4cc}, \texttt{\footnotesize ec2cb8dae1}, \texttt{\footnotesize ed2216380b}, \texttt{\footnotesize eea4ad9c04}, \texttt{\footnotesize eeeb9836b8}, \texttt{\footnotesize ef18cf0708}, \texttt{\footnotesize ef25276c25}, \texttt{\footnotesize f19ca0a52e}, \texttt{\footnotesize f248c2bcdc}, \texttt{\footnotesize f25f5e6f63}, \texttt{\footnotesize f2e6c43543}, \texttt{\footnotesize f38b0108a1}, \texttt{\footnotesize f3f016ba3f}, \texttt{\footnotesize f576071590}, \texttt{\footnotesize f6659a3107}, \texttt{\footnotesize f6a9b64a0d}, \texttt{\footnotesize f847086d15}, \texttt{\footnotesize f8d5147d1d}, \texttt{\footnotesize f8e13ab4ae}, \texttt{\footnotesize f8eac0ad24}, \texttt{\footnotesize f97de2c3e9}, \texttt{\footnotesize faba6e97d7}, \texttt{\footnotesize faec2f0468}, \texttt{\footnotesize fb152519ad}, \texttt{\footnotesize fb564c935d}, \texttt{\footnotesize fb893ffaf3}, \texttt{\footnotesize fb9b4c2f15}, \texttt{\footnotesize fd361ab85f}, \texttt{\footnotesize fd8560cfd6}, \texttt{\footnotesize fe1733741f}, and \texttt{\footnotesize ff17657f71}.

\subsection{Custom Stereo Losses}
\label{subsec:custom_stereo_losses}

As mentioned in \Cref{subsec:confidence}, we adapt $\mathcal{L}_\text{NS}$ for each event stereo model -- \ie, SE-CFF~\cite{nam2022stereo}, EMatch~\cite{zhang2025ematch}, E-StereoAnywhere~\cite{bartolomei2025stereo}, and E-FoundationStereo~\cite{wen2025foundationstereo}.
In particular, we started from the original loss proposed by the authors of each architecture, obtaining the following losses:

\begin{itemize}
    \item We adapt $\mathcal{L}_\text{NS}$ for SE-CFF~\cite{nam2022stereo} starting from their multi-scale disparity loss:
    \begin{equation}
        \mathcal{L}'_\text{NS} = \sum^L_s w_s \left[ \left( \lambda_\text{disp} \cdot \eta(\mathbf{C}^{(s)}_\text{Vsize};\mu_\text{Vsize}) \cdot \mathcal{L}^{(s)}_\text{disp} + \mathbf{M}^{(s)}_\text{auto} \cdot \lambda_\text{3p} \cdot (1-\eta(\mathbf{C}^{(s)}_\text{Vsize};\mu_\text{Vsize})) \cdot \mathcal{L}^{(s)}_\text{3p} \right) + \lambda_\text{smooth}\cdot\mathcal{L}^{(s)}_\text{smooth} \right],
        \label{eq:secff_loss}
    \end{equation}
    where $L$ is the number of scales, $w_s$ is the weight for the $s$-th scale, $\mathcal{L}^{(s)}_\text{disp}$ is a L1 loss computed at scale $s$, $\mathcal{L}^{(s)}_\text{smooth}$ is a gradient regularization term that ensure smooth disparity estimations, and $\lambda_\text{smooth}=0.1$ is the weighting term for $\mathcal{L}^{(s)}_\text{smooth}$.
    \item For other stereo networks -- \ie, EMatch~\cite{zhang2025ematch}, E-StereoAnywhere~\cite{bartolomei2025stereo}, and E-FoundationStereo~\cite{wen2025foundationstereo} -- we adopt a RAFTStereo-like~\cite{lipson2021raft} loss with further supervision for the initial disparity estimation:
    \begin{equation}
        \mathcal{L}''_\text{NS} = \left[\sum^N_i w_i \left[ \left( \lambda_\text{disp} \cdot \eta(\mathbf{C}_\text{Vsize};\mu_\text{Vsize}) \cdot \mathcal{L}^{i}_\text{disp} + \mathbf{M}_\text{auto} \cdot \lambda_\text{3p} \cdot (1-\eta(\mathbf{C}_\text{Vsize};\mu_\text{Vsize})) \cdot \mathcal{L}^{i}_\text{3p} \right) \right]\right] + \mathcal{L}^0_\text{NS},
        \label{eq:others_loss}
    \end{equation}
    where $N$ is the number of refinement steps, $w_i$ is the exponentially increasing weight for the $i$-th refined disparity, $\mathcal{L}^{i}_\text{disp}$ is a L1 loss computed with respect to the $i$-th refined disparity, and $\mathcal{L}^0_\text{NS}$ is the NeRF-supervised loss for the initial disparity.
\end{itemize}
As mentioned in the main paper (\cref{subsec:implementation_details}), the losses $\mathcal{L}'_\text{NS}$ and $\mathcal{L}''_\text{NS}$ are used for NVS data only -- where $\mathbf{C}_\text{Vsze}$, and the RGB triplet $\mathbf{I}_{LL}$, $\mathbf{I}_{L}$, and $\mathbf{I}_{R}$ are available.
For the other sources of data -- \ie, distilled data from \cite{gehrig2021dsec}, and ground-truth supervised trainings -- we maintain only the $\mathcal{L}^{(s)}_\text{disp}$ and $\mathcal{L}^{i}_\text{disp}$ terms respectively from $\mathcal{L}'_\text{NS}$ and $\mathcal{L}''_\text{NS}$.

\section{Additional Qualitative Results}

In this section, we collect additional qualitative results, including full samples from the EventHub data (\cref{subsec:eventhub_qualitatives}), predictions generated by event-based stereo networks (\cref{subsec:predictions_qualitatives}), and finally, a qualitative comparison between conventional RGB Stereo Foundation Models like FoundationStereo before and after fine-tuning on EventHub data against challenging night sequences (\cref{subsec:predictions_sfm_night_qualitatives}).

\subsection{Qualitative Samples from EventHub}
\label{subsec:eventhub_qualitatives}

We report a few training samples generated with our EventHub pipeline, obtained both by means of cross-modal distillation and by deploying novel view synthesis.

\Cref{fig:supp_dsec} shows three samples from the DSEC datasets, obtained through the former paradigm. 
From left to right, we display the left image from the color stereo pair, the left event frame, and the proxy disparity map generated by FoundationStereo \cite{wen2025foundationstereo} and projected over the event frame, as described in \cref{sec:method:distillation}. 
We can notice, in particular, the high level of detail of these predicted labels, crucial for providing the event stereo models with strong guidance.

\Cref{fig:supp_nds,fig:supp_scan} collect four examples from scenes available in the NeRFStereo \cite{tosi2023nerf} and ScanNet++ \cite{yeshwanth2023scannet} datasets, respectively. 
From left to right, we show rendered RGB and event frames, followed by rendered depth maps, confidence maps based on voxel sizes, and rendered depth maps masked according to confidence thresholding. 
The latter further highlight the importance of confidence thresholding in removing outliers in the rendered depth maps.

\begin{figure}[t]
    \centering
    \renewcommand{\tabcolsep}{1pt}
    \begin{tabular}{ccc}
         {\setlength{\fboxsep}{0pt}\setlength{\fboxrule}{1pt}\fbox{\includegraphics[width=0.32\linewidth]{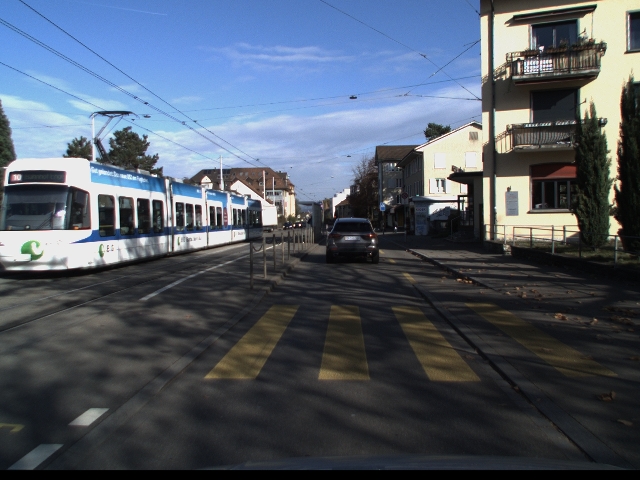}}} &
         {\setlength{\fboxsep}{0pt}\setlength{\fboxrule}{1pt}\fbox{\includegraphics[width=0.32\linewidth]{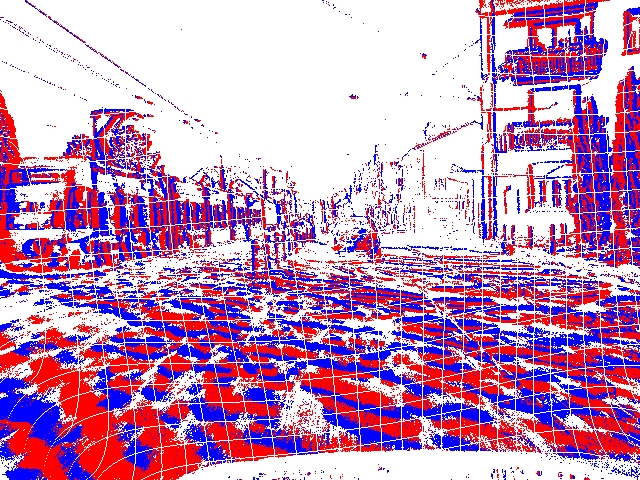}}} &
         {\setlength{\fboxsep}{0pt}\setlength{\fboxrule}{1pt}\fbox{\includegraphics[width=0.32\linewidth]{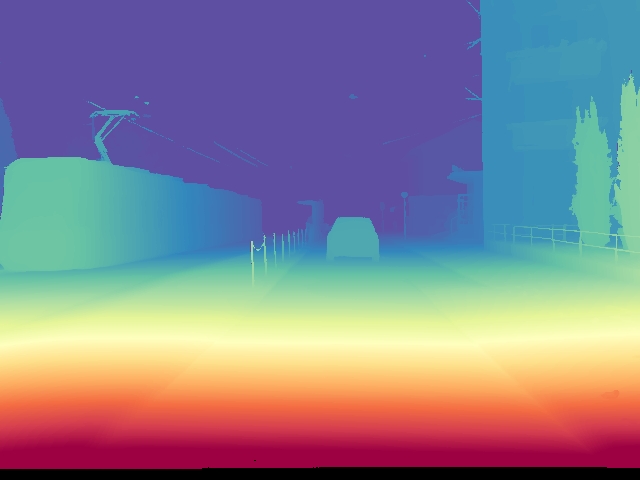}}} \\[-0.5pt]

         {\setlength{\fboxsep}{0pt}\setlength{\fboxrule}{1pt}\fbox{\includegraphics[width=0.32\linewidth]{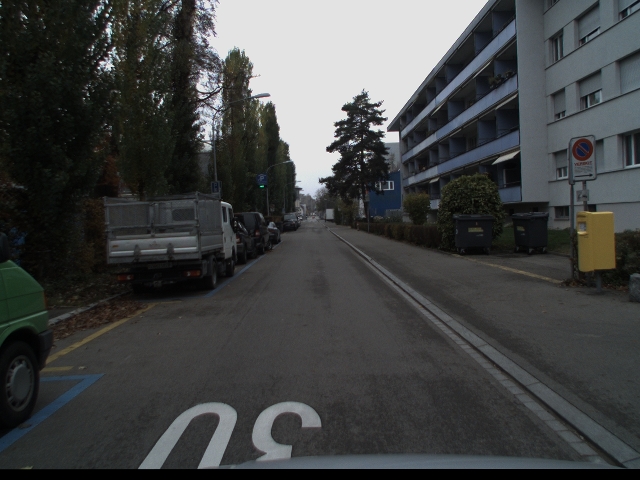}}} &
         {\setlength{\fboxsep}{0pt}\setlength{\fboxrule}{1pt}\fbox{\includegraphics[width=0.32\linewidth]{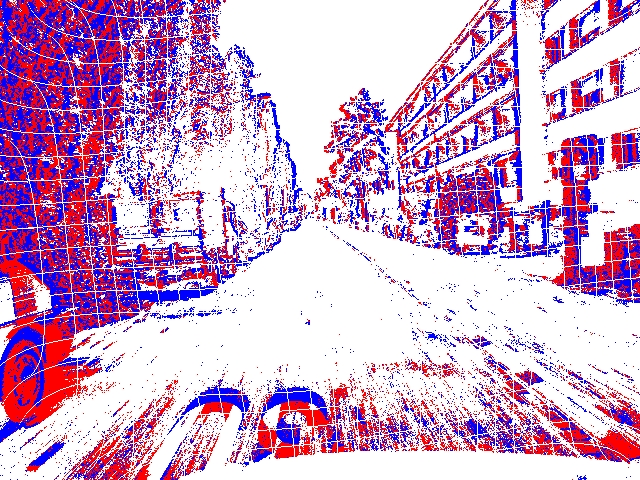}}} &
         {\setlength{\fboxsep}{0pt}\setlength{\fboxrule}{1pt}\fbox{\includegraphics[width=0.32\linewidth]{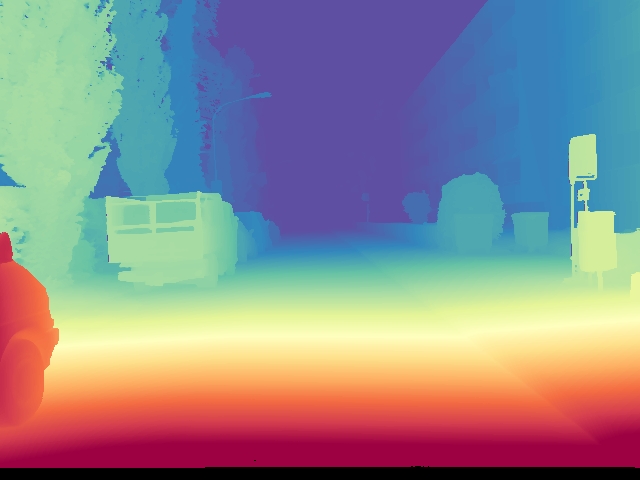}}} \\[-0.5pt]

         {\setlength{\fboxsep}{0pt}\setlength{\fboxrule}{1pt}\fbox{\includegraphics[width=0.32\linewidth]{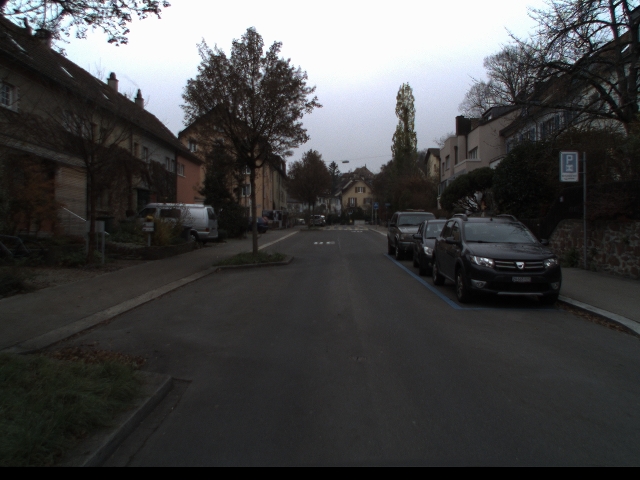}}} &
         {\setlength{\fboxsep}{0pt}\setlength{\fboxrule}{1pt}\fbox{\includegraphics[width=0.32\linewidth]{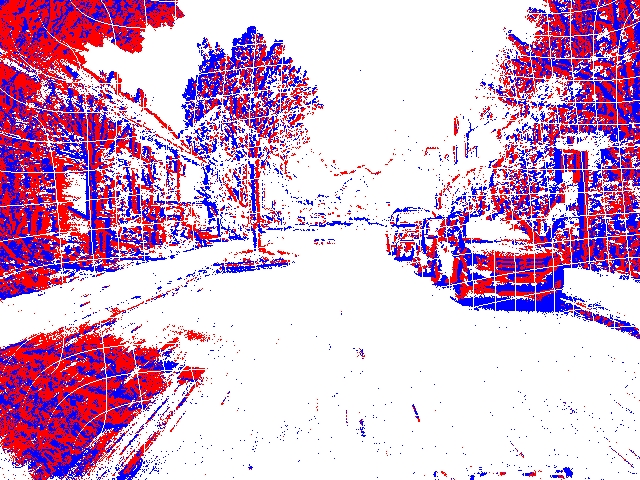}}} &
         {\setlength{\fboxsep}{0pt}\setlength{\fboxrule}{1pt}\fbox{\includegraphics[width=0.32\linewidth]{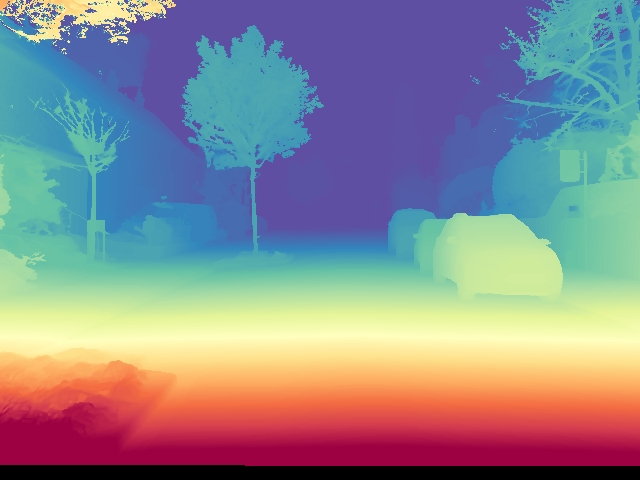}}} \\
         (a) & (b) & (c) \\
    \end{tabular}\vspace{-0.3cm}
    \caption{\textbf{Qualitative examples of training data generated by EventHub on DSEC \cite{gehrig2021dsec}.} (a) left color image, (b) left event frame, (c) proxy disparity labels predicted by FoundationStereo \cite{wen2025foundationstereo} and projected on the event frame.}
    \label{fig:supp_dsec}
\end{figure}

\begin{figure}[!htb]
    \centering
    \renewcommand{\tabcolsep}{1pt}
    \begin{tabular}{ccccc}
         {\setlength{\fboxsep}{0pt}\setlength{\fboxrule}{1pt}\fbox{\includegraphics[width=0.165\linewidth]{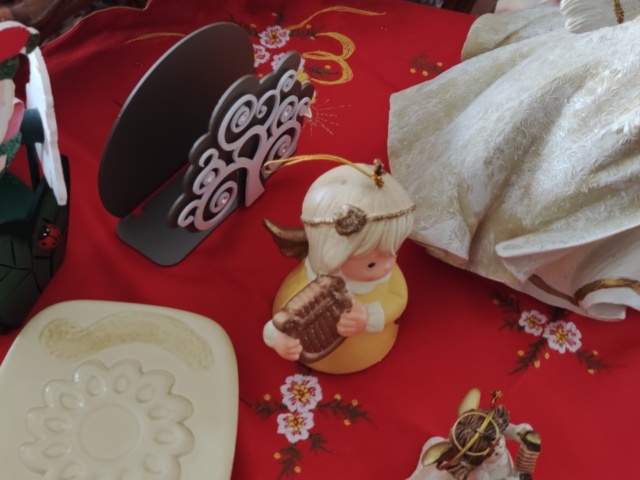}}} &
         {\setlength{\fboxsep}{0pt}\setlength{\fboxrule}{1pt}\fbox{\includegraphics[width=0.165\linewidth]{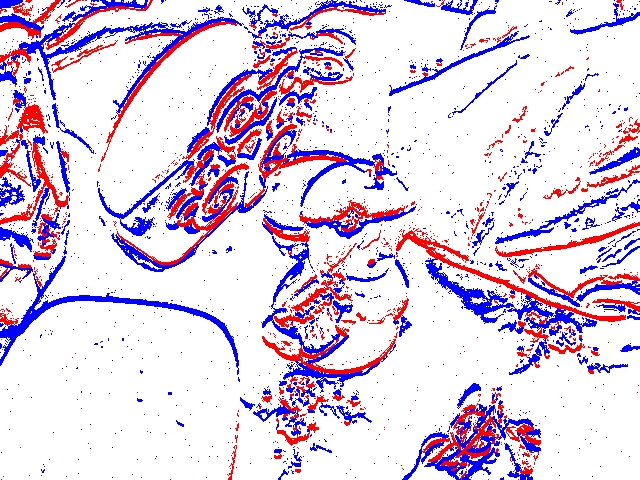}}} &
         {\setlength{\fboxsep}{0pt}\setlength{\fboxrule}{1pt}\fbox{\includegraphics[width=0.165\linewidth]{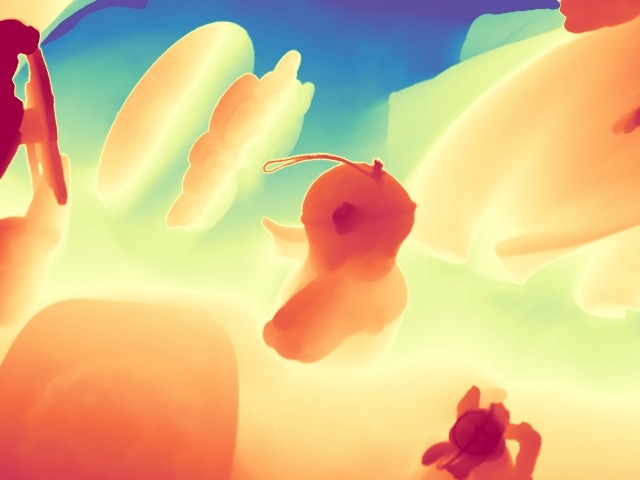}}} &
         {\setlength{\fboxsep}{0pt}\setlength{\fboxrule}{1pt}\fbox{\includegraphics[width=0.165\linewidth]{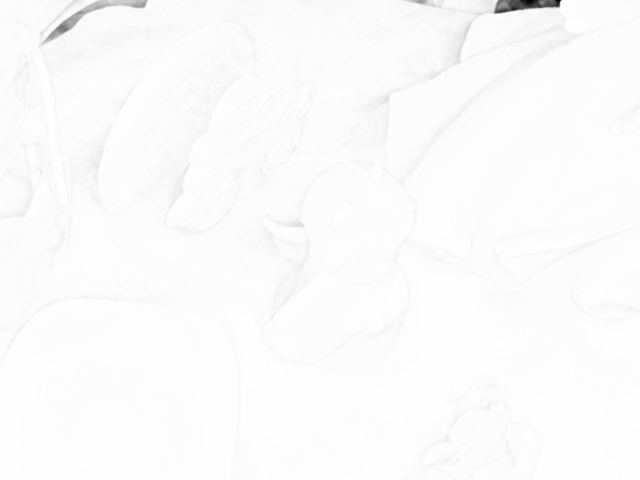}}} &
         {\setlength{\fboxsep}{0pt}\setlength{\fboxrule}{1pt}\fbox{\includegraphics[width=0.165\linewidth]{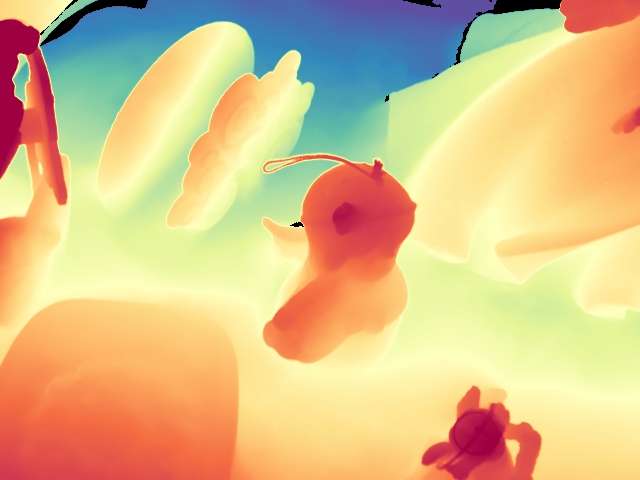}}} \\[-0.5pt]

         {\setlength{\fboxsep}{0pt}\setlength{\fboxrule}{1pt}\fbox{\includegraphics[width=0.165\linewidth]{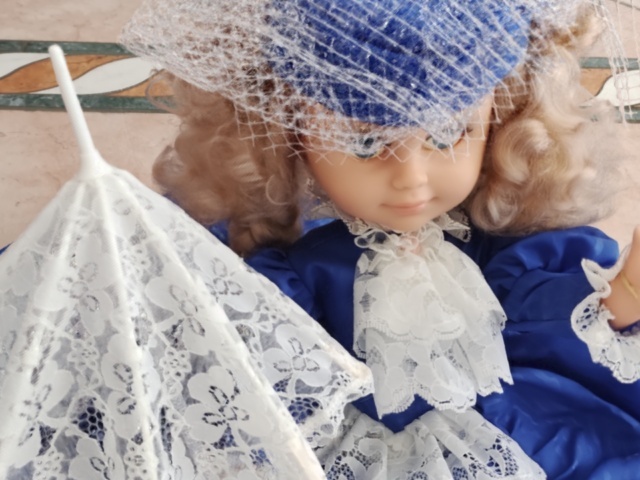}}} &
         {\setlength{\fboxsep}{0pt}\setlength{\fboxrule}{1pt}\fbox{\includegraphics[width=0.165\linewidth]{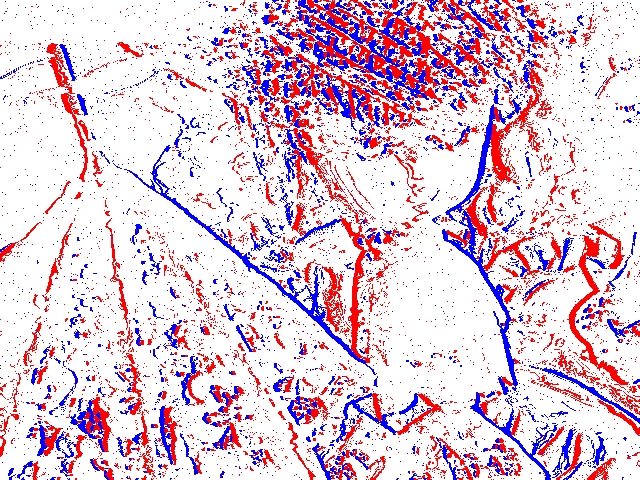}}} &
         {\setlength{\fboxsep}{0pt}\setlength{\fboxrule}{1pt}\fbox{\includegraphics[width=0.165\linewidth]{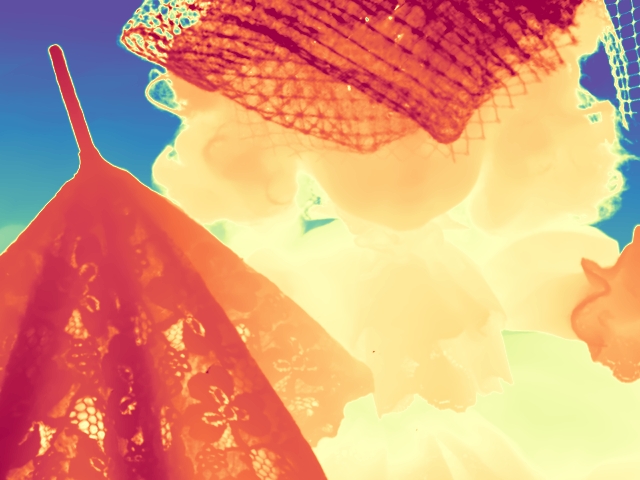}}} &
         {\setlength{\fboxsep}{0pt}\setlength{\fboxrule}{1pt}\fbox{\includegraphics[width=0.165\linewidth]{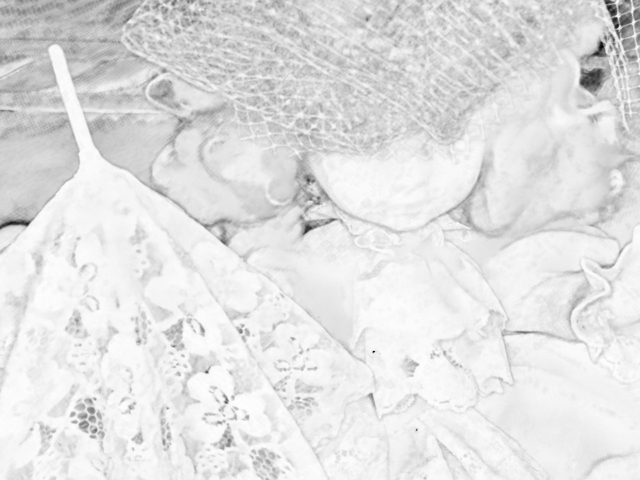}}} &
         {\setlength{\fboxsep}{0pt}\setlength{\fboxrule}{1pt}\fbox{\includegraphics[width=0.165\linewidth]{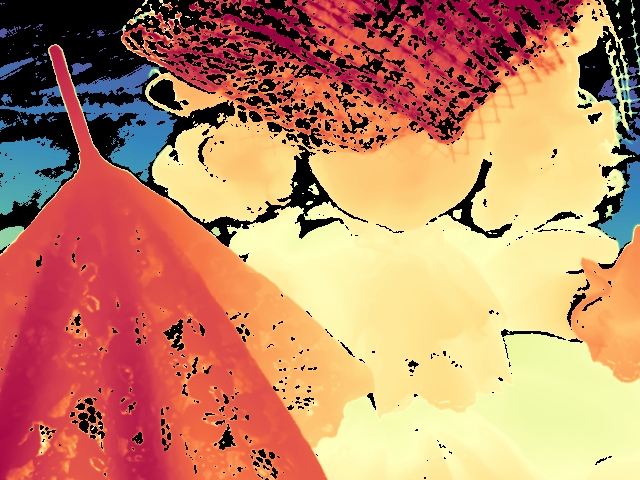}}} \\[-0.5pt]

         {\setlength{\fboxsep}{0pt}\setlength{\fboxrule}{1pt}\fbox{\includegraphics[width=0.165\linewidth]{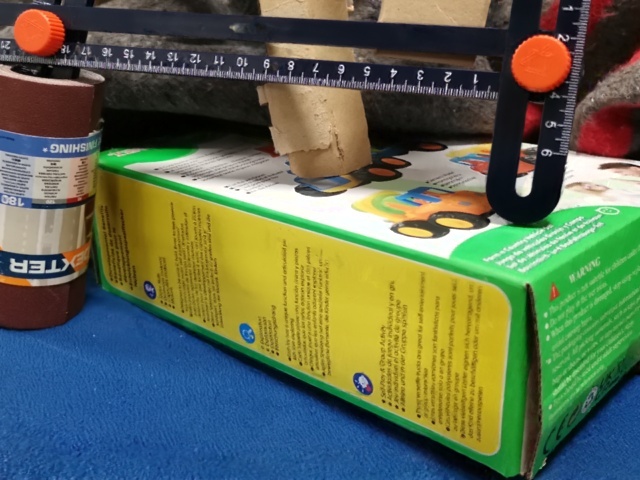}}} &
         {\setlength{\fboxsep}{0pt}\setlength{\fboxrule}{1pt}\fbox{\includegraphics[width=0.165\linewidth]{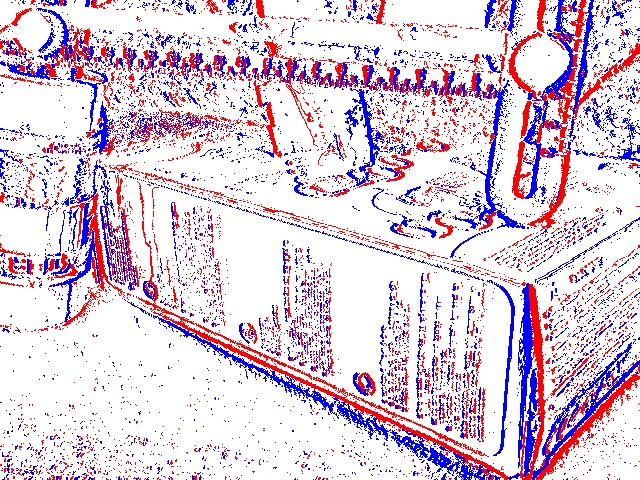}}} &
         {\setlength{\fboxsep}{0pt}\setlength{\fboxrule}{1pt}\fbox{\includegraphics[width=0.165\linewidth]{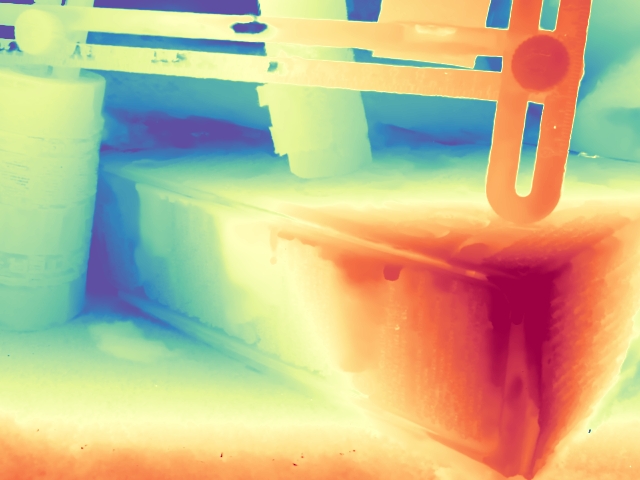}}} &
         {\setlength{\fboxsep}{0pt}\setlength{\fboxrule}{1pt}\fbox{\includegraphics[width=0.165\linewidth]{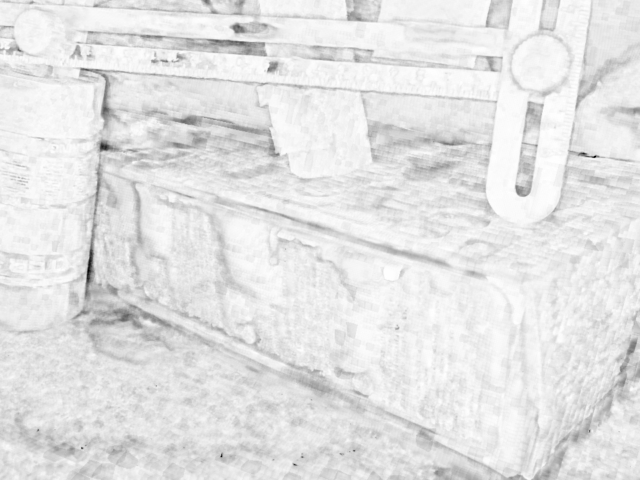}}} &
         {\setlength{\fboxsep}{0pt}\setlength{\fboxrule}{1pt}\fbox{\includegraphics[width=0.165\linewidth]{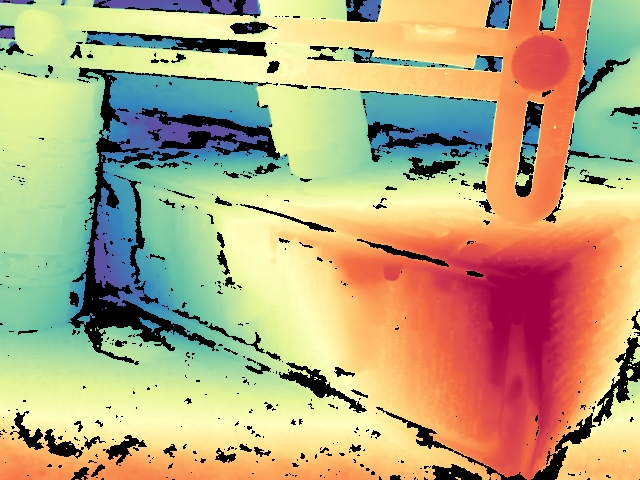}}} \\[-0.5pt]

         {\setlength{\fboxsep}{0pt}\setlength{\fboxrule}{1pt}\fbox{\includegraphics[width=0.165\linewidth]{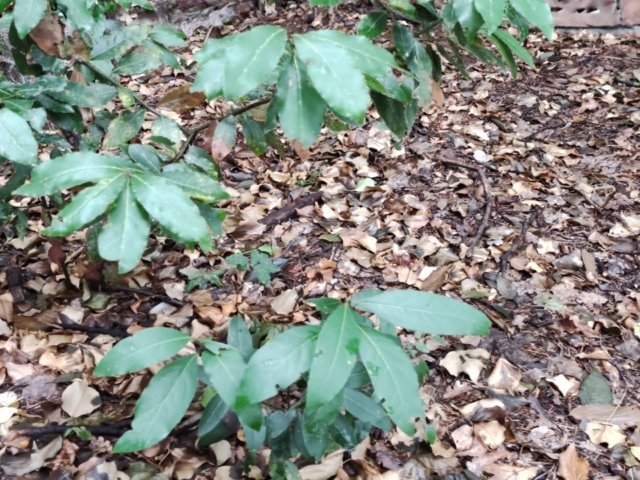}}} &
         {\setlength{\fboxsep}{0pt}\setlength{\fboxrule}{1pt}\fbox{\includegraphics[width=0.165\linewidth]{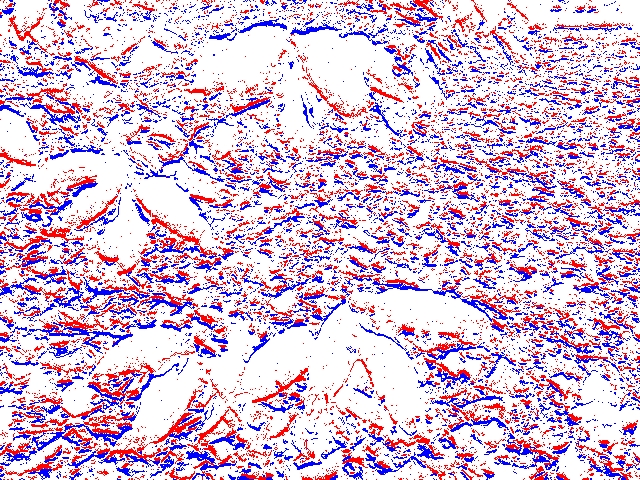}}} &
         {\setlength{\fboxsep}{0pt}\setlength{\fboxrule}{1pt}\fbox{\includegraphics[width=0.165\linewidth]{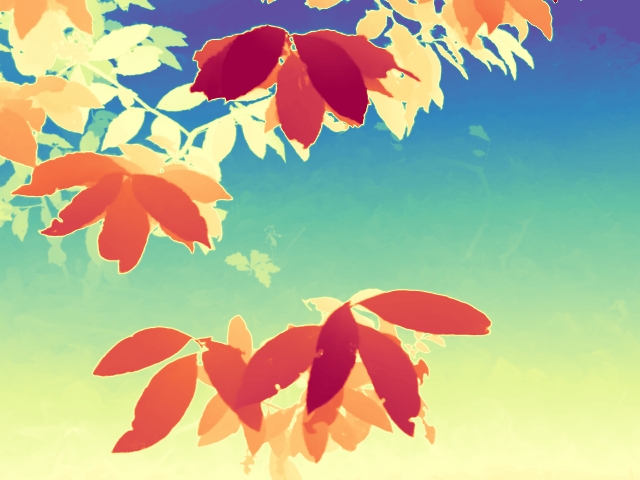}}} &
         {\setlength{\fboxsep}{0pt}\setlength{\fboxrule}{1pt}\fbox{\includegraphics[width=0.165\linewidth]{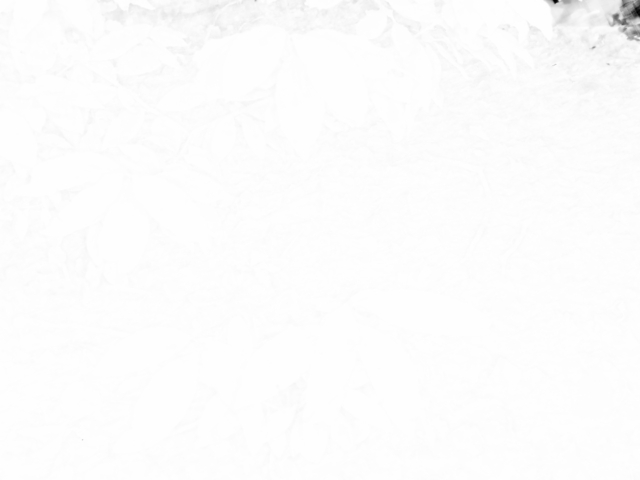}}} &
         {\setlength{\fboxsep}{0pt}\setlength{\fboxrule}{1pt}\fbox{\includegraphics[width=0.165\linewidth]{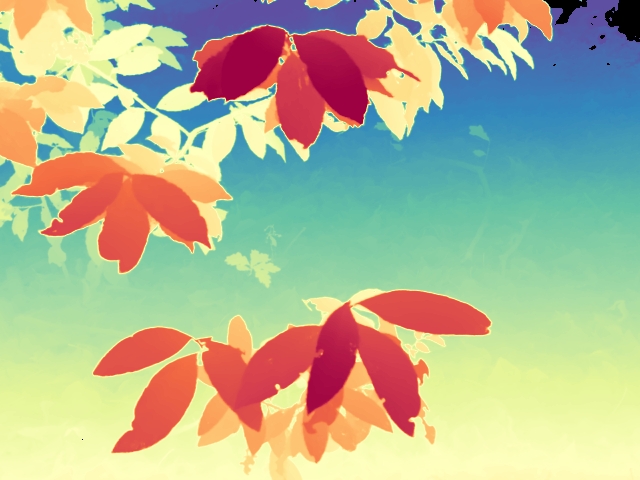}}} \\
         (a) & (b) & (c) & (d) & (e) \\
    \end{tabular}\vspace{-0.3cm}
    \caption{\textbf{Qualitative examples of training data generated by EventHub on NeRFStereo \cite{tosi2023nerf}.} 
    (a) rendered color image, (b) rendered event frame, (c) rendered depth and (d) confidence (the brighter, the higher the confidence in the estimated depth values), and (e) rendered depth masked according to confidence thresholding.}
    \label{fig:supp_nds}
\end{figure}

\begin{figure}[!htb]
    \centering
    \renewcommand{\tabcolsep}{1pt}
    \begin{tabular}{ccccc}
         {\setlength{\fboxsep}{0pt}\setlength{\fboxrule}{1pt}\fbox{\includegraphics[width=0.165\linewidth]{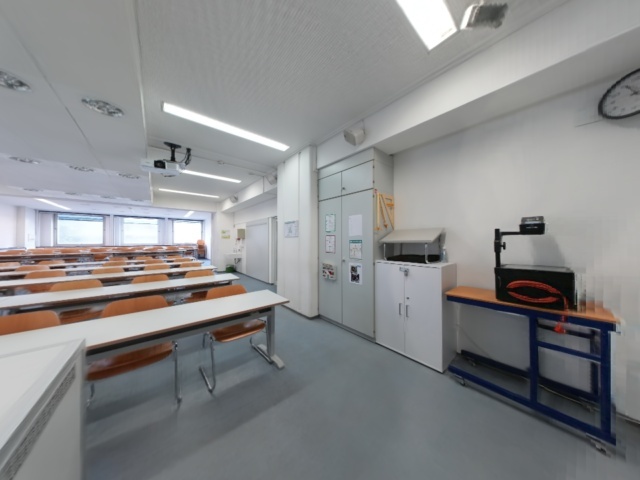}}} &
         {\setlength{\fboxsep}{0pt}\setlength{\fboxrule}{1pt}\fbox{\includegraphics[width=0.165\linewidth]{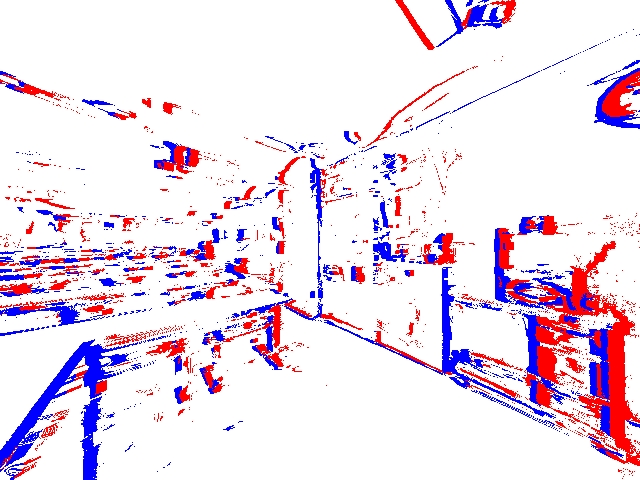}}} &
         {\setlength{\fboxsep}{0pt}\setlength{\fboxrule}{1pt}\fbox{\includegraphics[width=0.165\linewidth]{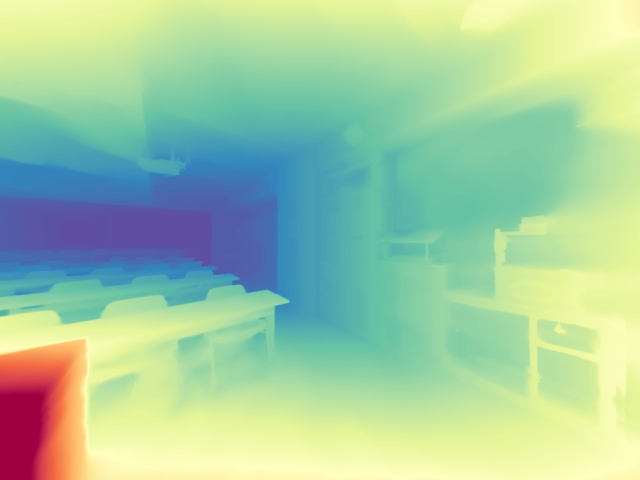}}} &
         {\setlength{\fboxsep}{0pt}\setlength{\fboxrule}{1pt}\fbox{\includegraphics[width=0.165\linewidth]{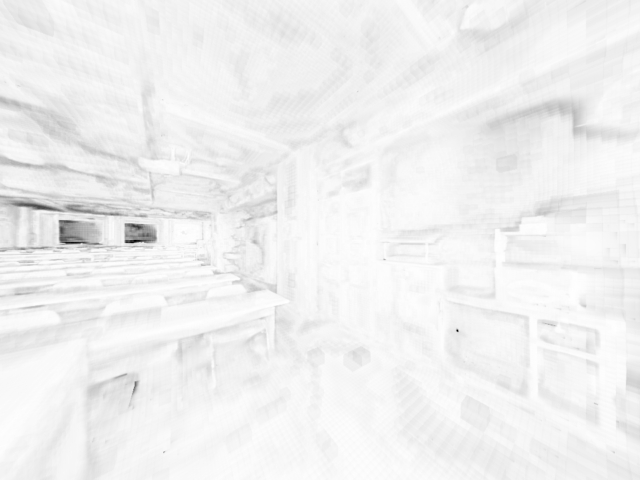}}} &
         {\setlength{\fboxsep}{0pt}\setlength{\fboxrule}{1pt}\fbox{\includegraphics[width=0.165\linewidth]{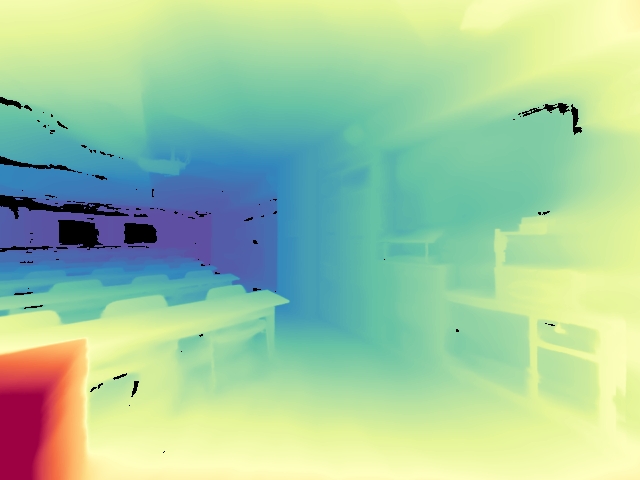}}} \\[-0.5pt]

         {\setlength{\fboxsep}{0pt}\setlength{\fboxrule}{1pt}\fbox{\includegraphics[width=0.165\linewidth]{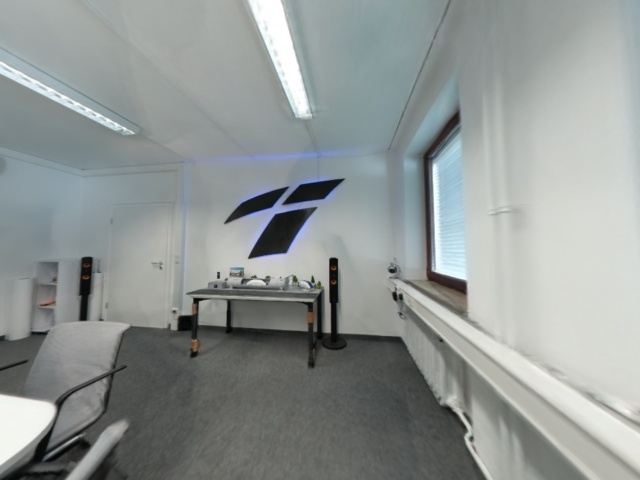}}} &
         {\setlength{\fboxsep}{0pt}\setlength{\fboxrule}{1pt}\fbox{\includegraphics[width=0.165\linewidth]{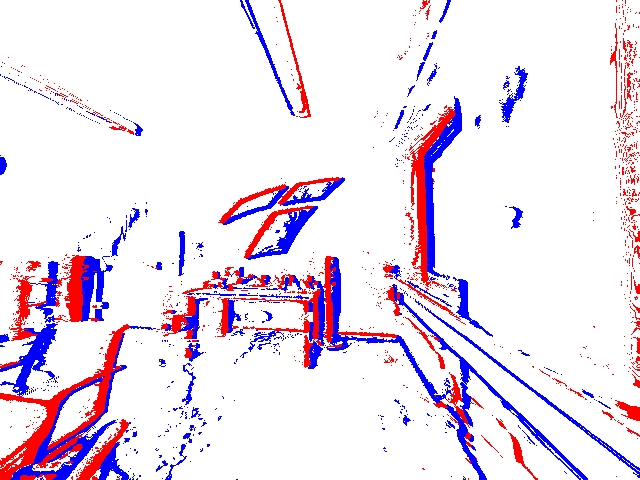}}} &
         {\setlength{\fboxsep}{0pt}\setlength{\fboxrule}{1pt}\fbox{\includegraphics[width=0.165\linewidth]{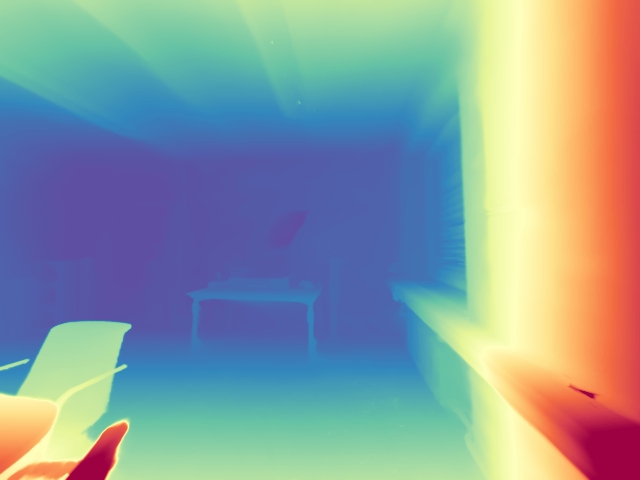}}} &
         {\setlength{\fboxsep}{0pt}\setlength{\fboxrule}{1pt}\fbox{\includegraphics[width=0.165\linewidth]{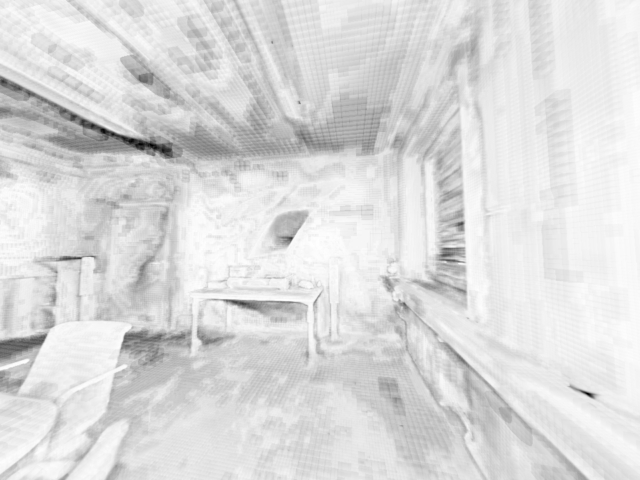}}} &
         {\setlength{\fboxsep}{0pt}\setlength{\fboxrule}{1pt}\fbox{\includegraphics[width=0.165\linewidth]{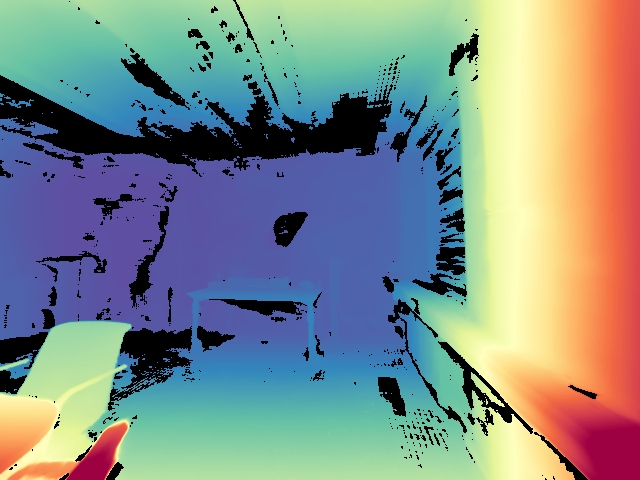}}}\\[-0.5pt]

         {\setlength{\fboxsep}{0pt}\setlength{\fboxrule}{1pt}\fbox{\includegraphics[width=0.165\linewidth]{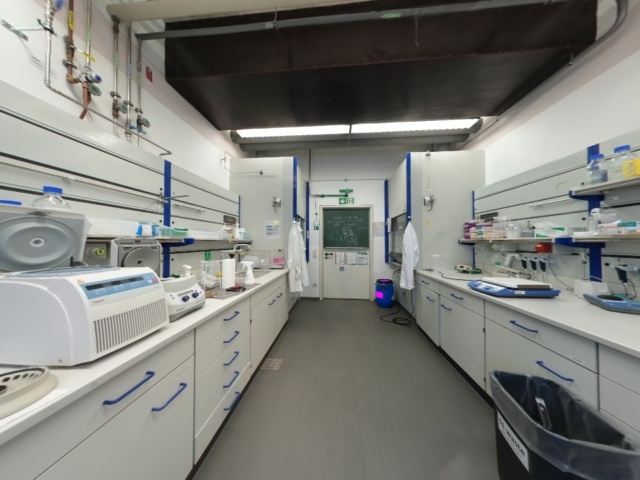}}} &
         {\setlength{\fboxsep}{0pt}\setlength{\fboxrule}{1pt}\fbox{\includegraphics[width=0.165\linewidth]{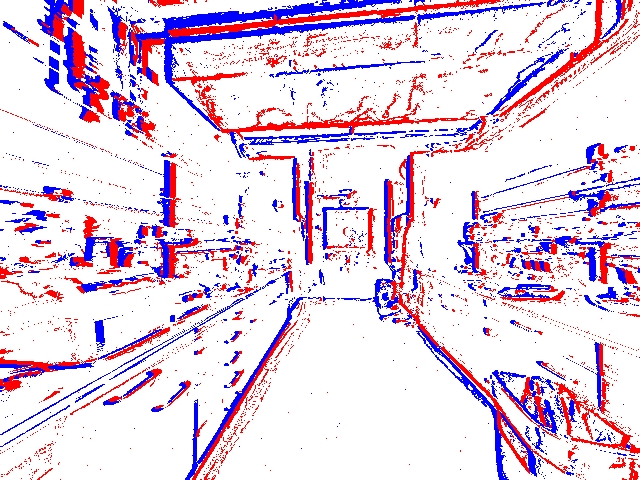}}} &
         {\setlength{\fboxsep}{0pt}\setlength{\fboxrule}{1pt}\fbox{\includegraphics[width=0.165\linewidth]{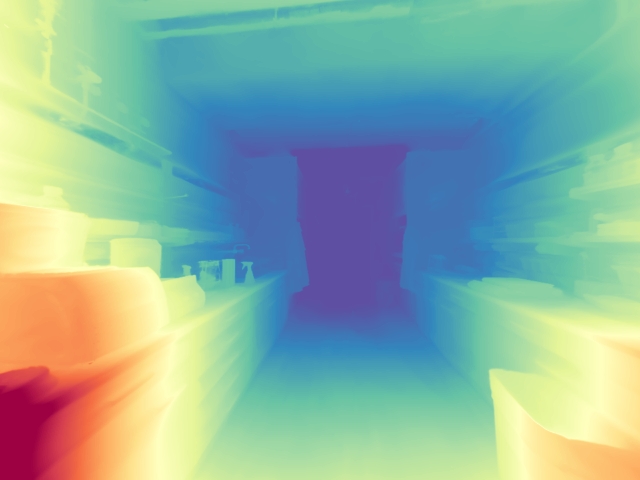}}} &
         {\setlength{\fboxsep}{0pt}\setlength{\fboxrule}{1pt}\fbox{\includegraphics[width=0.165\linewidth]{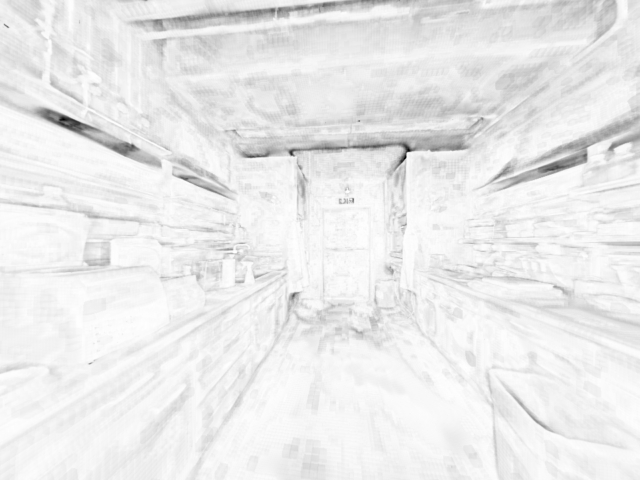}}} &
         {\setlength{\fboxsep}{0pt}\setlength{\fboxrule}{1pt}\fbox{\includegraphics[width=0.165\linewidth]{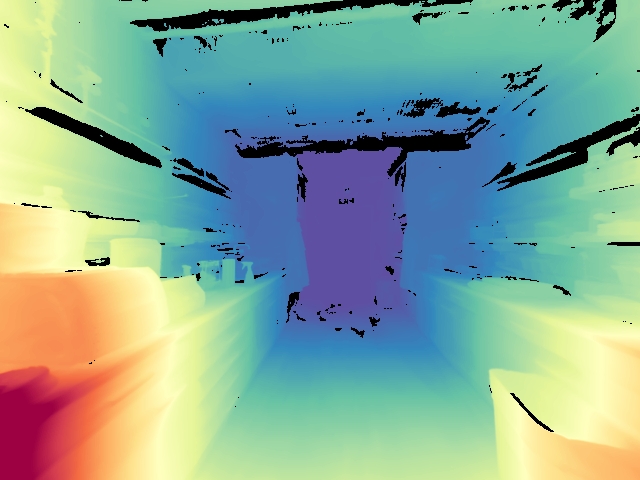}}} \\[-0.5pt]

         {\setlength{\fboxsep}{0pt}\setlength{\fboxrule}{1pt}\fbox{\includegraphics[width=0.165\linewidth]{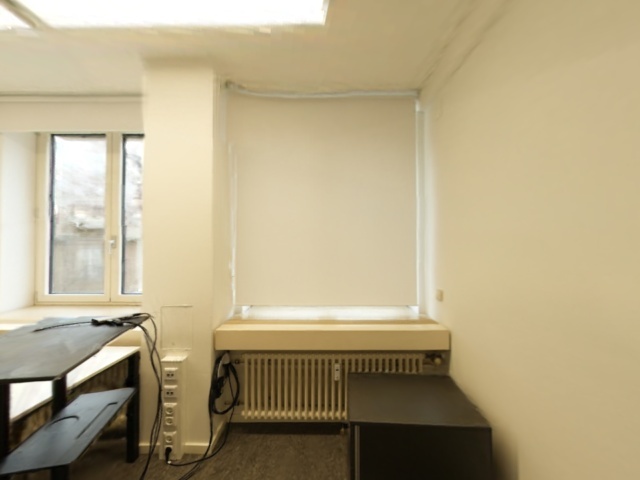}}} &
         {\setlength{\fboxsep}{0pt}\setlength{\fboxrule}{1pt}\fbox{\includegraphics[width=0.165\linewidth]{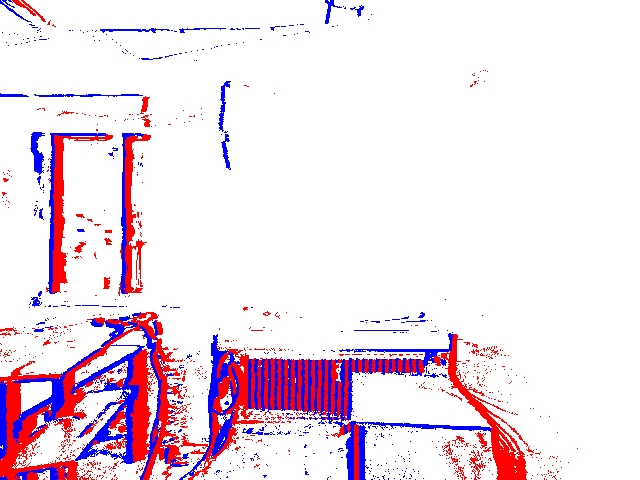}}} &
         {\setlength{\fboxsep}{0pt}\setlength{\fboxrule}{1pt}\fbox{\includegraphics[width=0.165\linewidth]{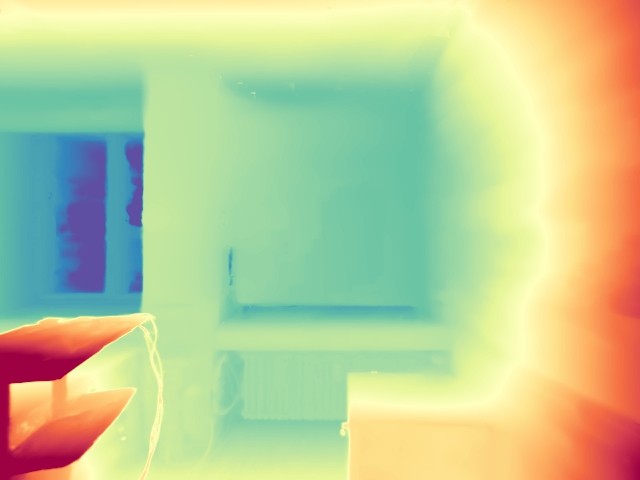}}} &
         {\setlength{\fboxsep}{0pt}\setlength{\fboxrule}{1pt}\fbox{\includegraphics[width=0.165\linewidth]{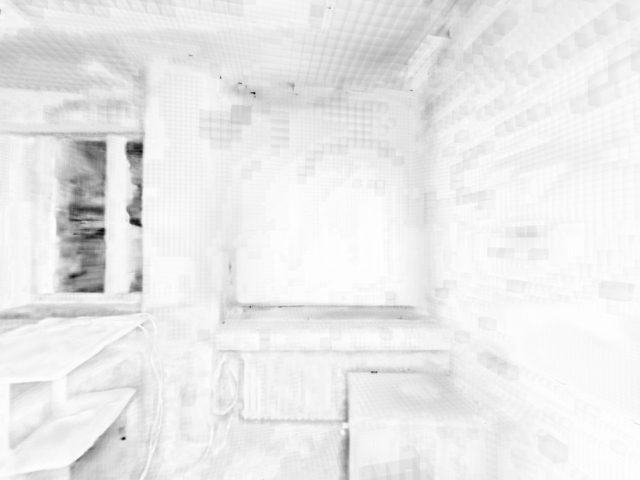}}} &
         {\setlength{\fboxsep}{0pt}\setlength{\fboxrule}{1pt}\fbox{\includegraphics[width=0.165\linewidth]{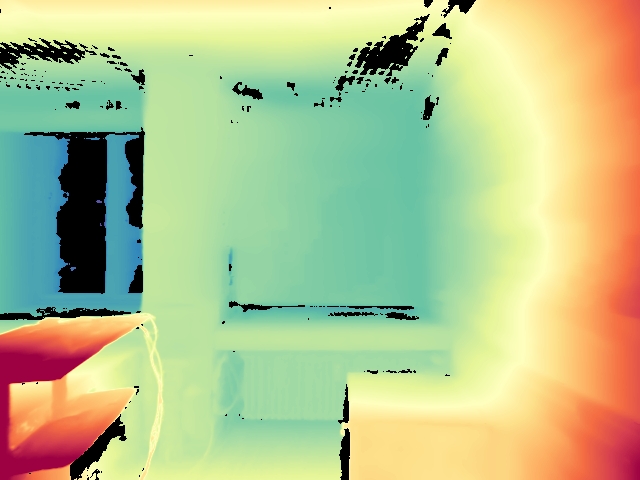}}} \\
        (a) & (b) & (c) & (d) & (e) \\
    \end{tabular}\vspace{-0.3cm}
    \caption{\textbf{Qualitative examples of training data generated by EventHub on ScanNet++ \cite{yeshwanth2023scannet}.} (a) rendered color image, (b) rendered event frame, (c) rendered depth and (d) confidence, (e) rendered depth masked according to confidence thresholding.}
    \label{fig:supp_scan}
\end{figure}

\subsection{Predictions from Event Stereo Networks}
\label{subsec:predictions_qualitatives}

We report additional qualitative results concerning event stereo models trained with different supervision flavors. 

\Cref{fig:qualitative_dsec_interlaken_00_g_00001,fig:qualitative_m3ed_outdoor_day_spot_outdoor_day_art_plaza_loop_00005,fig:qualitative_mvsec_outdoor_day_outdoor_day1_00015} collect two samples each, respectively, from DSEC \cite{gehrig2021dsec}, M3ED \cite{chaney2023m3ed} and MVSEC \cite{zhu2018multivehicle} datasets. 
On any dataset, we can clearly notice how MIX 4 allows for training any of the four models involved in our experiments at their best, with the novel models introduced by repurposing stereo foundation models from the RGB literature \cite{bartolomei2025stereo,wen2025foundationstereo} -- E-StereoAnywhere and E-FoundationStereo -- benefiting the most from the superior annotations produced by EventHub.

\begin{figure*}[t]
\centering
\renewcommand{\tabcolsep}{1pt}
\resizebox{1.0\textwidth}{!} {
	\begin{tabular}{cccccccc}
	\scriptsize\textbf{Events \& Ground Truth} &  & \scriptsize\textbf{SE-CFF}~\cite{nam2022stereo} & \scriptsize\textbf{EMatch}~\cite{zhang2025ematch} & \scriptsize\textbf{E-StereoAnywhere} & \scriptsize\textbf{E-FoundationStereo} \\
	\includegraphics[width=0.225\textwidth]{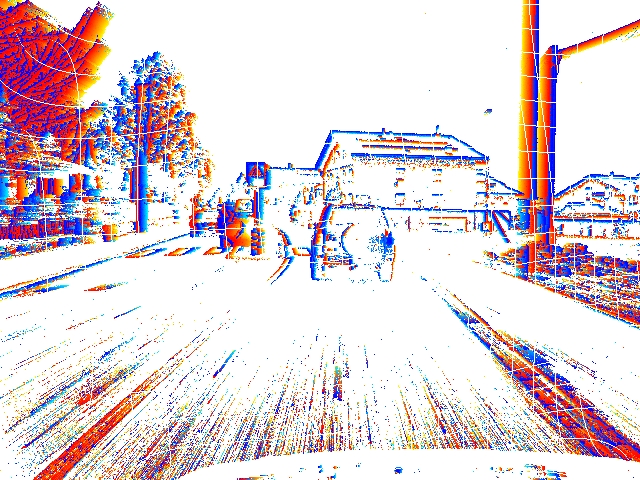} & \rotatebox[origin=l]{90}{\hspace{2em}\centering\scriptsize\textbf{LiDAR (GT)}} & \includegraphics[width=0.225\textwidth]{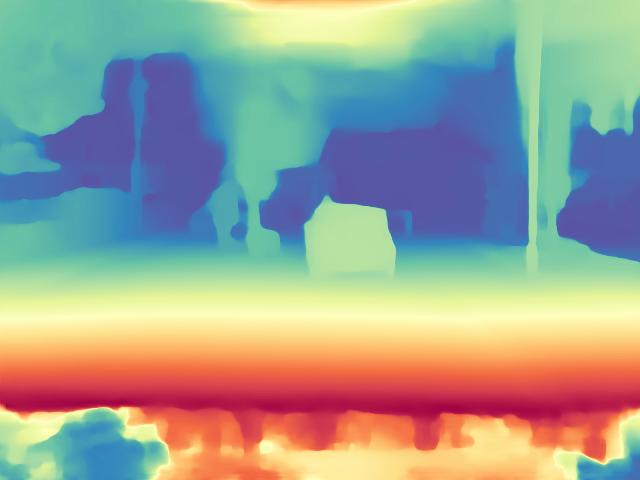} & \includegraphics[width=0.225\textwidth]{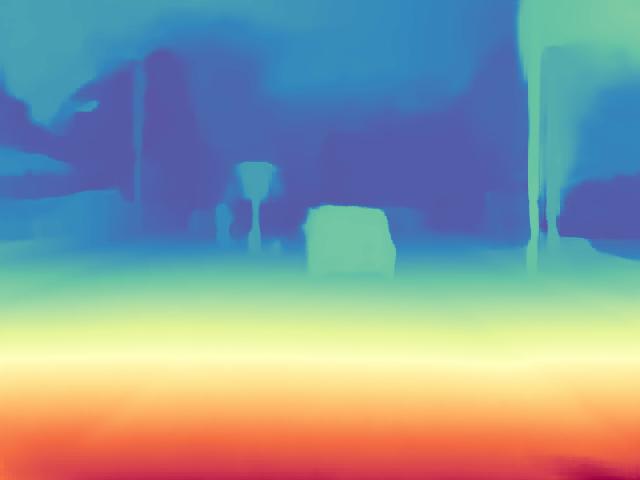} & \includegraphics[width=0.225\textwidth]{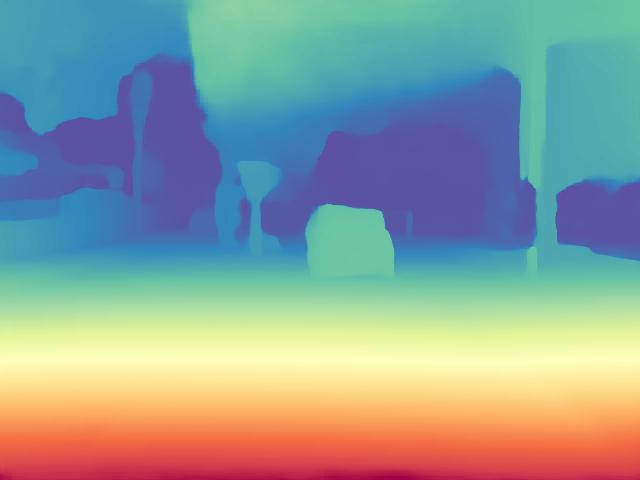} & \includegraphics[width=0.225\textwidth]{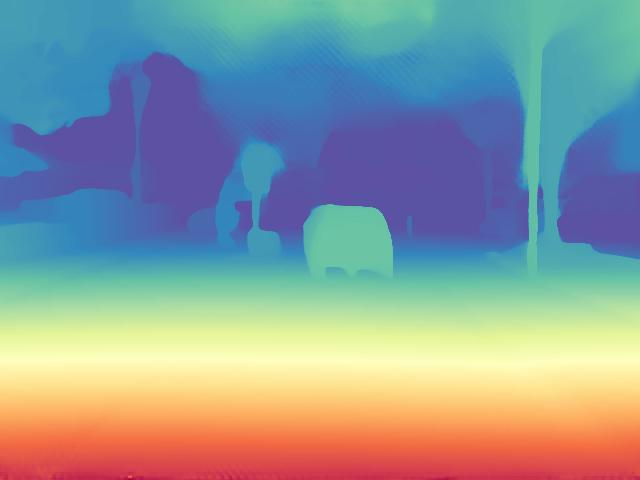} \\[-1pt]
	\includegraphics[width=0.225\textwidth]{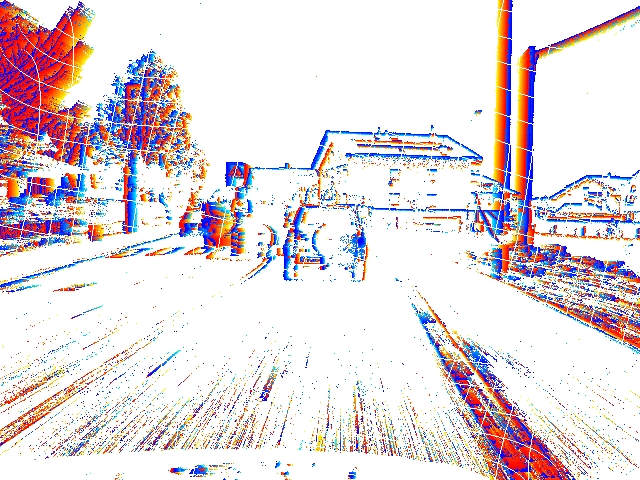} & \rotatebox[origin=l]{90}{\hspace{3em}\centering\scriptsize\textbf{MIX 3}} & \includegraphics[width=0.225\textwidth]{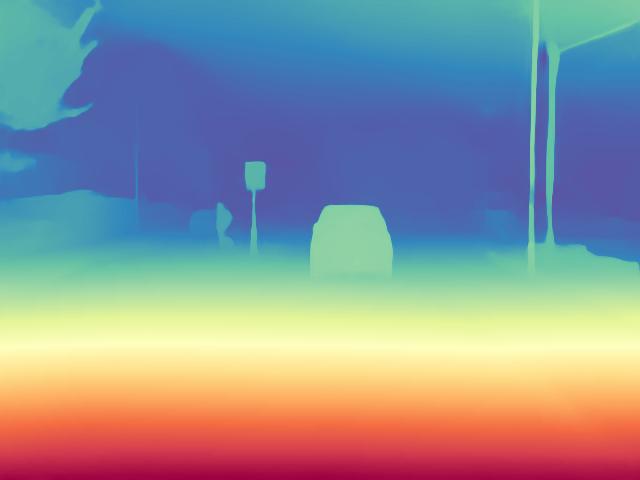} & \includegraphics[width=0.225\textwidth]{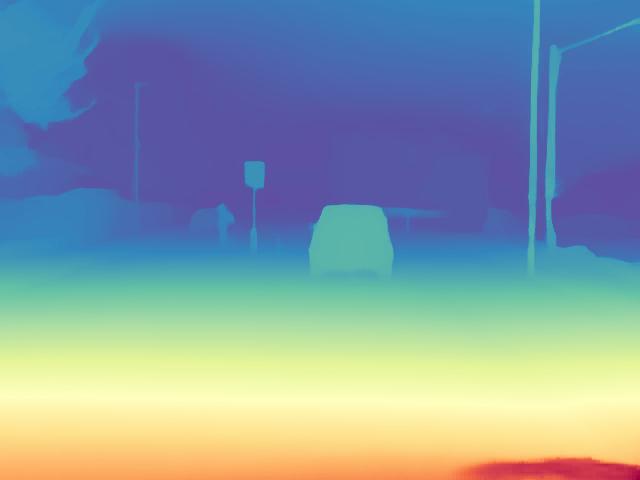} & \includegraphics[width=0.225\textwidth]{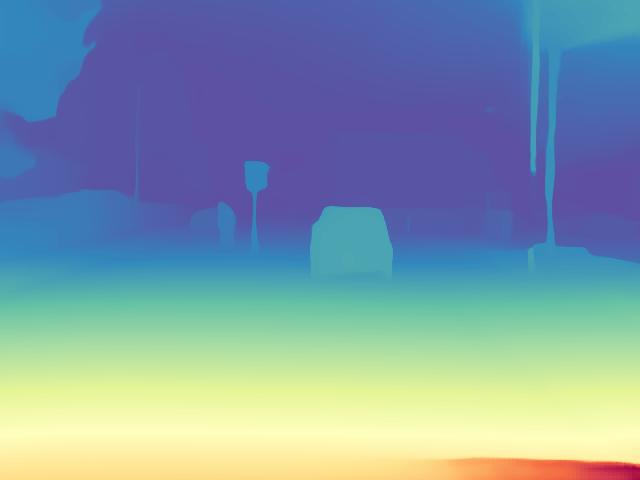} & \includegraphics[width=0.225\textwidth]{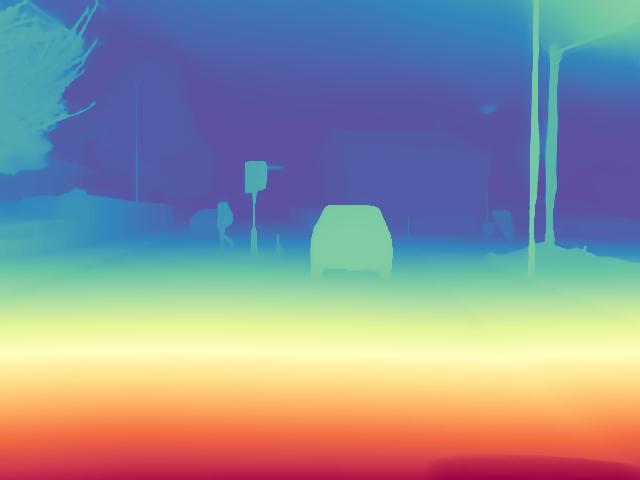} \\[-1pt]
	\includegraphics[width=0.225\textwidth]{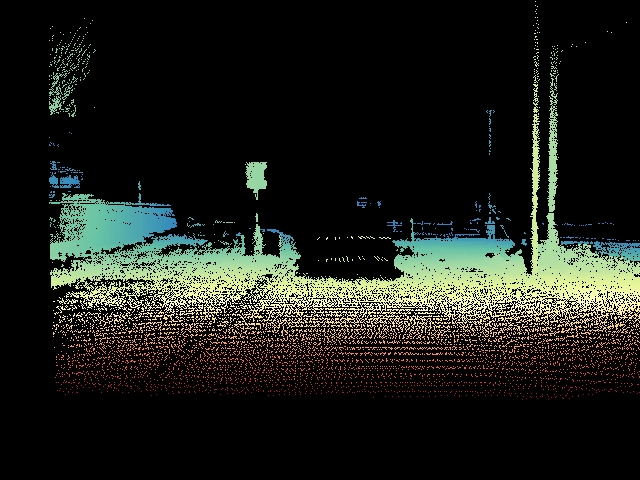} & \rotatebox[origin=l]{90}{\hspace{3em}\centering\scriptsize\textbf{MIX 4}} & \includegraphics[width=0.225\textwidth]{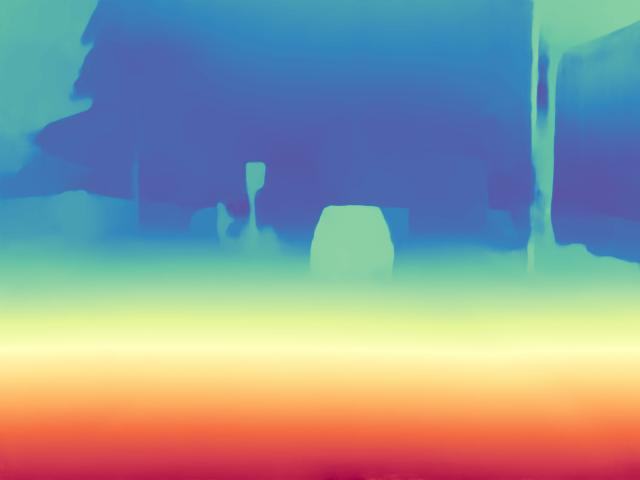} & \includegraphics[width=0.225\textwidth]{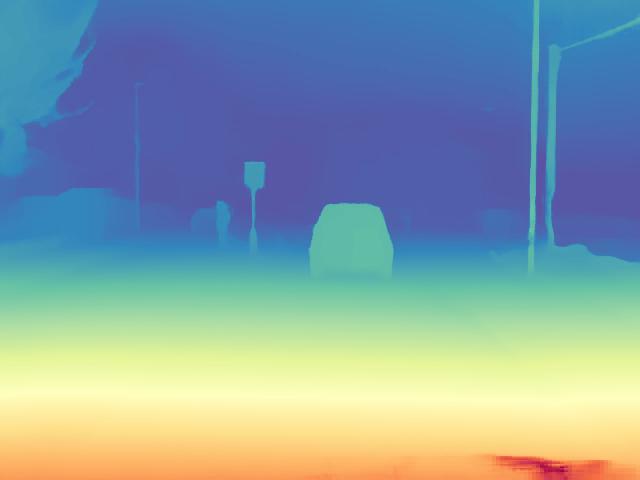} & \includegraphics[width=0.225\textwidth]{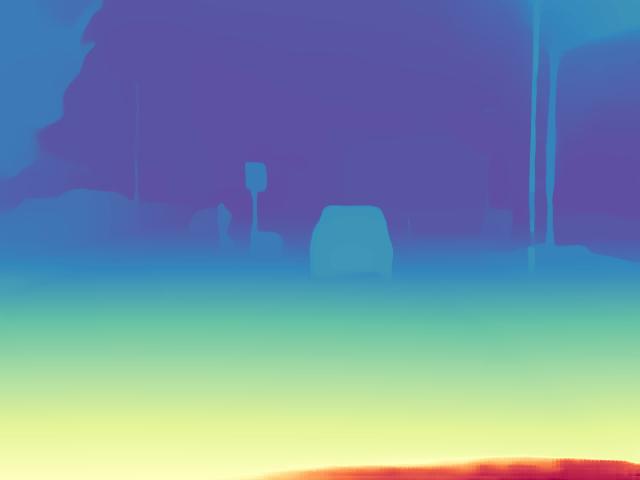} & \includegraphics[width=0.225\textwidth]{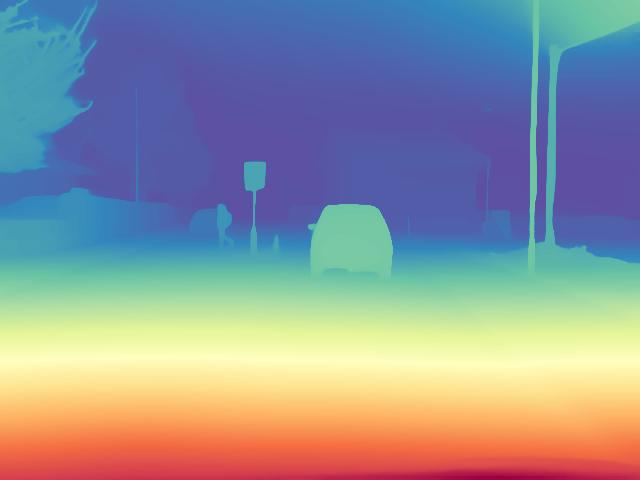} \\
	\end{tabular}
}\vspace{-0.3cm}
\caption{\textbf{Qualitative results on DSEC \cite{gehrig2021dsec} dataset.} Predictions by the four models trained with LiDAR labels, MIX 3 or MIX~4.}\vspace{-0.3cm}
\label{fig:qualitative_dsec_interlaken_00_g_00001}
\end{figure*}

\begin{figure*}[t]
\centering
\renewcommand{\tabcolsep}{1pt}
\resizebox{1.0\textwidth}{!} {
	\begin{tabular}{cccccccc}
	\scriptsize\textbf{Events \& Ground Truth} &  & \scriptsize\textbf{SE-CFF}~\cite{nam2022stereo} & \scriptsize\textbf{EMatch}~\cite{zhang2025ematch} & \scriptsize\textbf{E-StereoAnywhere} & \scriptsize\textbf{E-FoundationStereo} \\
	\includegraphics[width=0.225\textwidth]{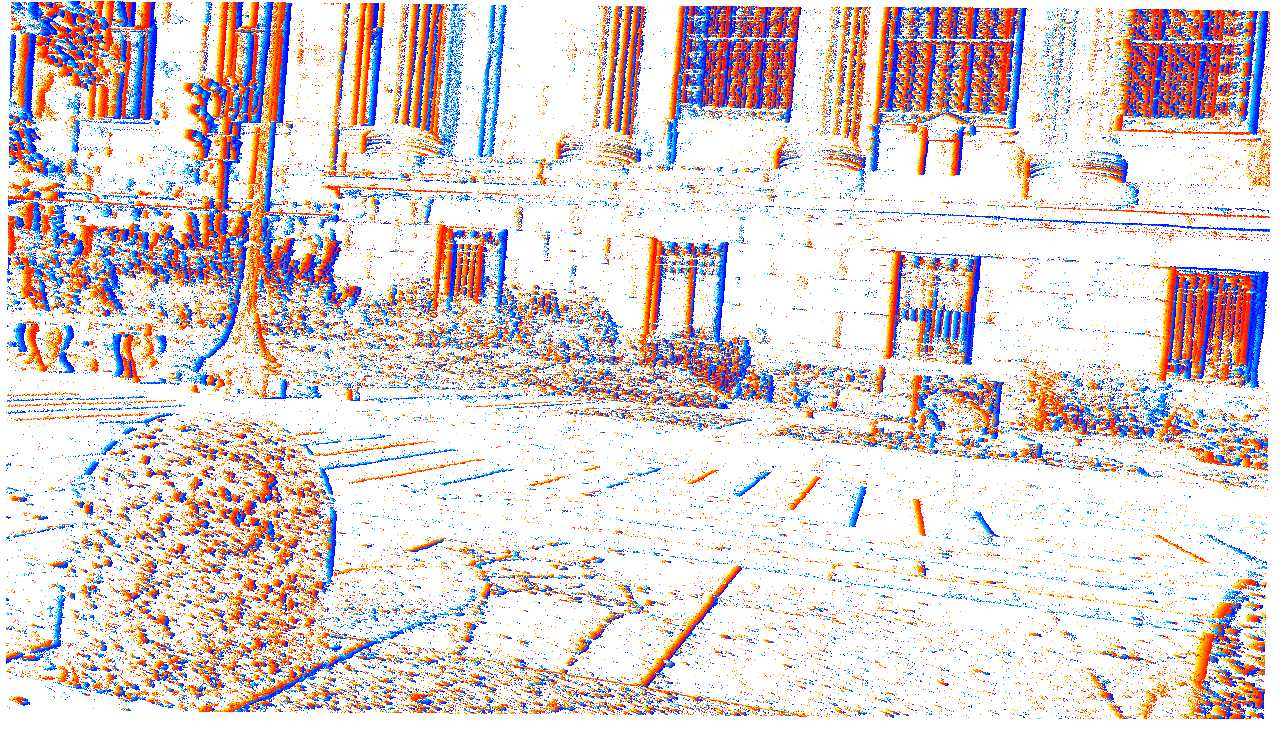} & \rotatebox[origin=l]{90}{\hspace{1.5em}\centering\scriptsize\textbf{LiDAR (GT)}} & \includegraphics[width=0.225\textwidth]{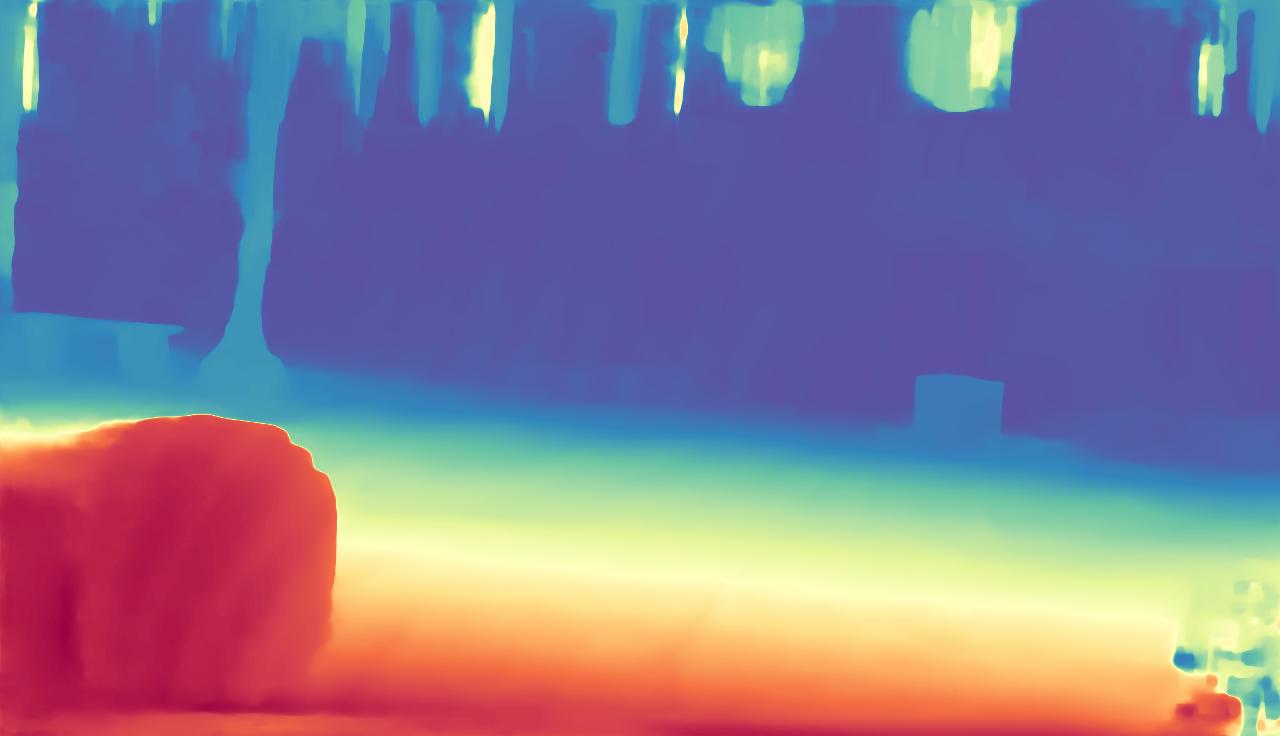} & \includegraphics[width=0.225\textwidth]{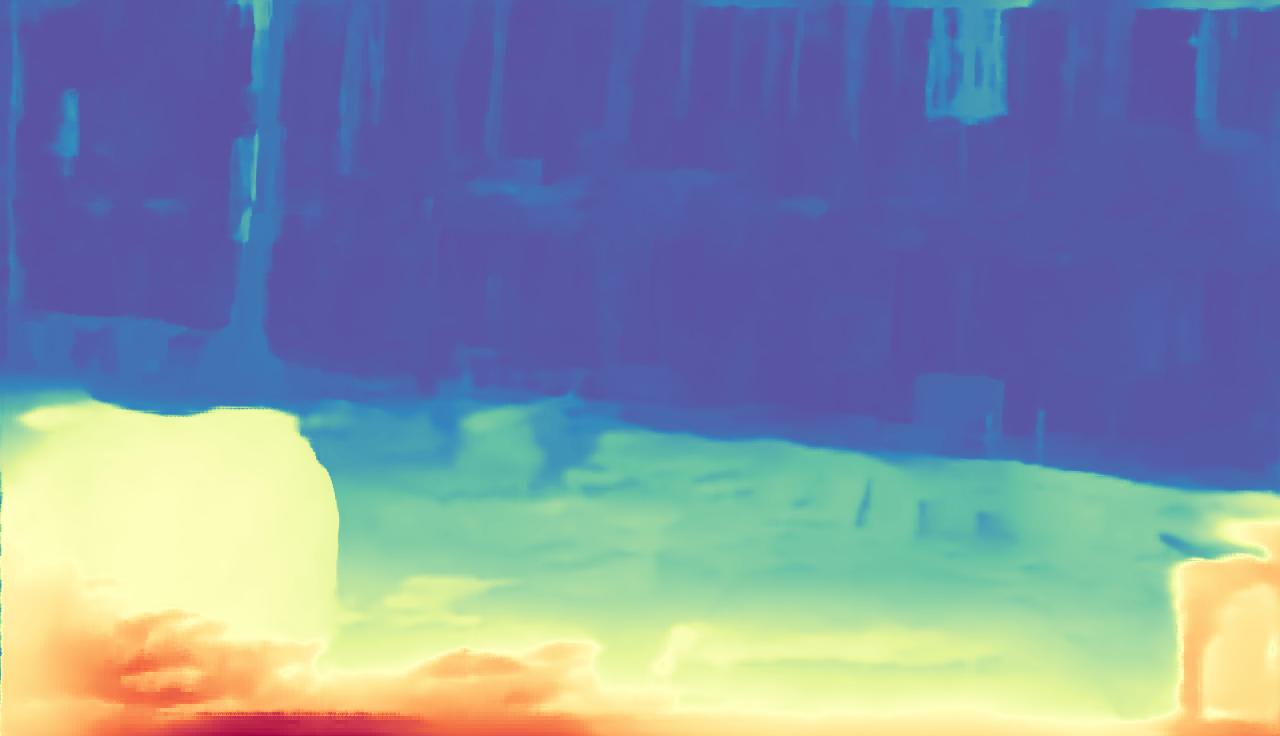} & \includegraphics[width=0.225\textwidth]{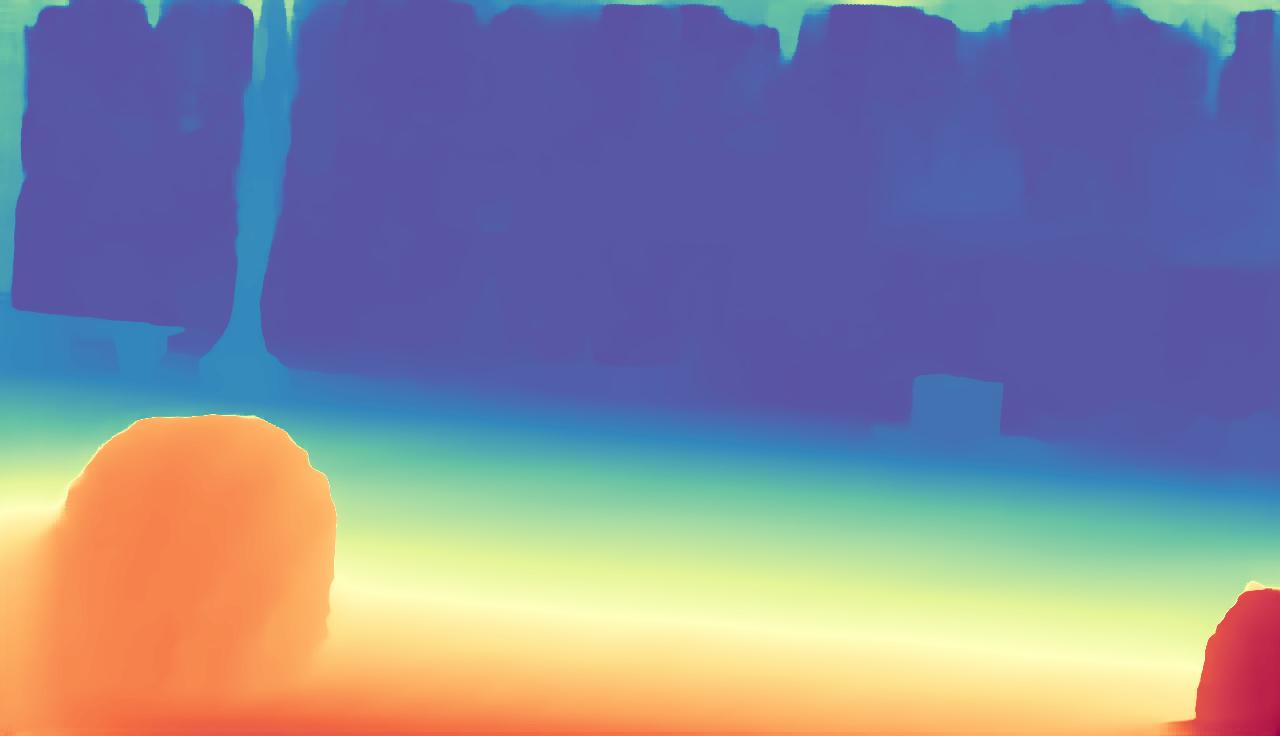} & \includegraphics[width=0.225\textwidth]{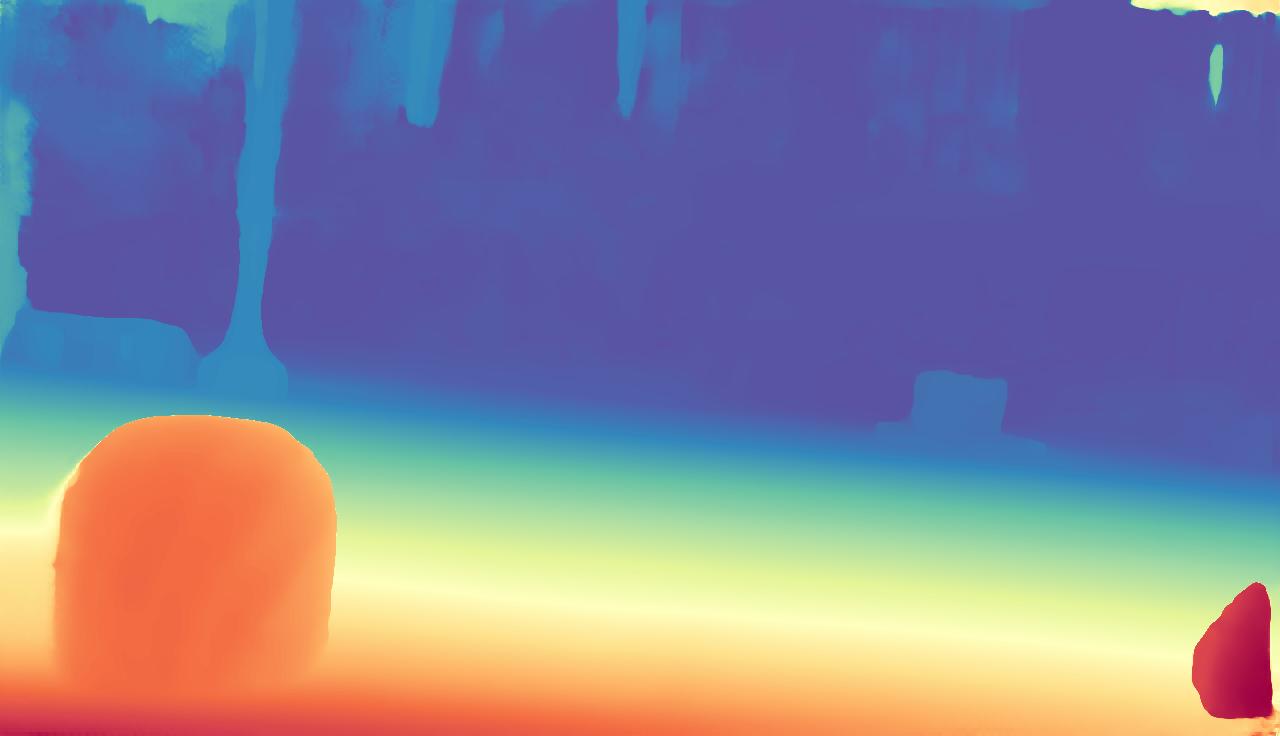} \\
	\includegraphics[width=0.225\textwidth]{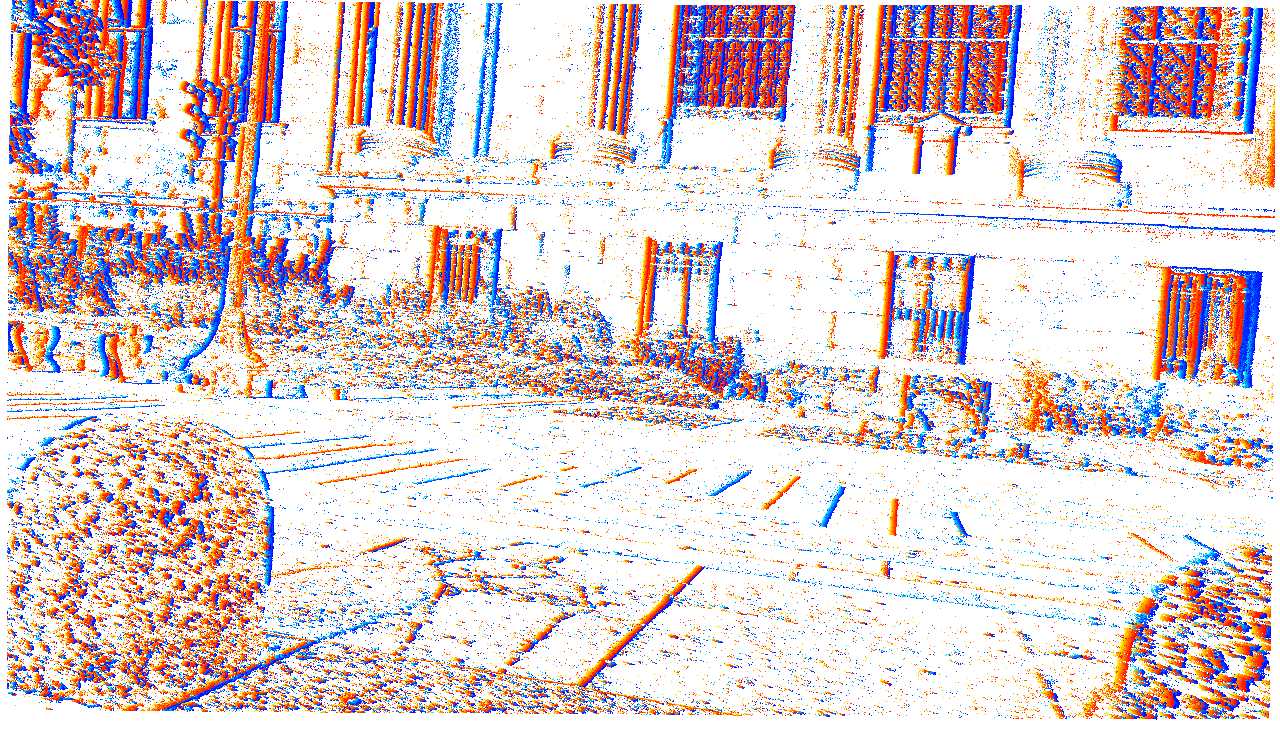} & \rotatebox[origin=l]{90}{\hspace{2em}\centering\scriptsize\textbf{MIX 3}} & \includegraphics[width=0.225\textwidth]{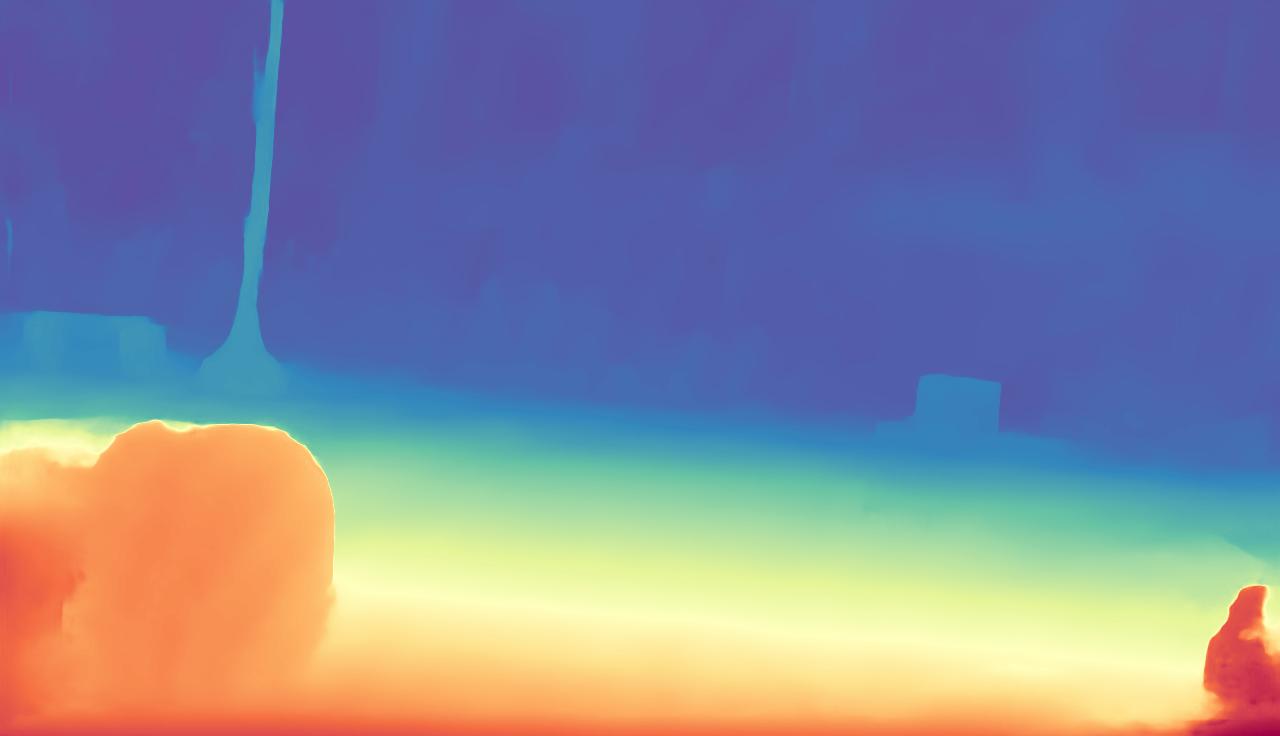} & \includegraphics[width=0.225\textwidth]{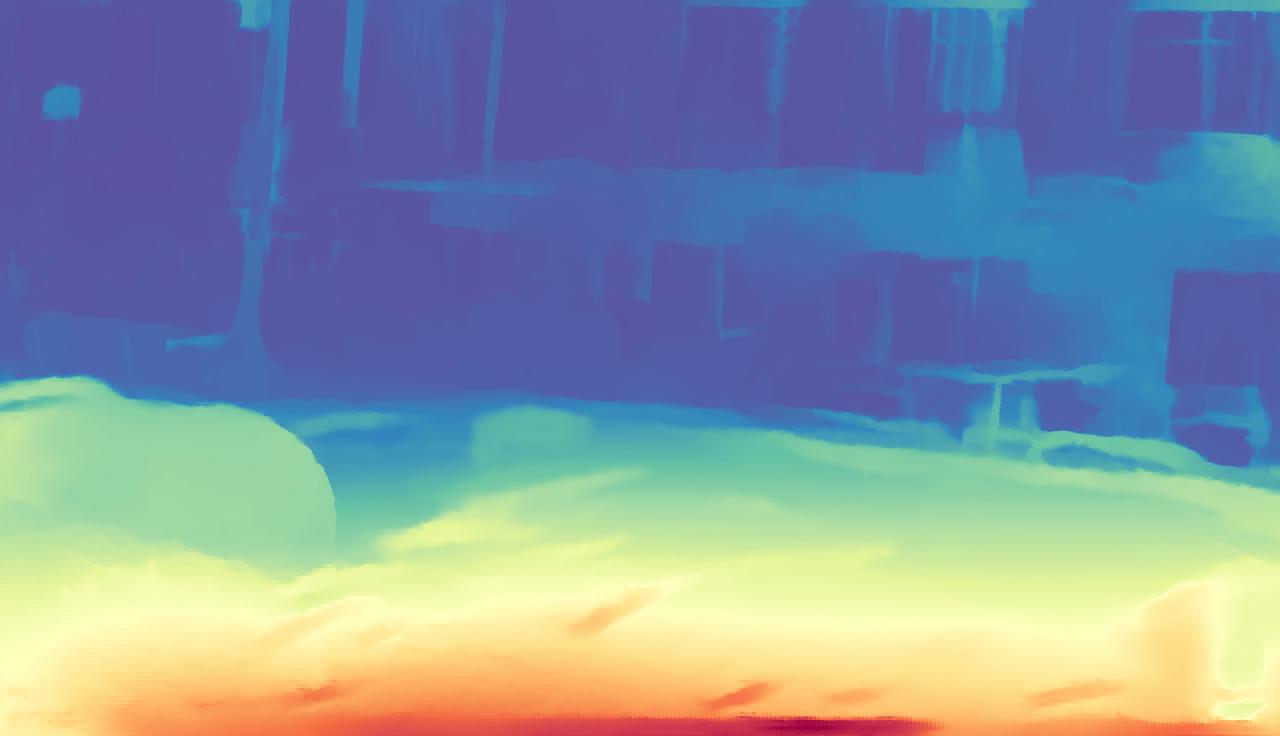} & \includegraphics[width=0.225\textwidth]{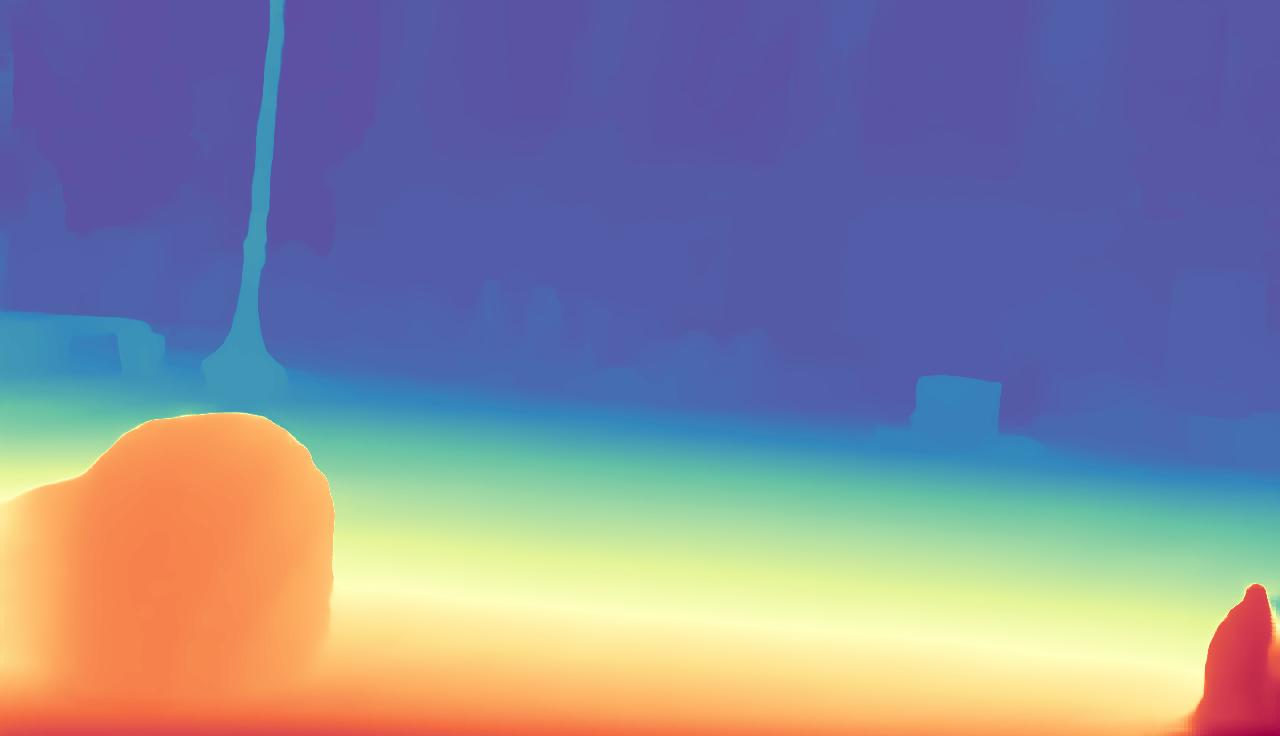} & \includegraphics[width=0.225\textwidth]{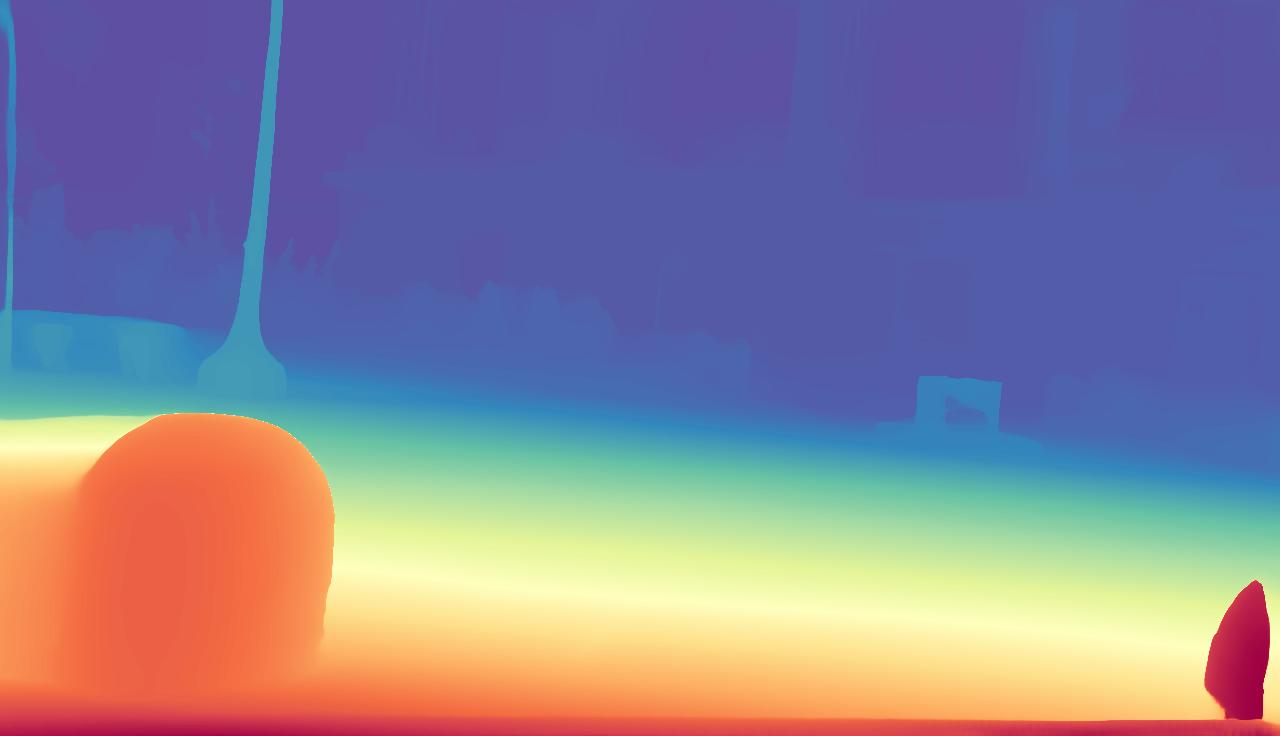} \\
	\includegraphics[width=0.225\textwidth]{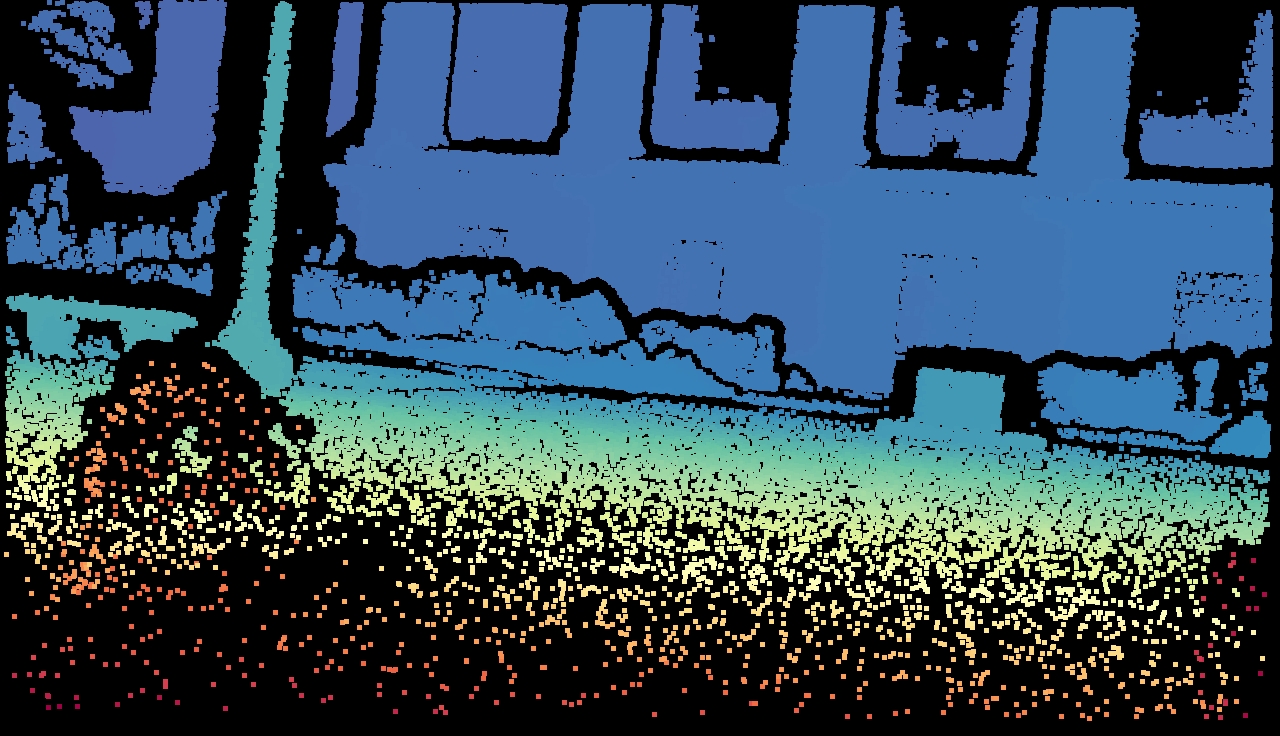} & \rotatebox[origin=l]{90}{\hspace{2em}\centering\scriptsize\textbf{MIX 4}} & \includegraphics[width=0.225\textwidth]{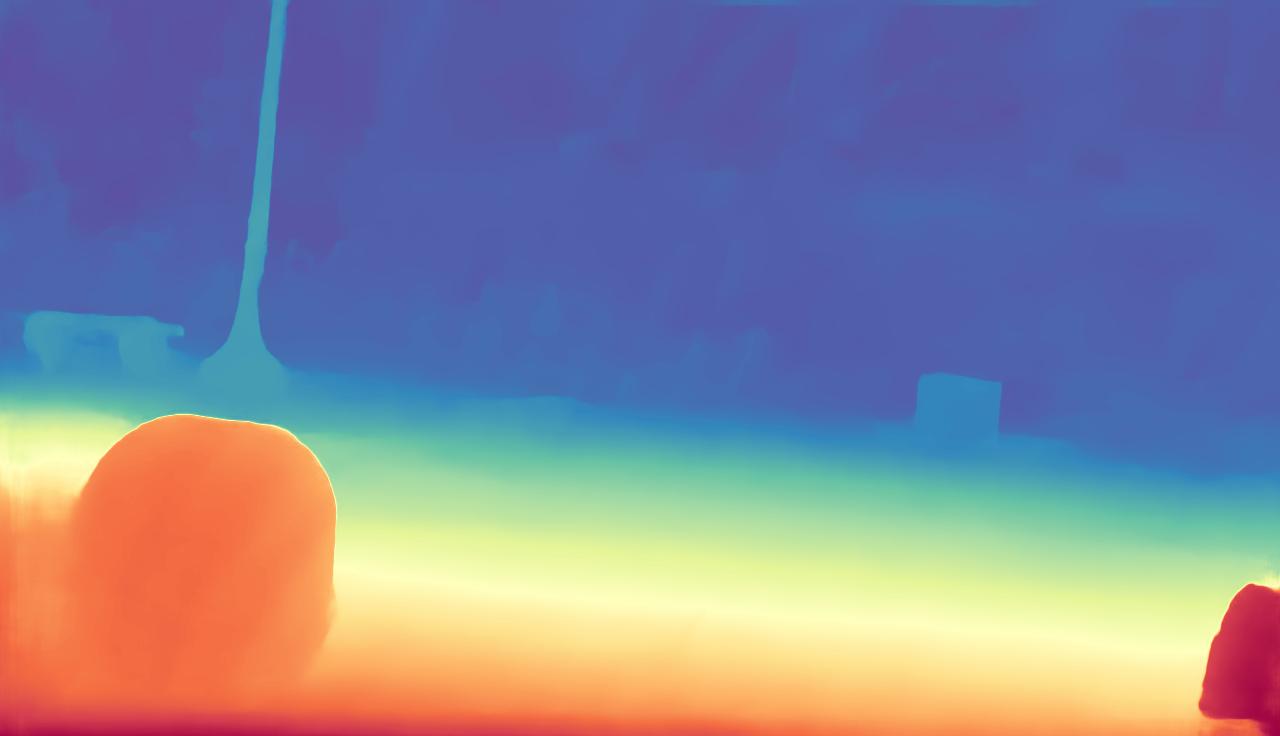} & \includegraphics[width=0.225\textwidth]{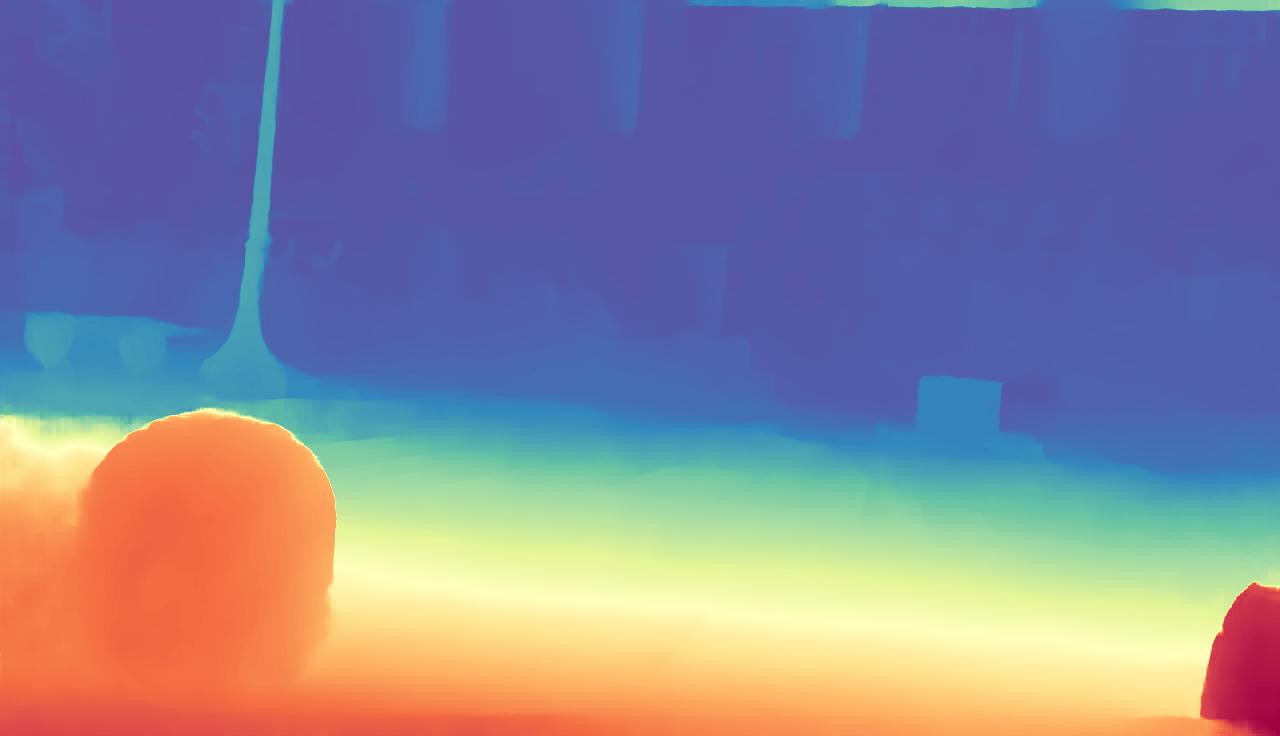} & \includegraphics[width=0.225\textwidth]{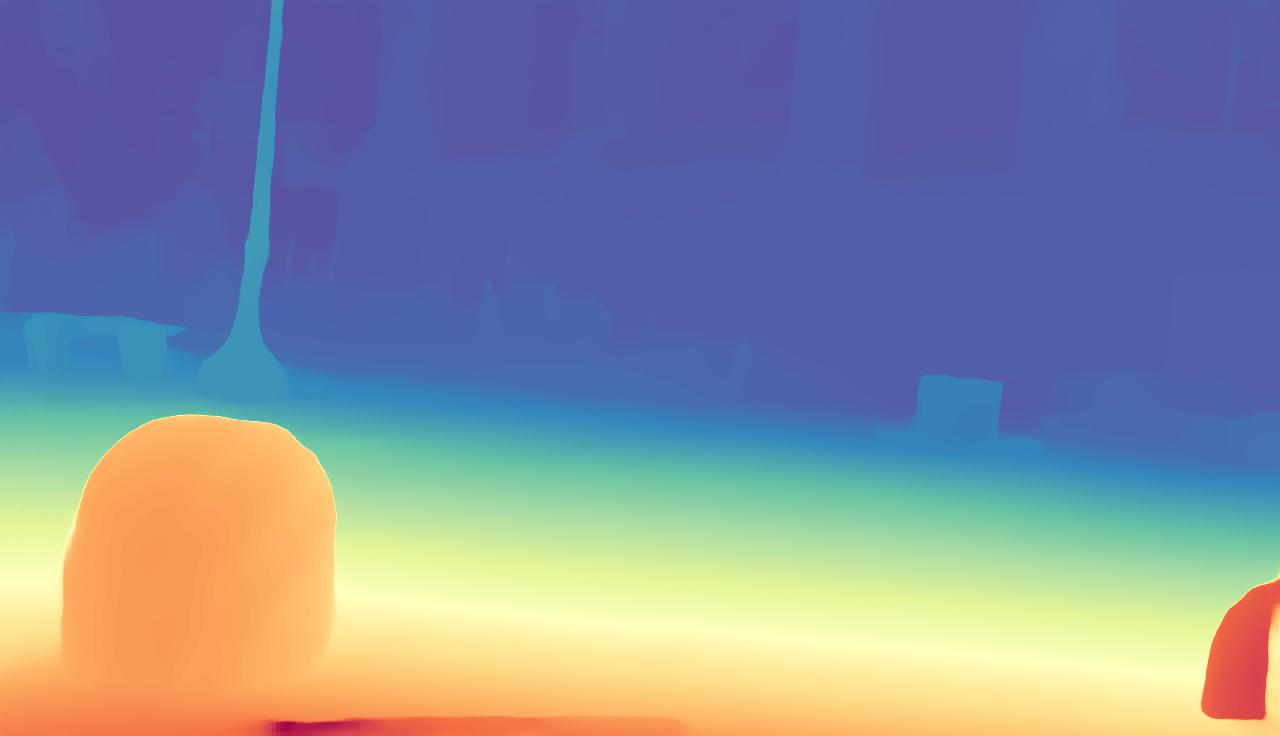} & \includegraphics[width=0.225\textwidth]{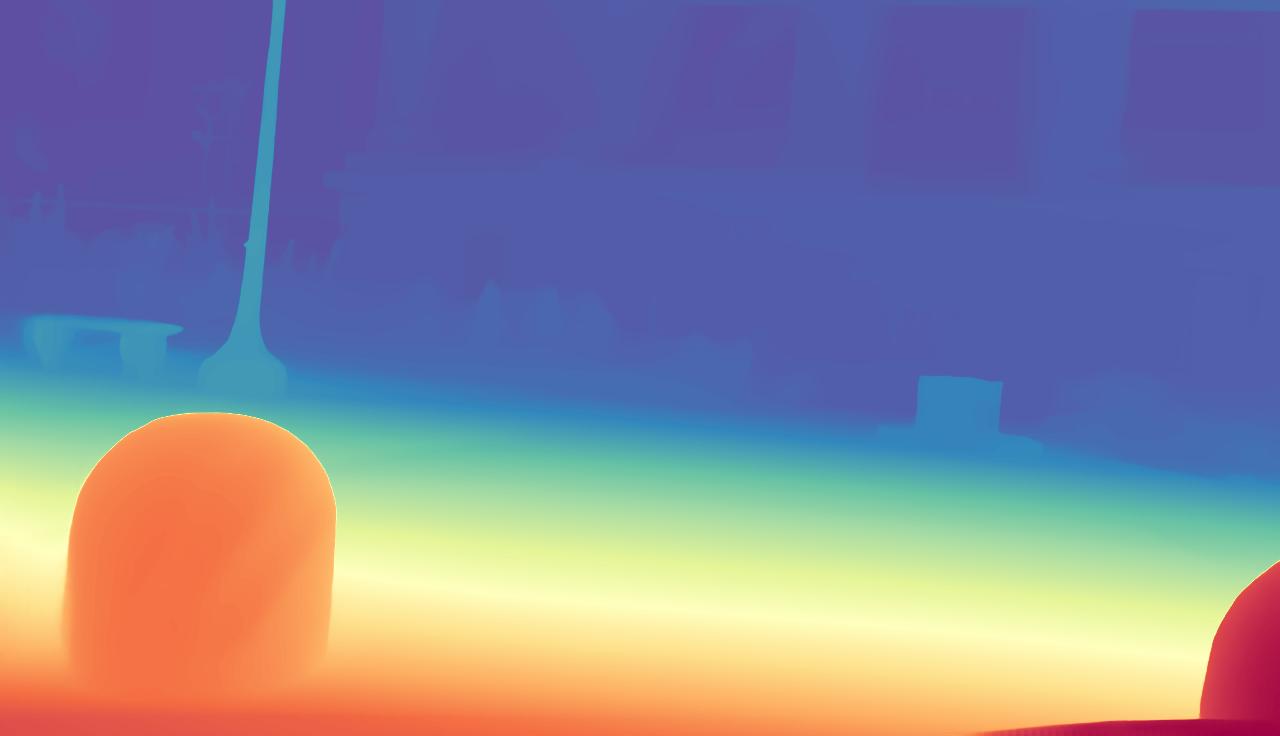} \\
	\end{tabular}
}\vspace{-0.3cm}
\caption{\textbf{Qualitative results on M3ED \cite{chaney2023m3ed} dataset.} Predictions by the four models trained with LiDAR labels, MIX 3 or MIX~4.}\vspace{-0.3cm}
\label{fig:qualitative_m3ed_outdoor_day_spot_outdoor_day_art_plaza_loop_00005}
\end{figure*}

\begin{figure*}[t]
\centering
\renewcommand{\tabcolsep}{1pt}
\resizebox{1.0\textwidth}{!} {
	\begin{tabular}{cccccccc}
	\scriptsize\textbf{Events \& Ground Truth} &  & \scriptsize\textbf{SE-CFF}~\cite{nam2022stereo} & \scriptsize\textbf{EMatch}~\cite{zhang2025ematch} & \scriptsize\textbf{E-StereoAnywhere} & \scriptsize\textbf{E-FoundationStereo} \\
	\includegraphics[width=0.225\textwidth]{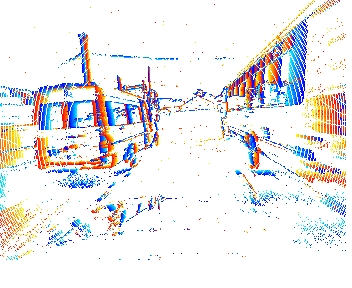} & \rotatebox[origin=l]{90}{\hspace{2em}\centering\scriptsize\textbf{LiDAR (GT)}} & \includegraphics[width=0.225\textwidth]{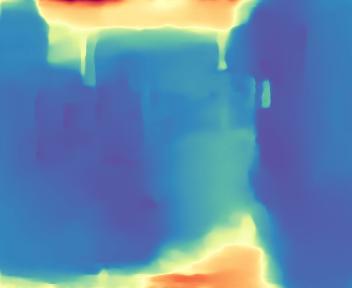} & \includegraphics[width=0.225\textwidth]{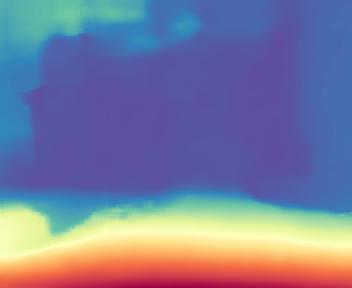} & \includegraphics[width=0.225\textwidth]{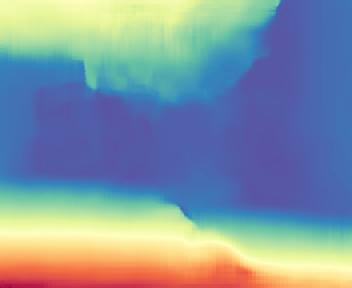} & \includegraphics[width=0.225\textwidth]{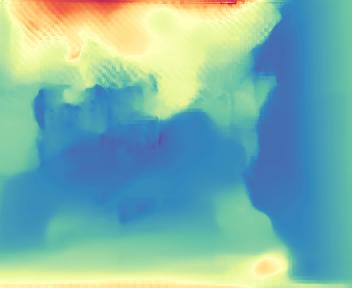} \\
	\includegraphics[width=0.225\textwidth]{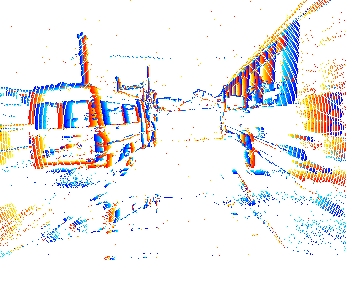} & \rotatebox[origin=l]{90}{\hspace{3em}\centering\scriptsize\textbf{MIX 3}} & \includegraphics[width=0.225\textwidth]{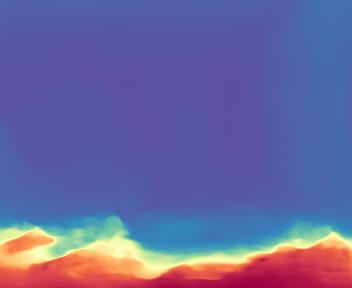} & \includegraphics[width=0.225\textwidth]{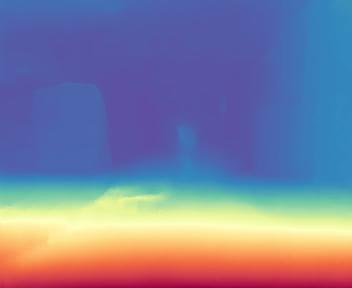} & \includegraphics[width=0.225\textwidth]{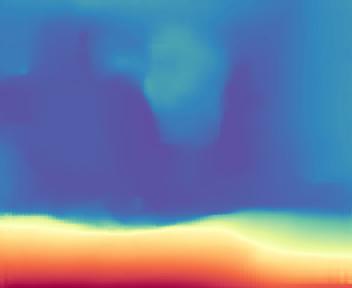} & \includegraphics[width=0.225\textwidth]{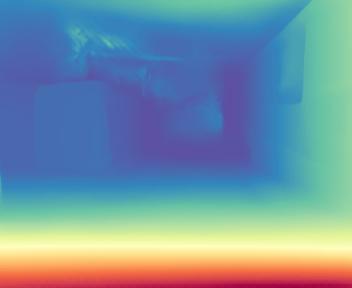} \\
	\includegraphics[width=0.225\textwidth]{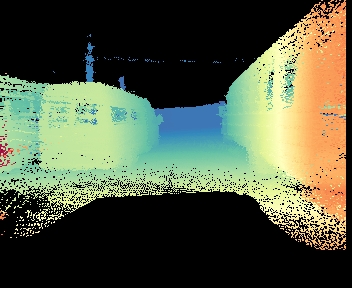} & \rotatebox[origin=l]{90}{\hspace{3em}\centering\scriptsize\textbf{MIX 4}} & \includegraphics[width=0.225\textwidth]{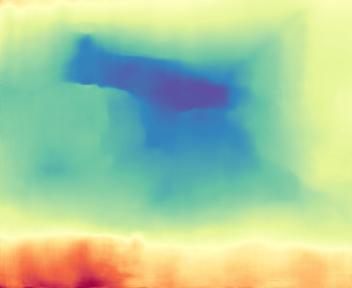} & \includegraphics[width=0.225\textwidth]{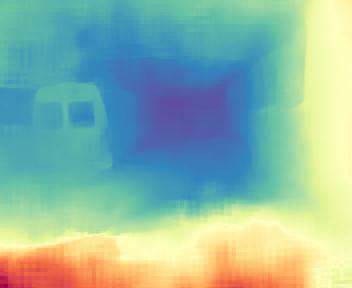} & \includegraphics[width=0.225\textwidth]{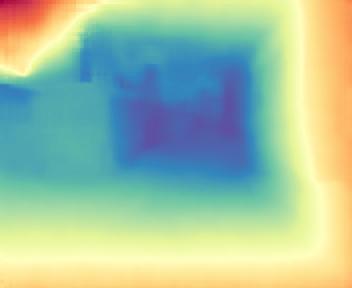} & \includegraphics[width=0.225\textwidth]{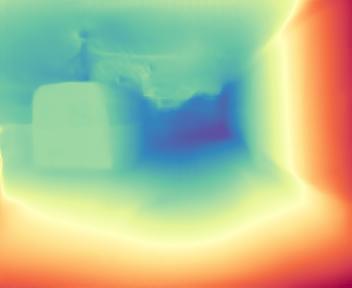} \\
	\end{tabular}
}\vspace{-0.3cm}
\caption{\textbf{Qualitative results on MVSEC \cite{zhu2018multivehicle} dataset.} Predictions by the four models trained with LiDAR labels, MIX 3 or MIX~4.}
\label{fig:qualitative_mvsec_outdoor_day_outdoor_day1_00015}
\end{figure*}

\subsection{Predictions from RGB SFMs at Night}
\label{subsec:predictions_sfm_night_qualitatives}

We conclude by showing qualitatively how we can improve the original stereo foundation models -- StereoAnywhere and FoundationStereo, from which we derived our new E-StereoAnywhere and E-FoundationStereo frameworks -- on challenging conditions where they struggle, by distilling the knowledge of E-StereoAnywhere and E-FoundationStereo themselves. 

\Cref{fig:backproxy_fs_vitl,fig:backproxy_fs_vits} collect two nighttime images from DSEC \cite{gehrig2021dsec} each. 
From left to right, we show (a) the left color image, then the predictions by FoundationStereo \cite{wen2025foundationstereo} respectively (b) before any further fine-tuning -- i.e., using the original weights -- and (c) after being fine-tuned on proxy labels distilled by E-FoundationStereo.
After fine-tuning, FoundationStereo learns to deal with this challenging domain and is able to better retain fine details in the predicted disparity maps.

\begin{figure}
    \centering
    \renewcommand{\tabcolsep}{1pt}
    \begin{tabular}{ccc}
        \includegraphics[width=0.32\textwidth]{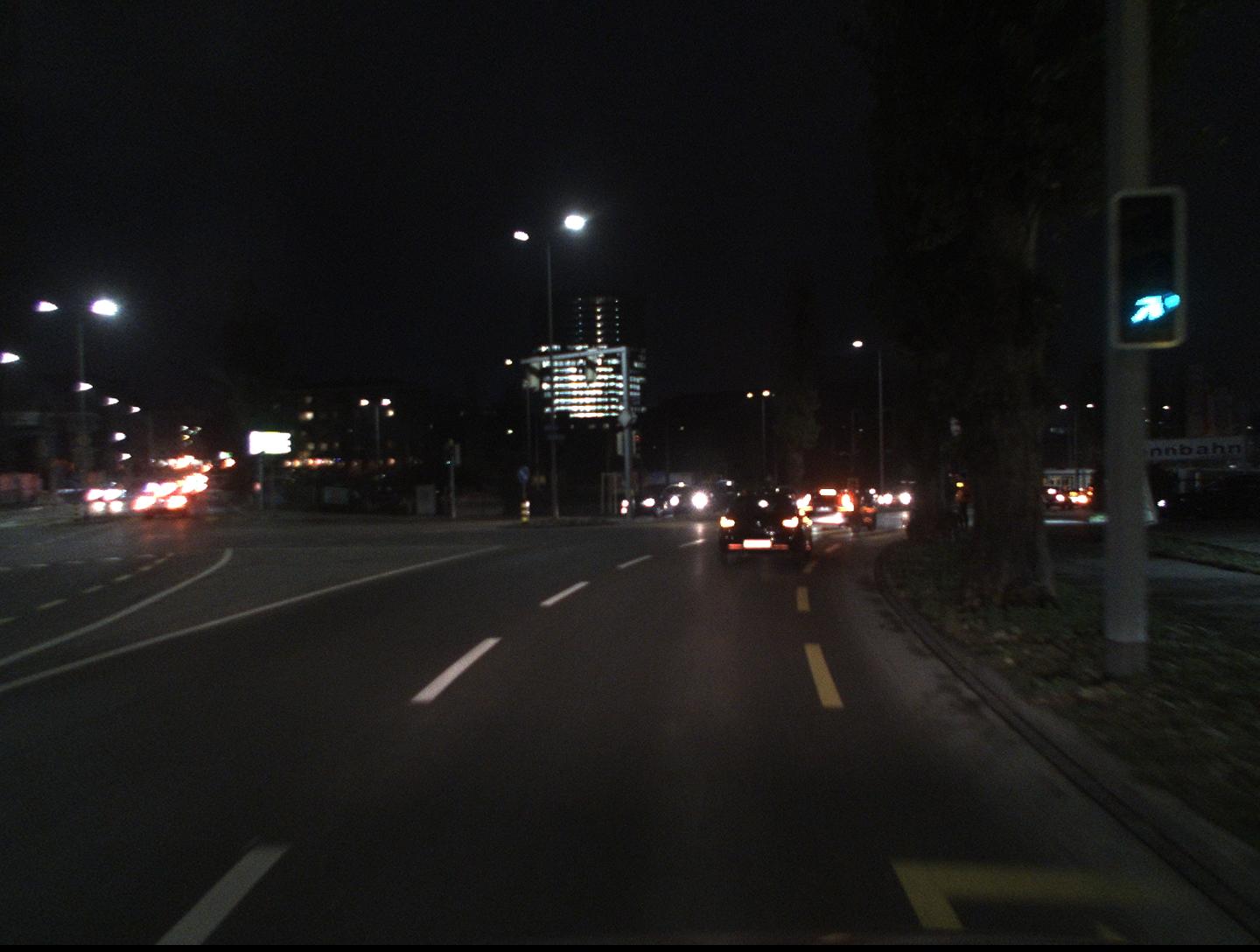} &
        \includegraphics[width=0.32\textwidth]{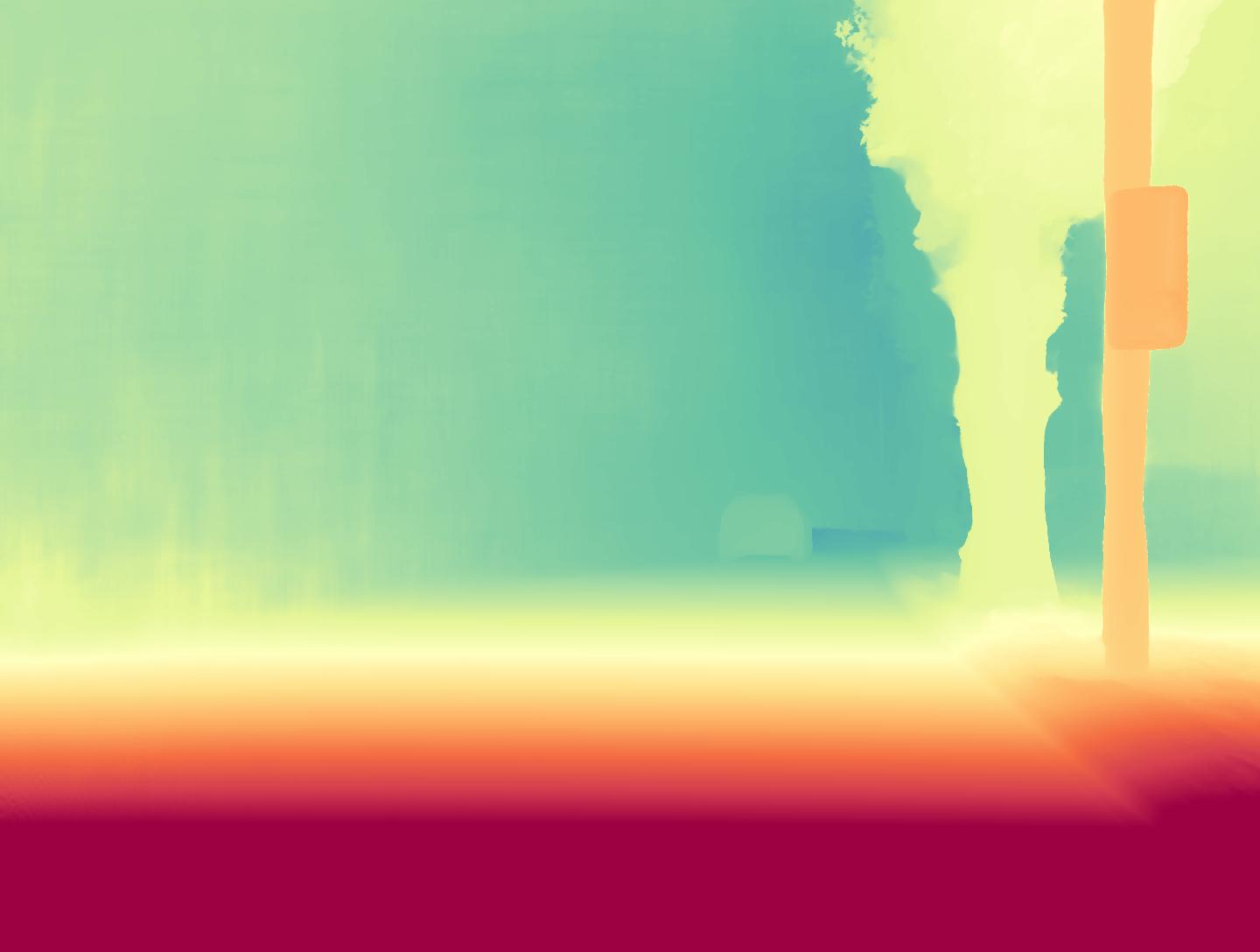} &
        \includegraphics[width=0.32\textwidth]{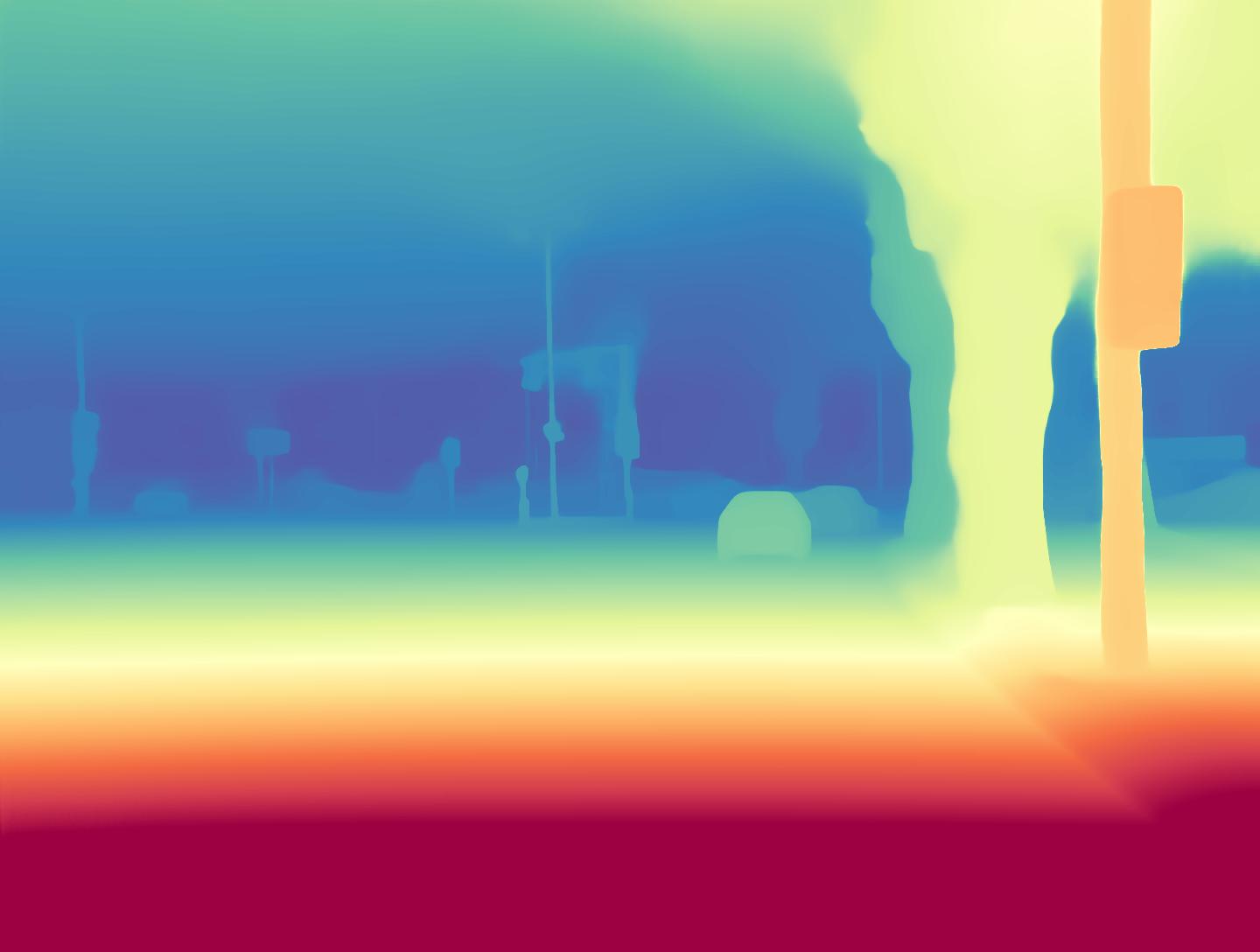}\\
        (a) & (b) & (c) \\
    \end{tabular}\vspace{-0.3cm}
    \caption{\textbf{Improved RGB FoundationStereo at Night.} Qualitative comparison on the {\footnotesize\texttt{zurich\_city\_09\_d}} night sequence. (a) left RGB, (b) prediction by baseline FoundationStereo VIT-L, and (c) its fine-tuned counterpart using proxy labels from E-FoundationStereo VIT-S.}
    \label{fig:backproxy_fs_vitl}
\end{figure}

\begin{figure}
    \centering
    \renewcommand{\tabcolsep}{1pt}
    \begin{tabular}{ccc}
        \includegraphics[width=0.32\textwidth]{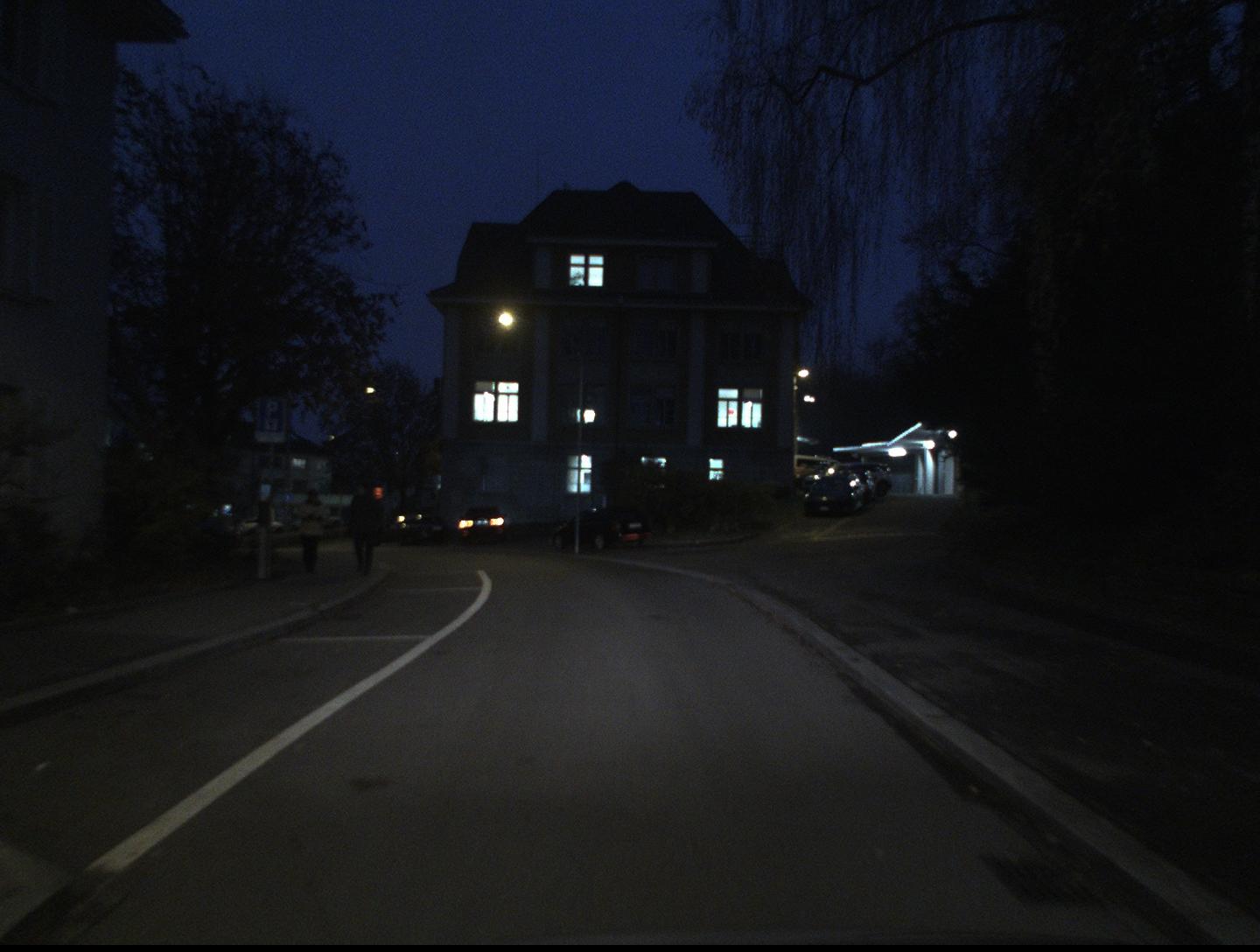} &
        \includegraphics[width=0.32\textwidth]{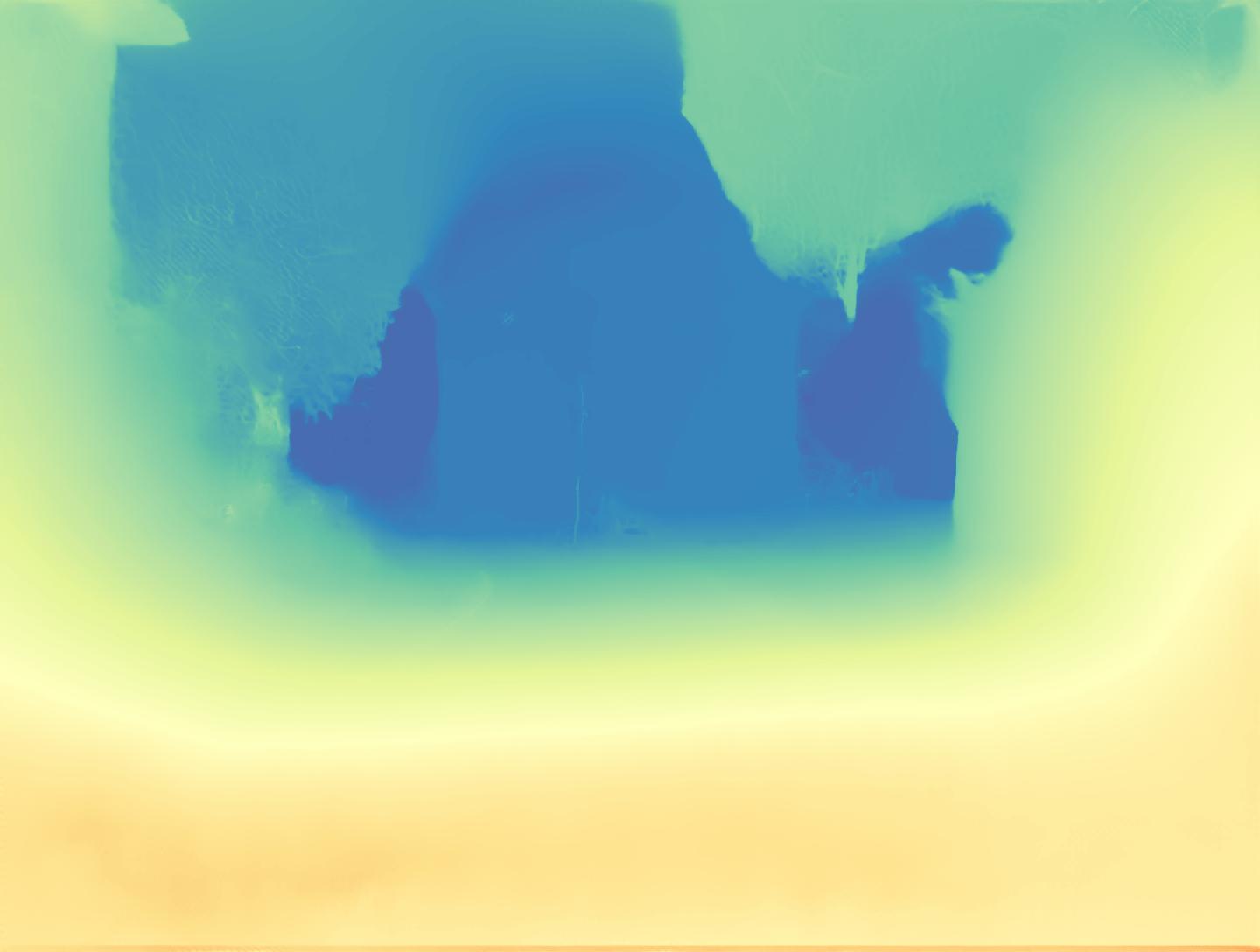} &
        \includegraphics[width=0.32\textwidth]{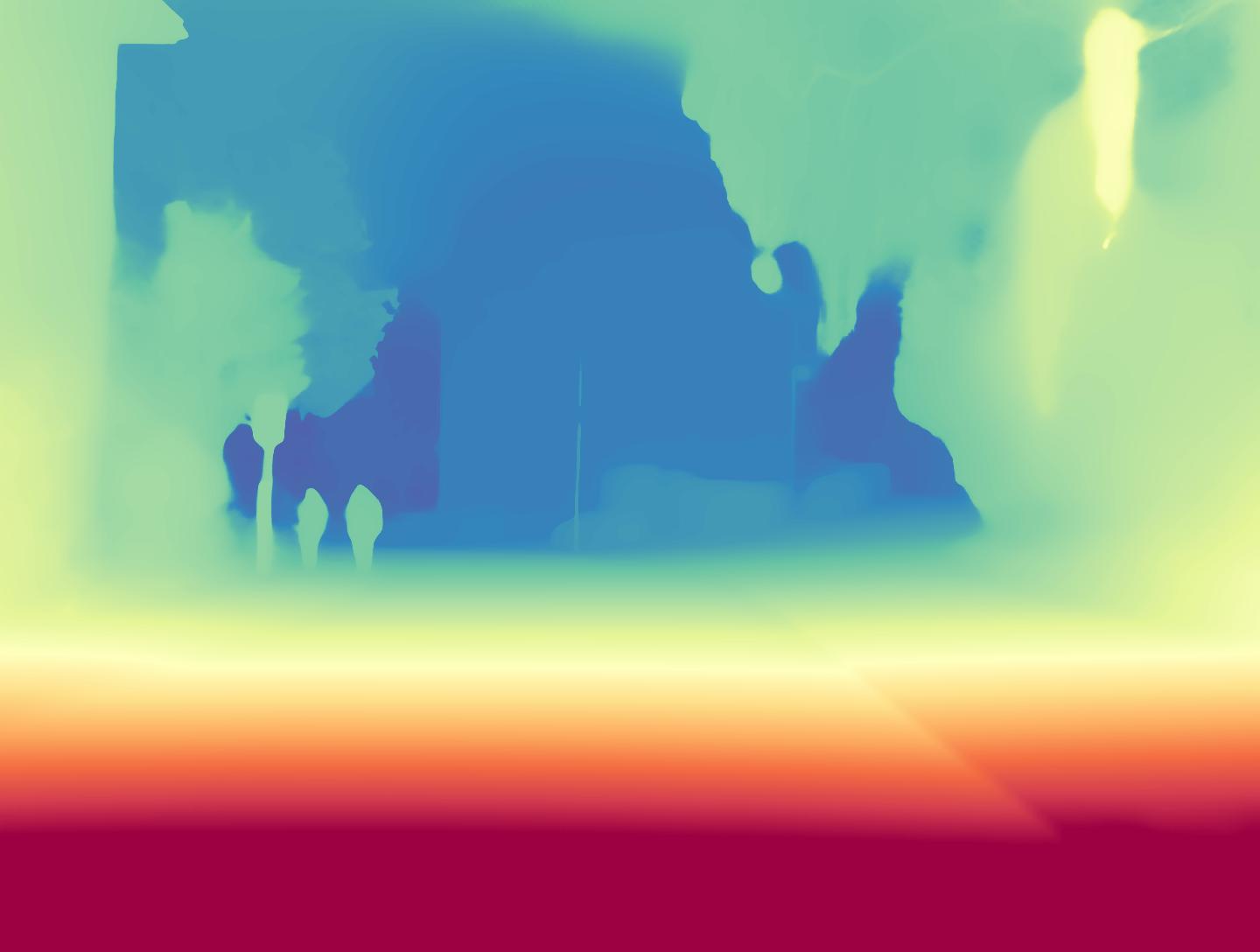} \\
        (a) & (b) & (c) \\
    \end{tabular}\vspace{-0.3cm}
    \caption{\textbf{Improved RGB FoundationStereo at Night.} Qualitative comparison on the {\footnotesize\texttt{zurich\_city\_10\_b}} night sequence. (a) left RGB, (b) prediction by baseline FoundationStereo VIT-S, and (c) its fine-tuned counterpart using proxy labels from E-FoundationStereo VIT-S.}
    \label{fig:backproxy_fs_vits}
\end{figure}

\twocolumn
\cleardoublepage
\textbf{Acknowledgment.}
The authors gratefully acknowledge the EuroHPC Joint Undertaking for awarding this project access to supercomputing resources under Proposal ID EHPC-DEV-2025D05-081.

{
    \small
    \bibliographystyle{ieeenat_fullname}
    \bibliography{main}
}

\else 

\textbf{Acknowledgment.}
The authors gratefully acknowledge the EuroHPC Joint Undertaking for awarding this project access to supercomputing resources under Proposal ID EHPC-DEV-2025D05-081.

{
    \small
    \bibliographystyle{ieeenat_fullname}
    \bibliography{main}
}

\clearpage

\fi

\end{document}